\documentclass{article}

\usepackage[english]{babel}
\usepackage{tabularx}
\usepackage[letterpaper,top=2cm,bottom=2cm,left=3cm,right=3cm,marginparwidth=1.75cm]{geometry}

\usepackage{amsmath} 
\usepackage{graphicx} 
\usepackage{url}
\usepackage{dblfloatfix}
\usepackage{longtable}
\usepackage{listings}
\usepackage{verbatim}
\usepackage{natbib}
\usepackage{makecell}
\usepackage{multirow}
\usepackage{arydshln}
\usepackage{amssymb}    
\usepackage{algorithm}  
\usepackage{algpseudocode} 
\usepackage{listings}
\title{Agent-based Condition Monitoring Assistance with Multimodal Industrial Database Retrieval Augmented Generation} 

\author{Karl Löwenmark, Daniel Strömbergsson, Chang Liu, Marcus Liwicki, Fredrik Sandin}
\begin{document}
\maketitle

\begin{abstract}
Condition monitoring (CM) 
plays a crucial role in ensuring reliability and efficiency in the process industry.
Although computerised maintenance systems effectively detect and classify faults, tasks like fault severity estimation, remaining useful life (RUL) prediction, and maintenance decisions still largely depend on human expert analysis.
The analysis and decision making automatically performed by current systems typically exhibit considerable uncertainty and high false alarm rates, leading to increased workload and reduced efficiency.

This work investigates the integration of large language model (LLM)-based reasoning agents within CM workflows to address analyst and industry needs, namely reducing false alarms, enhancing fault severity estimation, improving decision support, and offering explainable interfaces.
We propose MindRAG, a modular framework combining multimodal retrieval-augmented generation (RAG) with novel vector store structures designed specifically for CM data.
The framework leverages existing annotations and maintenance work orders as surrogates for labels in a supervised learning protocol, addressing the common challenge of training predictive models on unlabelled and noisy real-world datasets.

The primary contributions include: 
(1) an approach for structuring industry CM data into a semi-structured multimodal vector store compatible with LLM-driven workflows; 
(2) developing multimodal RAG techniques tailored for CM data; 
(3) developing practical reasoning agents capable of addressing real-world CM queries; 
and (4) presenting an experimental framework for integrating and evaluating such agents in realistic industrial scenarios.

Preliminary results, evaluated with the help of an experienced analyst, indicate that MindRAG provide meaningful decision support for more efficient management of alarms, thereby improving the interpretability of CM systems.
While promising, the approach highlights significant avenues for future enhancement, particularly concerning advanced retrieval mechanisms, long-term adaptability, and explainable severity and RUL predictions.

\pagebreak
\end{abstract}

\section{Introduction}
\label{paperE:introduction}

Condition monitoring (CM) for Predictive maintenance (PdM) of rotating machinery plays a vital role in ensuring the reliability and productivity of equipment in the process industry.
A major trend is the increased volume of data gathered, processed, and diagnosed to optimise sustainability, safety, and efficiency \cite{bleicher1991konzept, randall2011rolling, schwab2017fourth}.
Fault diagnosis is primarily done by computerised condition monitoring systems (CMS), which detect and classify faults, and primarily rely on human expert analysts for fault severity estimation (FSE), remaining useful life (RUL) prediction, and maintenance decisions, due to the complexity of these tasks.
CMS typically have high recall for fault detection and classification, but lower precision.
As a result, such systems are robust at detecting faults, but prone to false alarms, with less than 20\% of alarms being considered informative.
This increases the workload of analysts and limits time available for maintenance optimisation and root cause analysis.
Based on interviews with industry analysts from two major process industries in northern Sweden and an international equipment and CM systems supplier, four main desirable objectives (O1-O4) to current systems have been identified:
\begin{itemize}
    \item[\textbf{O1}] Reduced frequency of false alarms, in particular as a result of cable and sensor faults.
    \item[\textbf{O2}] More informative alarms with accurate FSE.
    \item[\textbf{O3}] Improved decision support and historic insights.
    \item[\textbf{O4}] Explainable models with simple interfaces for upskilling.
\end{itemize}

Intelligent fault diagnosis (IFD) is a machine learning (ML) based subset of CM, and has been extensively studied with the goal to use the vast amounts of data generated in for instance process industries to learn fault development features for improved fault diagnosis \cite{tiboni2022review}.
However, (publicly available) industrial CM datasets with reliable fault development labels are not available as far as we know, and such data is commonly characterised by heterogenous noise and feature spaces with non-linear fault development \cite{sergio_dataset, lowenmark2023dataset}.
Research has thus been primarily done on labelled lab test rig data with the goal to transfer models from lab to field domains \cite{roadmap, transfer_learning_RUL_review, deep_transfer_learning_review, deep_transfer_learning_review_2}.
Transfer learning and domain adaptation are promising tools, 
which despite many innovations remains challenging due to the aforementioned lack of labels and heterogenous environments \cite{transfer_learning_review}.

While labels are missing, industry datasets often feature \textit{annotations} and maintenance work orders (MWOs) with fault descriptions attached to signals from historic cases of fault development \cite{lowenmark2022processing}, which offer information similar to labels that can be used to facilitate IFD \cite{lowenmark2021technical, lowenmark2023labelling}.
Recent advances in large language model (LLM) reasoning have facilitated chatbot agents capable of imitating human reasoning to solve complex tasks by chain-of-thought reasoning and scaling of inference time \cite{kwa2025measuring}.
Such agents can load and analyse custom data unseen during training or in the prompt through for instance retrieval-augmented generation (RAG) \cite{Lewis2020}.
Therefore, we pose the question: can such reasoning agents be integrated in a CM workflow to assist with alarm management, severity assessment, decision support, and natural language-interfacing and explanations?

Aiming to address these questions, this paper presents four contributions:
\begin{enumerate}
    \item We introduce a new way to structure process industry CM data with a complex hierarchical structure of assets into a semi-structured multimodal graph-like vector store, compatible with machine learning models and especially LLM agent usage;
    \item We extend the RAG technique to work on multimodal CM data by adding new retrieval and generation mechanisms;
    \item We develop agents that can use these tools to answer relevant questions to meet industry needs;
    \item We present an experimental framework to implement and test an AI assistant for CM workflows based on items 1-3 above.
\end{enumerate}
%

\begin{figure}[ht]
    \centering
    
    \includegraphics[width=1.00\linewidth]{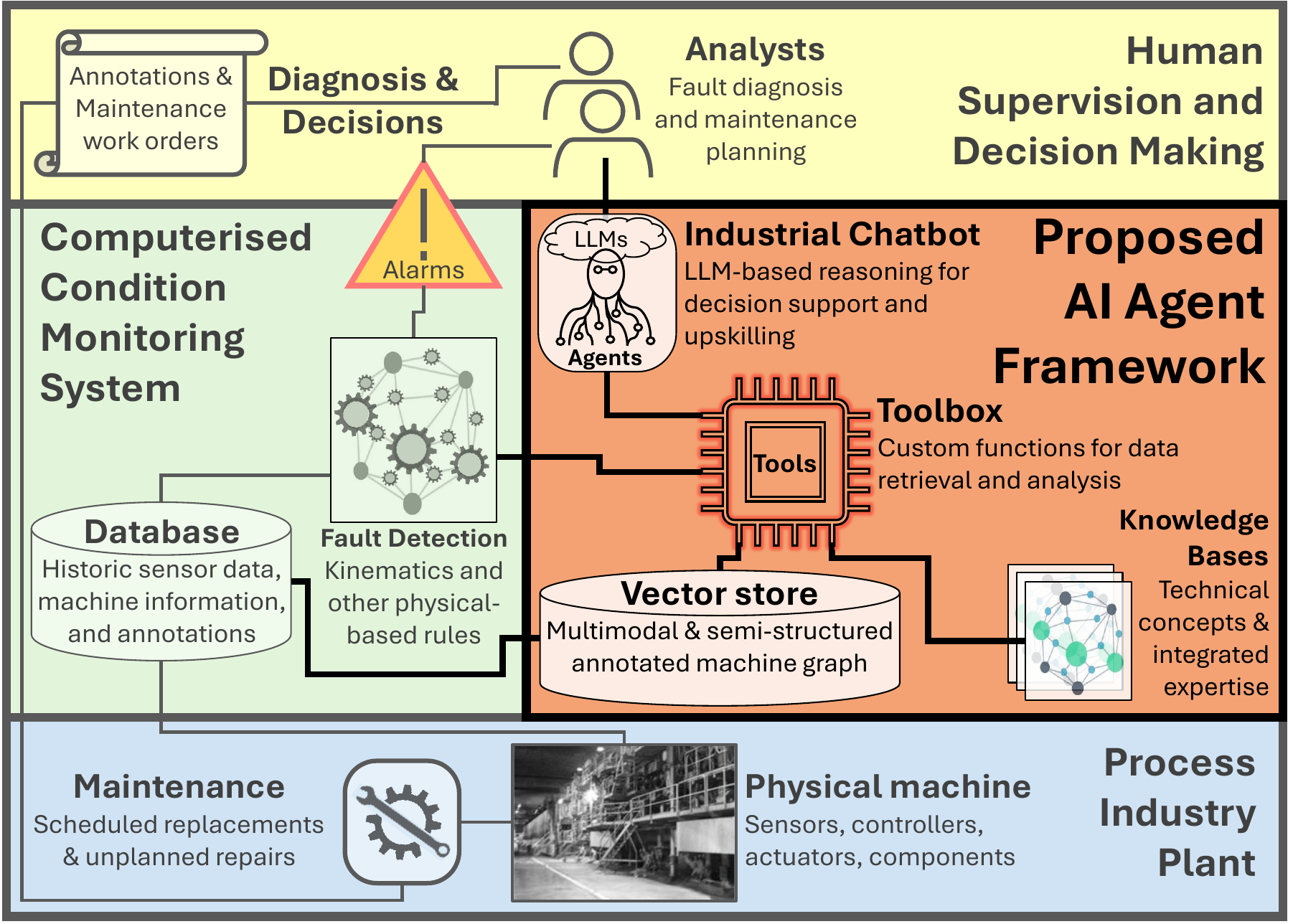}
    \vspace{-0.7cm}
    \caption{Condition monitoring for predictive maintenance enhanced with the proposed AI large language model-based agent framework.}
    \label{fig:complete_abstract}
\end{figure}

Figure \ref{fig:complete_abstract} illustrates how our approach fits into current industry systems and CM workflows: the processes start with the physical components in a process industry plant (blue), going through a computerised CMS layer (green) to human expert analysis (yellow).
The process industry plant is equipped with sensors mounted on various components, actuators, and controllers, that measure physical properties such as component vibration or sensor bias, as well as process information such as lubrication flow and shaft speed.
The resulting data samples are sent to a CMS database either due to anomaly detection conditions or scheduled sampling.
The CMS layer relies on a monitoring system to perform fault diagnosis on the sensor data transmitted to the database, primarily through fault detection with signal processing methods grounded in kinematics and physical-based properties of the components, such as features indicating ball-bass frequencies in the inner and outer race of bearings \cite{randall2011rolling}.
The amount of data analysed is far too large for human analysis for each stored recording.
Therefore, the CMS relies on rules set by analysts for each monitored component to generate alarms, such as threshold values for features computed by the CMS.
The alarms alert an analyst for follow-up analysis, and if significant fault development has occurred, follow-up annotations and MWOs are written, according to a PdM policy to minimise failure risk, machine downtime, and unnecessary maintenance interventions.

Our proposed solution offers an addition to the CMS layer by integrating LLM-based agents that can interact with the human experts through natural and technical language, and with the sensor data and CMS through \textit{tools} -- functions callable by the agent that return data about machine graph hierarchy, annotations, or sensor recording information.
The database is converted into a multimodal vector store containing machine graph information with information such as machine component names, sensor types, and machine parts; annotations, including annotation content and metadata such as date and author; and sensor data, including different data modalities, recording time, and sensor types (described in Section \ref{paper_E:vector_store}).
Agents are designed using the LangChain and LangGraph frameworks \cite{chaselangchain} and equipped with tools to retrieve and process data in the vector store (retrieval described in Section \ref{sec:retrieval}, agents in Section \ref{sec:agents}).
Knowledge bases that describe vector store, CMS, and process plant properties through both prompts, tool descriptions, and custom knowledge base files are available for agents to facilitate usage of industry-specific tools (described in Section \ref{sec:agents}).
The agents can interact with analysts through language queries and responses, with different agents having different tools to address different needs identified in Table \ref{tab:knowledge_gap}.
Thus, looking through hundreds of signal recordings and annotations to find similar past cases as guidance in a complicated fault severity analysis case is reduced to one or a few language queries, with the agent handling the tedious steps of data retrieval, analysis, and presentation.

\subsection{Task Descriptions}
To decompose the task of CM in PdM, Table \ref{tab:knowledge_gap} describes a set of analysis categories, identified in workshops with industry experts and CM researchers, and showcases the types of queries that industrial experts commonly pose, along with the corresponding data and knowledge requirements.
The industrial experts often perform text-based analyses, such as documenting equipment failures or reviewing historic maintenance logs.
Thus, there is significant potential to embed these capabilities into AI-based systems \cite{brundage2021technical}.

Table \ref{tab:knowledge_gap} is divided into two main modalities -- text-only analysis in a CM graph database, and combined text- and signal analysis in a multimodal CM graph database.
General historic insight is the least complex task, requiring only text-retrieval from a flattened corpus of historic documentations.
The example questions highlight specific agent requirements, such as knowing what is considered a ``fault" in the specific industry database, and accurately identifying these, which can be difficult if retrieval is only done with standard RAG.
The second question exemplifies the need for AI agent reasoning;
to correctly answer this question, an agent must first find all unplanned maintenance stops, then count each fault incidence that can be considered as associated to the planned maintenance stop, then return the most common type of fault given these conditions.

\begin{table}[ht]

    \centering
    \caption{Condition monitoring tasks with corresponding example questions, and data and knowledge used by experts for the analysis. The categories are ordered in levels of difficulty, from easier to more difficult when trying to automate the task with AI.
    }
    \vspace{-0.5cm}
    \begin{tabular}{p{1.8cm}m{7.3cm}m{5.3cm}}
    \label{tab:knowledge_gap}
   \\
    \hline
    \textbf{Analysis Category} & \textbf{Example CM questions} & \textbf{Data and knowledge requirements} \\
    \hline
    \multicolumn{3}{c}{\textbf{Text-based analysis -- knowledge retrieval}} \\
    \hline
    \multirow{3}{2cm}{General historic insight} & What faults were most common this month? & \multirow{-1}{5.3cm}{Retrieve and analyse historic faults using technical language knowledge.} \\
     & What faults most commonly preclude unplanned maintenance stops? &  \\
     \hdashline
     & Which drying cylinders have required the most extra lubrication? &  \\
     
     \multirow{-3}{2cm}{Context-ualised historic insight} & Which sensor positions have had the most faults? & \multirow{-3}{5.3cm}{Retrieve and analyse historic faults w.r.t machine graph components using technical language and machine-specific knowledge.}\\
     
     & Which components have required the most unplanned stops or critical faults? & \\
    \hline
    \multicolumn{3}{c}{\textbf{Text-and-signal-based analysis -- intelligent decision assistance}}\\
    \hline
    \multirow{4}{2cm}{Fault detection \& classifi-cation} & What bearing faults are currently being monitored? & \multirow{2}{5.3cm}{Detect and classify faults based on machine graph sensor data using condition monitoring knowledge and contextual knowledge} \\ 
    & Are there any fault indications in areas where maintenance is planned within a month? & \\
    & & \\
    \hdashline
    & & \\
    \multirow{4}{2cm}{Fault severity estimation} & What alarms require immediate attention? & \multirow{4}{5.3cm}{Estimate fault severity based on machine graph sensor data using condition monitoring and machine-specific knowledge.}\\
    & What components are trending towards failure? &  \\
    & Which bearing faults have deteriorated the most since last week? & \\
    & & \\
    \hdashline
    \multirow{7}{2cm}{Remaining useful life prediction \& maintenance recommen-dations} & What maintenance actions are most urgent right now? & \multirow{5}{5.3cm}{Plan maintenance action based on fault severity, time until planned maintenance stop, fault effects on production, etc., using condition monitoring and tacit knowledge}\\
    & How long until this component will require replacement? & \\
     & What action would you suggest? & \\
    & Are there any other nearby components that warrant replacement at the next stop? & \\
    \hline
    \end{tabular}
\end{table}

For contextualised historic insight, the agent must perform the same reasoning steps, but also take machine graph hierarchy information into account, such as different machine types (``which drying cylinders") and sensor positions, and use those as part of analysis or retrieval.
To facilitate these types of queries, an industry-specific approach with custom retrieval functions can be leveraged, extending a typical RAG approach to CM data.
For example, ``Which sensor positions have had the most faults?" could be solved by reading all annotations in a database, sorting them into different sensor positions, then checking how many faults are in which class, which is expensive and scales poorly with larger datasets.
However, if the agent knows all types of historic faults and sensor positions, and can query conditioned with CM-specific retrieval mechanisms, it can instead ask for number of faults given a specific sensor position, then repeat this for each sensor position.

Fault diagnosis tasks based on signal analysis are here split into thee subtasks -- fault detection \& classification, fault severity estimation, and remaining useful life prediction \& maintenance decisions, based on how difficult each task is to model, and what impacts it has on CM workflows.
In general, to answer these questions, an agent must be equipped with tools to jointly retrieve signal and text data, through for instance a multimodal RAG approach built around the graph-like hierarchy of CM data from a process industry machine.
Fault detection \& classification are grouped as they serve primarily to generate alarms, and already operate with high recall in CM workflows.
A chatbot agent could assist analysis with these tasks through improved alarm management, both by improving precision and earlier detection, and through a natural language interface which facilitates seamless organisation of alarms based on language inputs.
As with text-only questions, the agent must use tools and knowledge bases to include only relevant data into further analysis.
The agent then relies on past human analysis accessed through annotations to analyse signals and summarise the results, to for instance provide insight into areas with planned maintenance which have undetected faults that might warrant intervention.

FSE is considerably more difficult, and analysts highlight this function as the most impactful area of improvement.
A chatbot augmented with the ability to estimate severity of faults based on signal features and related historic insights would help analysts prioritise analysis efforts and ask questions such as ``Which bearing faults have deteriorated the most since last week?".
To answer this question, an agent must first search for what bearing faults are currently monitored, or alternatively first analyse all data to detect possible bearing faults.
The severity of detected faults should then be estimated based on data from the last week, though prior data can be included in the model if trend-behaviour is important for the agent's FSE.
Finally, severities have to be compared across different modes of degradation to provide the user with a relevant answer.

RUL prediction \& maintenance recommendations are the final stage of improved AI CM assistance; a model capable of accurate and robust RUL prediction would be very close to having the tools to autonomously implement or recommend maintenance policy.
A RUL-enhanced agent should be able to prioritise maintenance actions, highlight components with low remaining run-time, and suggest how to proceed.
To predict RUL, especially with changing fault evolution parameters such as component load, a model for how fault severity will evolve is desirable \cite{transfer_learning_RUL_review}.
While our agents have access to annotations describing maintenance actions, and thus can make maintenance recommendations based on historic cases, they lack a robust view of how the fault will evolve, and merely interpolate when maintenance is prudent based on similarities.
Thus, maintenance predictions are primarily indicators of fault severity, not an accurate and robust representation of actual RUL, and the agent is designed with the goal to improve historic insights \& upskilling and alarm management mainly, while the experimental framework presented towards the end offers a path towards complete AI assistance in CM for the process industry.

%


%

\bigskip

The goal of this study is to investigate and develop an agent framework including a novel vector store and custom tools and knowledge bases to meet the needs identified in Table \ref{tab:knowledge_gap}.
Key questions and hypothesised answers are:
\begin{itemize}
    \item[\textbf{Q1:}] \textbf{How can language-based domain/tacit knowledge be integrated with condition monitoring models?}
    \vspace{-8pt}
    \begin{itemize}
        \item[A1:] LLMs with access to annotations and a growing knowledge base.
    \end{itemize}
    \item[\textbf{Q2:}] \textbf{How can LLM chatbot agents be used to reduce the impact of common false alarms?}
    \vspace{-8pt}
    \begin{itemize}
        \item[A2:] LLM chatbot agents with bespoke tools to load, navigate, and process the condition monitoring data.
    \end{itemize}
    \item[\textbf{Q3:}] \textbf{How can LLM chatbot agents facilitate decision support?} 
    \vspace{-8pt}
    \begin{itemize}
        \item[A3:] Through language-based historic insights and knowledge transfer through  Multimodal data compilation, text-to-signal interface, integration of tacit and domain knowledge from annotations and expert interactions.
    \end{itemize}
\end{itemize}
%

\subsection{Paper Structure}
This paper is organised as follows.
Section~\ref{sec:related_work} provides and overview of the background and related work of LLM agents and RAG approaches to data similar to process industry CM data.
Section~\ref{paper_E:vector_store} describes how CM data is transformed from an unstructured database backup to a semi-structured graph-like multimodal CM vector store.
The retrieval functions that can operate on the vector store are then described in Section~\ref{sec:retrieval}.
%
%
Section~\ref{sec:agents} describes tools that facilitate agent calling of the RAG functionality, agents capable of using these tools, and how the tools were evaluated.
Section~\ref{sec:framework} describes how agents can be integrated in a CM workflow and experimental frameworks to evaluate this integration.
Section~\ref{PaperE:sec:results} describes the results of the experiments from Sections~\ref{sec:agents} and ~\ref{sec:framework}.
Section \ref{PaperE:sec:discussion} discusses these results and the implications of them.

\section{Background}
\label{sec:related_work}

\subsection{Large Language Model Agents}
LLM agents combine the text processing capabilities of modern LLMs with the customisability of autonomous agents equipped with various tools to integrate data and handle tasks beyond the scope of normal LLMs \cite{agentic_AI_review}.
The LLM is typically a transformer-based natural language model \cite{vaswani2017attention}, trained on massive text corpora via self-supervised learning to capture linguistic patterns and world knowledge \cite{brown2020language}.
The LLM functions as a probabilistic maximum likelihood generative model: given an input prompt or context, it produces an output by sequentially predicting the most likely next token.
Despite relying on statistical correlations rather than explicit symbolic logic, sufficiently large models can exhibit sophisticated reasoning and abstraction capabilities.
This enables an LLM to serve as the ``brain" of an agent, interpreting instructions and mapping them to actions (in language form).
In AI terms, the LLM agent fits the definition of an intelligent agent perceiving its environment (through input text) and acting upon it via outputs—towards achieving specified goals \cite{agentic_AI_review}.
The theoretical basis for LLM agents thus combines data-driven generalisation with the sense-act cycle from classical agent frameworks, yielding flexible behaviour without task-specific programming.

The LLM part of LLM agents excel in scenarios that involve unstructured data, open-ended reasoning, or multi-step problem solving where traditional approaches struggle.
For example, in analysis of free-form text, LLM-based agents can integrate context, extract relevant information, and generate coherent summaries or answers, surpassing rule-based systems that require structured input \cite{brown2020language}.
By adding tools for external function calls and processing of specific modalities of data, they can also interact with structured knowledge sources like knowledge graphs in a flexible way.
An LLM agent can translate natural language queries into graph traversals or interpret graph-derived facts in context, enabling reasoning over knowledge graphs without manual query formulation \cite{pan2024unifying}.

LLM agents are adept at complex, multi-step decision-making tasks.
Techniques such as chain-of-thought prompting \cite{wei2022chain} and the ReAct paradigm \cite{yao2023react} allow the model to decompose a problem into intermediate reasoning steps and actions, iteratively refining its approach.
This capability to plan and reason over several turns (e.g., asking clarifying questions, retrieving information, then synthesising an answer), facilitates scaling not only model, input data, and training, and but also inference time \cite{sardana2023beyond, snell2024scalingllmtesttimecompute, wu2024inference}.
%
%
%
%
In essence, when a task requires parsing ambiguous, unstructured inputs or dynamically combining evidence across modalities or steps, LLM agents provide a more 
general solution than bespoke algorithms.

Frameworks like LangChain and LangGraph have emerged to facilitate such structured reasoning and retrieval-augmented generation (RAG) with LLMs \cite{chaselangchain}.
LangChain provides a modular toolkit for building LLM-centric applications, allowing developers to compose sequences of prompt executions, tool calls, and memory lookups as reusable ``chains"; it also supports agentic behavior, where an LLM chooses which action to take next (e.g., whether to invoke a search tool or answer directly) instead of following a fixed script, an approach that makes it straightforward to implement for instance RAG workflows \cite{Lewis2020}.

In a RAG system, the agent first retrieves relevant context from a knowledge base or vector store, and then generates a response grounded in that retrieved information.
LangGraph extends this idea by introducing a graph-based orchestration framework for LLM agents.
Rather than a linear chain, LangGraph represents the agent's workflow as a directed graph of nodes  through which an agent state is passed, where each node might be a prompt, a computation, or a data retrieval step.
By explicitly encoding the possible decision paths and maintaining state, e.g. the conversation history or intermediate results, LangGraph allows for fine-grained control over multi-step reasoning processes, including conditional branching and parallel subtasks.
These frameworks illustrate how imposing structure and integrating external knowledge retrieval into LLM agents can enhance reasoning reliability and factual accuracy, enabling complex decision-making systems that are less prone to knowledge gaps or the so-called hallucination effect \cite{dhuliawala2023chain}.

\subsection{Retrieval-Augmented Generation}
\label{sec:rag-intro}
Retrieval-augmented generation is a framework that integrates LLMs with an external knowledge source through a retrieval module to enhance factual accuracy and domain specificity of the generated outputs \cite{Lewis2020, Karpukhin2020, Borgeaud2022}.
Rather than relying solely on the internal parameters of a language model, RAG dynamically retrieves relevant information from a knowledge source 
at inference time.
This retrieval step injects new, domain-specific, or otherwise up-to-date knowledge into the generation process, making RAG particularly appealing for knowledge-intensive tasks such as open-domain question answering, summarisation, and dialogue systems~\cite{izacard2020leveraging, Izacard2022}.

This approach provides two main benefits: 
\begin{enumerate}
    \item \textbf{Adaptability:} The system can incorporate new or domain-specific knowledge without expensive model re-training. 
    \item \textbf{Transparency:} Retrieved passages or data snippets can be exposed to end-users, improving explainability and trust.
\end{enumerate}

\label{sec:rag-theory}
The RAG framework, as formulated by Lewis et al \cite{Lewis2020}, couples a retrieval module, denoted by $\mathcal{R}$, with a generative module $\mathcal{G}$, typically an LLM. Let $\mathbf{x}$ be the input query (for instance, a user question).
The retriever identifies top-$k$ relevant documents from a knowledge store $\mathcal{K}$ as
\begin{equation}
    \{\mathbf{d}_1, \dots, \mathbf{d}_k\} \;=\; \mathcal{R}(\mathbf{x}),
\end{equation}
where each $\mathbf{d}_i$ is typically a textual chunk or structured data snippet (e.g., a paragraph, table entry, or short passage).
The probability of retrieving document $\mathbf{d}_i$ given $\mathbf{x}$ is often based on dense passage retrieval
\cite{Karpukhin2020} with a bi-encoder architecture \cite{Lewis2020}:
\begin{equation}
    P_\mathcal{R}(\mathbf{d}_i \mid \mathbf{x}) \;\propto\; \exp\bigl(\phi(\mathbf{x})^\top \,\psi(\mathbf{d}_i)\bigr),
    \label{eq:retriever-prob}
\end{equation}
where $\phi(\cdot)$ and $\psi(\cdot)$ represent vector encoders for the query and documents, respectively \cite{Karpukhin2020}, typically transformer LLMs in the case of text-to-text RAG.

Conditioned on both the query and the retrieved documents, the generative model $\mathcal{G}$ produces the output $\mathbf{y}$ based on the input query $\mathbf{x}$ and the retrieved documents $\mathbf{d}$ with the joint conditional probability as
\begin{equation}
P(y \mid x) \approx 
\sum_{d \in \mathrm{Top\text{-}k}(p(d \mid r))}
P_\mathcal{R}(d \mid x) \, P_\mathcal{G}\bigl(y \mid x, d\bigr)
\;=\;
\sum_{d \in \mathrm{Top\text{-}k}(p(d \mid r))}
P_\mathcal{R}(d \mid x)
\prod_{i=1}^{N}
P_\mathcal{G}\bigl(y_i \mid x, d, y_{1:i-1}\bigr),
\label{eq:rag-joint}
\end{equation}
where $P_\mathcal{G}$ represents the generative model that constitutes $\mathcal{G}$.


The practical appeal of RAG lies in its ability to adapt to new information and specialised domains without the costly process of retraining or fine-tuning large models from scratch \cite{Borgeaud2022}.
For instance, in customer support applications, a RAG-based system can query an evolving FAQ database to accurately answer user questions about newly released products \cite{pandya2023automating}.
%
%
Similarly, in academic or scientific contexts, RAG can be useful for literature reviews and knowledge discovery, as it can retrieve scholarly articles and produce concise summaries of the most pertinent findings (albeit with the need for researchers to review model outputs to ensure scientific rigorousness) \cite{thus2024exploring}.
Beyond open-domain question answering, RAG is increasingly adopted in settings such as code generation, where retrieved code snippets provide relevant examples, and healthcare, where patient data or medical guidelines are retrieved to inform a recommendation or diagnosis\cite{cabello2024megmedicalknowledgeaugmentedlarge}.
A benefit of RAG is its \emph{modularity}; models can be paired with various retrievers and diverse knowledge sources, enabling flexible and up-to-date generation capabilities without re-training model parameters.

For many real-world tasks, especially in domains with heterogeneous data, custom scoring or weighting functions can be introduced into Eq.~\eqref{eq:retriever-prob} and Eq.~\eqref{eq:rag-joint} to align retrieval with data-specific properties.
For instance, 
\begin{equation}
    \widetilde{P_\mathcal{R}}(\mathbf{d}_i \mid \mathbf{x}) \;\propto\; \alpha \cdot \exp\bigl(\phi(\mathbf{x})^\top \,\psi(\mathbf{d}_i)\bigr) \;+\; \beta \cdot \mathrm{meta}(\mathbf{d}_i),
\end{equation}
where $\mathrm{meta}(\mathbf{d}_i)$ is a function encoding domain-specific metadata (e.g., fault type, sensor ID, etc.), and $\alpha, \beta$ are hyperparameters.
Such modifications help integrate domain knowledge to increase the likelihood that more relevant or trustworthy documents are retrieved first.
The query, documents, vector encoders, retriever, and generator, can all be extended to function on more modalities of data than pure text documents, which facilitates \textit{multimodal RAG}.

\subsection{Multimodal RAG}
\label{sec:multimodal-rag}
Multimodal RAG (MRAG) generalises the RAG paradigm to handle multiple data modalities, such as text, images, sensor data and audio \cite{yu2025visragvisionbasedretrievalaugmentedgeneration}.
Instead of restricting $\mathbf{x}$ and $\mathbf{d}$ to be textual, and $\mathcal{R}$, $\mathcal{G}$, $\phi$ and $\psi$ to be functions of input text, they are instead allowed to include or consist of data of any processable modality. 
To facilitate processing a variety of input modalities, $\mathcal{R}$ and $\mathcal{G}$ can be defined as having separate encoders $\phi_m$ and $\psi_m$ and generators $\mathcal{G_m}$ for different modalities $m$.
Alternatively, multimodal models that can encode both text and images into a shared embedding space, such as CLIP \cite{Radford2021clip}, can be used so that
\begin{equation}
    \phi_{\mathrm{text}}(\mathbf{z}) \;\;\mathrm{and}\;\; \phi_{\mathrm{image}}(\mathbf{z}) \;\; \in \;\mathbb{R}^D,
\end{equation}
where $\mathbf{D}$ is the dimension of the shared embedding space. 
Retrieval is then based on cosine similarity or another metric over these multimodal embeddings \cite{Kim2021vilt}.

Formally, let $\mathbf{z}$ represent a multimodal query of an image-text pair, so that 
\begin{align}
    \mathbf{z} &= [\mathbf{x}_\mathrm{text}, \mathbf{x}_\mathrm{image}], \\
    \mathbf{d}_i &= [\mathbf{d}_{i,\mathrm{text}}, \mathbf{d}_{i,\mathrm{image}}],
\end{align}
where $\mathbf{x}_\mathrm{image}$ could be a an image encoding, and $\mathbf{d}_{i,\mathrm{sensor}}$ represents a text description of the image, e.g. a caption from a social media website.
%
%

One can then define specialised retrieval probabilities with two separate encoders $\phi_{\mathrm{text}}$ and $\phi_{\mathrm{image}}$ to do retrieval over two modalities simultaneously:
\begin{equation}
    P_\mathcal{R}(\mathbf{d}_i \mid \mathbf{z}) \;\propto\; \exp\!\Bigl(\phi_{\mathrm{text}}(\mathbf{x}_\mathrm{text})^\top \,\psi_{\mathrm{text}}(\mathbf{d}_{i,\mathrm{text}}) \;+\; \phi_{\mathrm{image}}(\mathbf{x}_\mathrm{image})^\top \,\psi_{\mathrm{image}}(\mathbf{d}_{i,\mathrm{image}})\Bigr).
    \label{eq:MM_retriever}
\end{equation}
If using only one modality to retrieve multimodal data, the other modality is set to zero which facilitates seamless transition between different user inputs.
If instead relying on a multimodal encoder such as CLIP, retrieval probabilities can go from one modality to another as
\begin{equation}
    P_\mathcal{R}(\mathbf{d}_i \mid \mathbf{z}) \;\propto\; \exp\!\Bigl(\phi_{\mathrm{multimodal}}(\mathbf{z})^\top \,\psi_{\mathrm{multimodal}}(\mathbf{d}_{i})\Bigr),
    \label{eq:MM_retriever_joint}
\end{equation}
where $\mathbf{z}$ in this example can be any combination of text and image.
In these examples, multimodal RAG facilitates searching for images and/or text similar to a text query, image query, or a text-image query pair.
Thus, a user could search for image retrieval of the input text query ``a cute dog", and $\mathcal{R}$ would then retrieve images with similar text encodings using 
Equation \ref{eq:MM_retriever}, or directly retrieve images with shared joint encoding space as the input query using 
Equation \ref{eq:MM_retriever_joint}.
A user could also use a multimodal query to e.g., attach a cute dog picture to the query, now aiming to retrieve similar dogs with similar captions, etc.

The generative module $\mathcal{G}$ can likewise be a combination of uni-modal models, e.g. an LLM generating text output based on multimodal retrieval, but ideally $\mathcal{G}$ also operates as a multimodal generator that can take all used modalities into account.
For the image-text example, the joint conditional probability then becomes
\begin{equation}
P(y \mid z) \approx 
\sum_{d \in \mathrm{Top\text{-}k}(p(d \mid r))}
P_\mathcal{R}(d \mid z) \, P_\mathcal{G}\bigl(y \mid z, d\bigr)
\;=\;
\sum_{d \in \mathrm{Top\text{-}k}(p(d \mid r))}
P_\mathcal{R}(d \mid z)
\prod_{i=1}^{N}
P_\mathcal{G}\bigl(y_i \mid z, d, y_{1:i-1}\bigr),
\label{eq:rag-multi-joint}
\end{equation}
where $P_\mathcal{R}(\mathbf{d}_i \mid \mathbf{z})$ is defined with Equation \ref{eq:MM_retriever} or Equation \ref{eq:MM_retriever_joint}, and $P_\mathcal{G}$ is generated with a generative model compatible with the retrieved input modalities.

%


Recent advances in MRAG demonstrate significant progress in integrating vision-language models with retrieval methods across diverse application domains \cite{yu2025visragvisionbasedretrievalaugmentedgeneration}, such as vision-language web-search \cite{zhang2024visionsearchassistantempower} or text-to-cloth fashion retrieval \cite{zhao2024unifashionunifiedvisionlanguagemodel}.
%
MRAG models have also shown generalisation to complex real-world scenarios, such as autonomous driving, by retrieving relevant multimodal examples (e.g., sensor data and dashcam images) to inform in-context learning, thus improving explanation coherence and decision transparency \cite{yuan2024ragdrivergeneralisabledrivingexplanations}.
Similar improvements have been observed in multitask vision-language settings, such as image captioning and visual question answering, where retrieving contextually relevant multimodal content significantly enhances generative performance \cite{rao2024ravenmultitaskretrievalaugmented, ding2024rablipmultimodaladaptiveretrievalaugmented}.

Several challenges exist across MRAG frameworks, including modality imbalance, computational overhead, and the difficulty of aligning heterogeneous data sources within unified embedding spaces \cite{abootorabi2025askmodalitycomprehensivesurvey, zhao2023retrievingmultimodalinformationaugmented}.
Quantitative studies have specifically highlighted the trade-off between retrieval effectiveness and model memorisation, emphasising the importance of carefully calibrated retrieval mechanisms for generalisation and scalability \cite{carragher2025quantifyingmemorizationretrieverperformance}.
Particularly in specialised domains such as medical imaging, the effectiveness of MRAG systems depend on precise modality alignment, domain-specific pre-training, and robust fusion strategies capable of integrating complex, modality-specific semantics \cite{medical_MMRAG}.
%

\subsection{Industrial Applications and Challenges}
\label{sec:industrial-rag}

In high-stakes industrial settings, such as process industry manufacturing or energy production, \emph{condition monitoring} involves collecting real-time or periodic data (vibration, temperature, acoustic signals, etc.) to detect incipient equipment failures \cite{Jardine2006, tiboni2022review}.
Conventional analytics approaches often fail to capture complex correlations or lack transparency in how recommendations are generated \cite{yan2017industrial}. 

By incorporating RAG, engineers could link sensor readings with relevant documents such as:
\begin{itemize}
    \item \textbf{Technical machine information documents:} Retrieve sections describing fault modes, machine component characteristics, and operational data.
    \item \textbf{Historical fault description annotations:} Retrieve fault descriptions associated with sensor data.
    \item \textbf{Maintenance records:} Provide best practices or official procedures from the maintenance history.
\end{itemize}
Such retrieval can provide \emph{traceable} knowledge, while the generative model can produce interpretable recommendations or summarise possible fault causes.
This traceability could help operators or inspectors verify the source documents, bridging the gap between AI inferences and established engineering knowledge \cite{zhang2025llm, Alsaif2024Electronics, KG_LLM_FD, chen2025faultgptindustrialfaultdiagnosis}.

Key challenges in deploying industrial RAG include:
\begin{enumerate}
    \item \textbf{Heterogeneous Data:} Sensor measurements, textual field notes, and schematics must be processed under a unifying retrieval model. Multimodal RAG thus becomes essential.
    \item \textbf{Domain Adaptation:} Industrial taxonomies and naming conventions are characterised by \textit{technical language} \cite{technical_language_processing, lowenmark2022processing}, which requires specialised embeddings and knowledge integration \cite{long2023adaptcontextsretrievalaugmenteddomain}.
    \item \textbf{Real-time Inference:} Large-scale systems monitoring thousands of machines simultaneously demand efficient retrieval and generation pipelines.
\end{enumerate}

Extending MRAG to an industrial context, we might consider combining sensor readings, diagnostic codes, and textual descriptions within the same retrieval mechanism.
For instance, let $\mathbf{z}$ and $\mathbf{d}$ instead be
\begin{align}
    \mathbf{z} &= [\mathbf{x}_\mathrm{text}, \mathbf{x}_\mathrm{sensor}], \\
    \mathbf{d}_i &= [\mathbf{d}_{i,\mathrm{text}}, \mathbf{d}_{i,\mathrm{sensor}}],
\end{align}
where $\mathbf{x}_\mathrm{sensor}$ could be a time-series embedding (e.g., vibration signals), and $\mathbf{d}_{i,\mathrm{sensor}}$ might represent reference patterns or historical fault data. 
With text annotated signals and an appropriate signal encoder, equations \ref{eq:MM_retriever} and \ref{eq:rag-multi-joint} can then be adapted to facilitate text-based sensor retrieval, or sensor-based text-retrieval.
The latter in particular could facilitate direct fault diagnosis by retrieving fault descriptions based on sensor characteristics, thus mimicking human diagnosis by generating a response conditioned on the most similar past analysis.
Multimodal industry models \cite{lowenmark2021technical} could also extend this further by facilitating joint multimodal retrieval per Equation \ref{eq:MM_retriever_joint}, and unlock direct multimodal generation in Equation \ref{eq:rag-multi-joint}.
Such models could directly retrieve signals that are the most similar to a text query, or vice versa, and generate fault descriptions based both on similar past descriptions and an internal multimodal mapping from an input signal to the most appropriate output text.

Recent research highlights significant potential in integrating LLMs with specialised retrieval techniques for advanced CM.
For instance, conversational frameworks leveraging LLMs can retrieve real-time industry data \cite{chatcnc}.
These advancements align closely with the ongoing industry shift towards more user-centric, contextually responsive diagnostic methodologies within CM \cite{HEREDIAALVARO2025103007}.

Industry data is often formatted as or akin to graphs, which can be leveraged by integrating graph properties into the RAG approach \cite{peng2024graphretrievalaugmentedgenerationsurvey, li2025simpleeffectiverolesgraphs}, to for instance incorporate medical knowledge graphs to augment question-answering models in this domain \cite{cabello2024megmedicalknowledgeaugmentedlarge}.

\label{sec:multimodal_fault_detection}

One example of an industrial application of MRAG is \emph{Multimodal Large Language Model-Based Fault Detection and Diagnosis in Context of Industry 4.0} \cite{Alsaif2024Electronics}, where the authors propose a system that leverages multimodal large language models to support fault detection and diagnosis in industrial environments.
According to the paper, the method integrates data from multiple modalities—such as sensor signals, textual maintenance logs, and visual data—by encoding each modality into a shared embedding space.
A retrieval module then selects relevant external evidence based on the encoded query, which may combine sensor data and text.
The integrated information is used to condition a large language model that generates diagnostic outputs and fault explanations.
Experimental evaluations demonstrate improvements in diagnostic accuracy and interpretability compared to baseline approaches, while also highlighting challenges related to data heterogeneity, scalability of multimodal retrieval, and domain-specific adaptation in Industry 4.0 settings.
However, the functionality is primarily limited to a question-answering agent for diagnostic information, and does not analyse or diagnose signals, instead operating with data from an Online Support Forum and fault codes, descriptions, and corrective actions from multiple industrial control system manufacturers, not raw sensor data, when doing inference.

\bigskip

In conclusion, LLM agent and MRAG approaches have to potential to be augment CM workflows if aptly applied to annotated industry data.
Such approaches require the development of a semi-structured multimodal vector store from unstructured CM data, data- and task-specific retrieval functions, and generative models implemented in accordance with industry needs.

\section{Multimodal Condition Monitoring Data and Vector Store}
\label{paper_E:vector_store}

To meet the functionality demands introduced in Section \ref{paperE:introduction} and Table \ref{tab:knowledge_gap}, with regard to the background laid out in Section \ref{sec:related_work}, sensor data must be fused with machine hierarchy data and associated text description annotations in a vector store compatible with multimodal data retrieval and generation.
Table \ref{tab:data_needs} describes data, retrieval, and generation requirements mapped to these industry demands, separating these requirements into \textit{documents} -- as commonly used in RAG approaches -- and \textit{recordings} connected to these documents, an unexplored addition to RAG and AI agent approaches.
To facilitate retrieval and generation, CM data must thus be processed to set up a vector store that accommodates multidimensional multimodal data, where each document consists of multiple recordings in a way that facilitates retrieval and generation both at the document-level and at the recording level.

\begin{figure}
    \centering
    \includegraphics[width=1.0\linewidth]{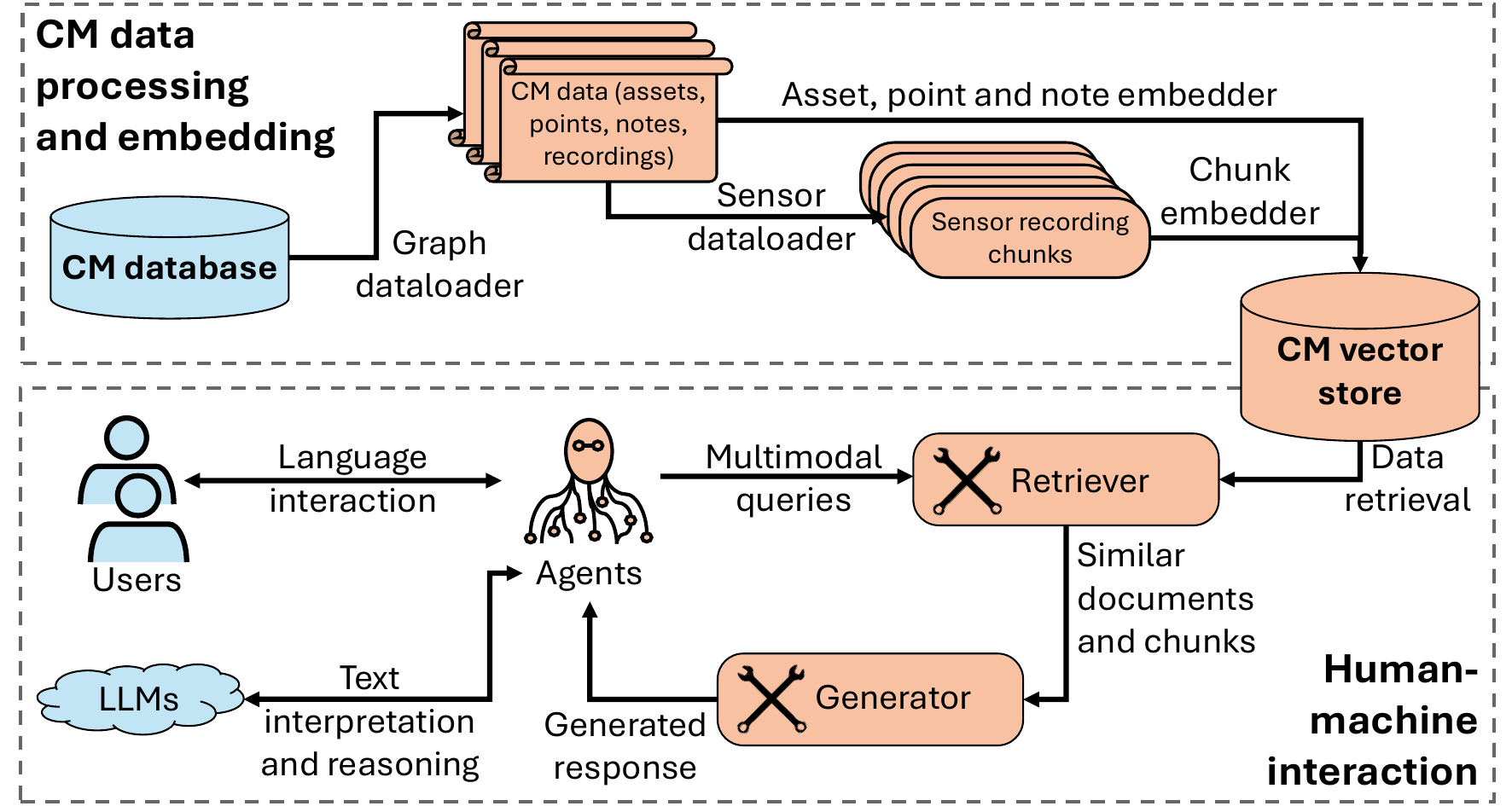}
    \caption{Information flow from CM database to CM vector store and from vector store to generated response.}
    \label{fig:RAG_pipeline}
\end{figure}

Figure \ref{fig:RAG_pipeline} outlines a RAG workflow, where blue coloured elements depict existing elements, i.e. (annotated) CM databases, potential users (analysts), and LLMs, and orange elements depict development requirements.
To facilitate Multimodal industrial database RAG (MindRAG), the CM database must be processed into CM documents and recording chunks, which form the basis of the CM vector store.
The vector store is then accessed with custom retriever functions, which is described in Section \ref{sec:retrieval}.
The retrieved data is then used by custom agents to generate a response, described in Section \ref{sec:agents}.
In the figure, the generator, agents, and LLMs, are shown separately, though in practice generation is done by the agents through specific tools and prompts using an LLM.

This section describes how machine sensor, hierarchy graph, and text, data is processed into a multimodal vector store, starting with an explanation of the physical background of the CM data, followed by a step-by-step description of how to transform CM data to a CM vector store.
Machine information, such as component name and placement, and sensor placement, structured to mirror the machine hierarchy tree, is here referred to as hierarchy graph data, which is textual data like annotations, but completely technical with only abbreviations and jargon.

%

%

\begin{table}[t]
\caption{Data, retrieval, and generation functions required to meet various industry demands.}
\label{tab:data_needs}
\centering
\begin{tabular}{p{5cm}p{10cm}}
    \hline
    \hline
    \textbf{Analysis Category} & \textbf{Data, retrieval, and generation required for AI assistance} \\
    \hline
    \hline
    & \\
    Historic fault insight & \multirow{-1.5}{10cm}{Retrieve and analyse historic fault description annotations and maintenance work order \textit{documents} connected to the machine hierarchy graph, based on annotation and graph retrieval distance metrics, and generate a response grounded in the most relevant annotations and hierarchy graph components.}\\
    & \\
    \cdashline{1-1}
    & \\
    Component-specific fault insight & \\
& \\
    \hline
    & \\
    Fault detection and classification & \multirow{7}{10cm}{Retrieve and analyse sensor data \textit{recordings} connected to the machine hierarchy graph \textit{documents}, either with separate distance metrics or in a joint embedding space. Generate a response conditioned on hierarchy graph and text data associated with relevant recordings.} \\ 
    & \\
    \cdashline{1-1}
    & \\
    Fault severity estimation  &  \\
    & \\
    \cdashline{1-1}
    & \\
    Maintenance decision making &  \\
    & \\
    \hline
    \hline
\end{tabular}
\end{table}

\subsection{Machine Hierarchy Graph}
\label{sec:PM1}
The machine, at its highest level, is conceptualised as a set of sections, each filling one role in the production of kraftliner paper.
First, the paper pulp enters the forming section winders, split into upper and lower winders.
Each winder consists of rollers of a different size, but with similar working conditions of humid but not hot environments, as well as motors and gear boxes.
The pulp then enters the press sections, containing rollers with similar working conditions as the winders.
The bulk of the paper machine consists of drying sections, characterised by hot and decreasingly humid working environments as the pulp gets dryer further into the machine.
The drying sections consist of small articulated drying rollers and large drying cylinders, which are the most common components in the machine.
Then, the paper enters through the sizing rollers into the calender section, consisting of rollers operating in a dryer environment than earlier in the machine.
Finally, the paper enters the reel section to be wound up on spools for further processing or shipping.

Each section consists of groups of similar components at similar placements, e.g., ``drying cylinders 1A FS", which consist of individual components, e.g., seven drying cylinders (on the free side, i.e. the opposite end of the rollers from where the electric motor driving them is mounted) for the aforementioned group.
The rollers are held by bearings at the extremes, where the drive end is coupled to a motor which enables the movement of the roller.
Each component, denoted as an \textit{asset} in the vector store, has at least one sensor, typically an accelerometer, attached.
The sensors are mounted either on a vertical, horizontal, or axial, position, which in the case of rollers is on the house of the bearings, where the sensors measure vibrations originating from the bearings.
The sensors are connected to a collection box, which processes the acceleration measurements and saves them as recordings of acceleration, velocity, and different enveloped signals on their time waveform and spectrum.

%
Data is stored for each type of filter, as well as for different sampling rates or duration.
Each data stream is called a \textit{point} in the vector store, and each stored signal is called a \textit{recording}.
Recordings are dictionaries representing signals instances, defined by a spectrum array, a time series array, and various metadata, such as shaft speed information, parent point information, time and reason for storing measurement, etc.
In total there are 33 keys, though some are empty and others are not sufficiently relevant to the data analysis to warrant including in the vector store.

\begin{figure}[b]
    \centering
    \includegraphics[width=0.95\linewidth]{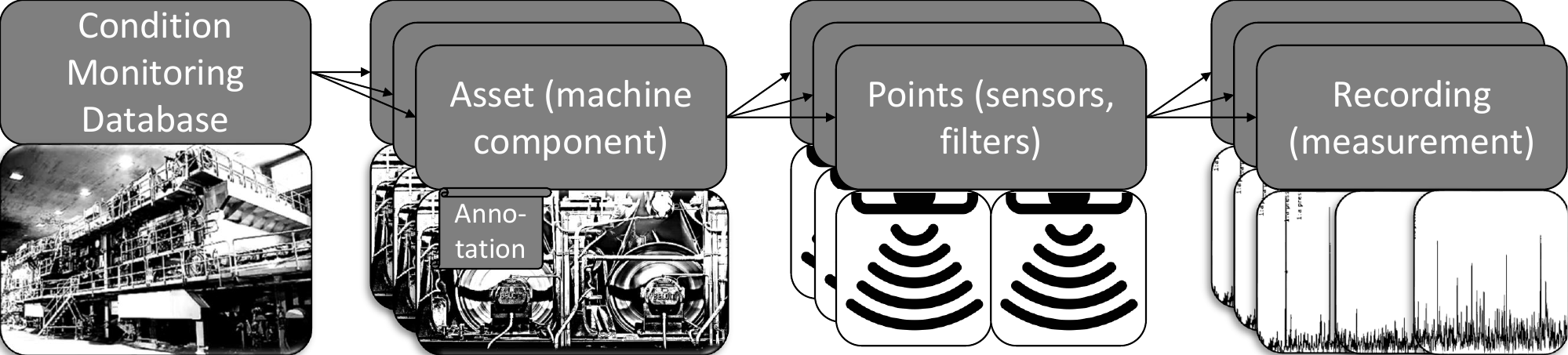}
    \caption{Hierarchy schema illustrating the graph-like structure of data in a process industry machine.}
    \label{fig:schema_black}
\end{figure}

Assets can also be associated with one or more \textit{annotations}, which are technical language descriptions of fault properties, maintenance work orders, maintenance actions, and other information relevant to the process.
Annotations are primarily defined by their content and the date, though other metadata such as annotation author or title are also available.
Points and annotations are thus linked to assets (components), which form component groups that make up the sections of the machine.
Sensor data is linked to the points in the form of recordings, forming a tree structure that is illustrated in Figure \ref{fig:schema_black}

\subsection{Setting up a Condition Monitoring Vector Store}

The condition monitoring database is defined as a 150GB SQL backup file extracted from the SKF Observer computerised CM system, connected to the paper machine hierarchy described in Section \ref{sec:PM1}.
Data is accessed through JSON queries generated with the Phoenix API.
Since the amount of data is large, and almost all data is unannotated and unlabelled, guided data extraction is crucial for effective and efficient research and testing.
Therefore, data extraction and processing is split into multiple steps, where each part of the data serves one function in the final agent implementation.
These steps are:
\begin{enumerate}
    \item extracting graph hierarchy data (assets, annotations, points),
    \item extracting sensor data (recordings),
    \item filtering recordings,
    \item processing filtered recordings,
    \item generating features from processed recordings,
    \item creating vector store.
\end{enumerate}

\subsection{Extracting and Processing Graph Hierarchy Data}
\label{sec:hierarchy}
The first step in setting up the CM vector store is to extract the graph hierarchy data, as described in Section \ref{sec:PM1}.
A hierarchy list is created from the hierarchy data as a flattened version of the tree of components where each element in the list corresponds to an asset.
Notes are then extracted and mapped to the corresponding asset.
The annotations in the dataset are stored for up to five years.
Since not every component has warranted human analysis in the last five years, while some have malfunctioned multiple times, an asset can have anywhere from zero to over ten associated annotations.

Points are extracted from the asset list, and stored alongside the asset and associated notes.
An asset normally has between two and four points, where each point defines one signal type from the same sensor, typically velocity (raw spectra) and one or more envelope filtered signals.
Some assets have upwards to or above ten associated points with a mix of signal types from multiple sensors.

Fault descriptions and maintenance actions described in the annotations can sometimes be relevant to every point in an asset, such as a sensor fault at an asset with one sensor, or a bearing fault that is visible both in the velocity and envelope signals.
However, some bearing faults are visible only in a specific signal type, which means that some points will be associated with fault descriptions that are not representative of the features in the point.
Likewise, a sensor fault in an asset with multiple sensors will naturally only exhibit those features on the affected points, while all other points will be associated with an erroneous fault description.
Furthermore, sensors on for instance the free side and drive side of a roller can both exhibit bearing fault features due to the vibration travelling through the component, in which case both sensors can reflect a fault physically present only in one subpart of the asset, but digitally present in some signal types on multiple points.

\subsection{Extracting CM Signal Data}
Sensor data is connected to the points in the database, and can be extracted within a chosen time span with the point ID.
Since the amount of data is large, but essentially all unlabelled and unannotated, language guided extraction is key to efficient data-driven fault diagnosis.
For most research tasks, annotated data is more valuable than unannotated data, and the simplest form of language guided extraction is to extract data from before and/or after annotation dates.
However, the content of the annotations can also be used to filter what data is extracted for more efficient computation.
This can be achieved in three ways:
\begin{itemize}
    \item Rule-based inference -- cheap and efficient but low resolution and prone to errors.
    \item LLM-based inference -- computationally more expensive but far more robust than rule-based inference. Requires assistance with interpreting technical terms based on tacit knowledge.
    \item Human-based inference -- reliable if done by experts, but costly and time consuming.
\end{itemize}

Rule based inference is the simplest form of language guidance; a researcher or engineer can analyse the corpus of annotations and define rules to classify note contents into different fault classes.
For example. all annotations featuring the word ``replaced" can be classified as having fault-free data after, and fault features before, the annotation date.
Annotations featuring ``cable" or ``sensor", but not ``replaced", are then instead mapped to corresponding fault features both before and after the annotation date, etc.
However, rule-based analysis of language is exceptionally difficult to scale without forcing users to conform to strict taxonomies, and the rule-based system must be able to account for negations, spelling mistakes, nuanced descriptions, etc.

Therefore, an LLM can be used as an intermediate layer between user requests and data extraction.
An LLM with proper assistance can analyse note contents and suggest appropriate intervals for data extraction.
For example, one note from the database reads (translated to English) ``BPFO on DS bearing visible on FS. The sensor is broken on 64372 DS. Replace the sensor".
Interpreting this annotation with a rule-based system that correctly identifies that the free side sensor shows features that indicate a BPFO in the drive side, but that the drive side sensor is broken and should be replaced, is beyond challenging, but an LLM augmented with tools to interpret the technical language and translate it to extraction instructions is capable of such a task with much better results.
However, using an LLM requires either the installation of an in-house model on a powerful computer or in a data centre, or calls to external providers of LLMs, which can be either difficult with a start-up cost, or expensive over time with possible data leaks of vulnerable data. 

Human experts can analyse all annotations and create extraction rules, which would result in data extraction closest to the ground truth without manually inspecting each recording, but also take up valuable expert time and scale poorly whenever different data, either with different time spans, or from another dataset, is analysed.

In this study, all three methods are used for different volumes of data.
First, all annotated recordings are extracted and used in all tests, with time differences up to 50 days before and 20 days after annotation dates.
The annotations are classified both by a rule-based system for initial analysis, and then by an LLM in later stages of the model, such as for fault diagnosis inference.
Finally, the results are verified both by a human analysis and by an LLM evaluation agent.

\subsection{Processing CM Signal Data}

The extracted recording data is formatted as nested dicts, where for instance each recording has 33 keys, some of which are elements that are dicts of dicts.
To make data storage and processing more efficient, the recordings need to be restructured into training data, which is done in two steps; first, the data is filtered based on recording keys and metadata conditions, such as sensor type, machine speed, or recording date.
%
The filtered data is then converted into training data, structured so that every note with the associated asset and points is a unique element in the list (the same asset will thus appear multiple times with different notes for unique note-asset pairs).
Each point connected to a note contains a list of recordings.
The recording is structured so that of the 33 keys, four data types are contained -- time series, spectra, trend levels, and sensor bias --and three types of metadata -- machine shaft speeds, recording dates, and time delta between recording date and note date.

The time series and spectra are the main vibration measurements, the spectra being the fast fourier-transformed time series, though for envelope points, the time series is also passed through the envelope filter, making it difficult to interpret.
These measurements are recorded normally four times per day, 6.4 seconds per recording with 8192 elements in the time series array and 3200 in the spectrum distributed up to 500Hz.
Some recordings deviate from this and have higher or lower sampling rates or durations, which is processed in the filtering step.
The trend levels are the average vibration values streamed constantly from the sensors to the database, and are thus a low-resolution approximation of the state of the component, but with frequent data points, hundreds per day.
The sensor bias is the bias voltage over the sensor, and is streamed along the levels as frequent data points.

When transforming the CM data to a training dataset, the goal is to produce sets of inputs and target outputs (psuedo-labels).
Since the CM database is unlabelled, the annotations and the time distance to them act as proxy labels.
As previously described, annotations are stored at the asset level and are based on evolution of features in all asset points over many recordings.
Annotations are also based on tacit knowledge such as which components are more vulnerable to faults, which components are easier to replace, when the next maintenance stop is, feedback from maintenance engineers, etc.
Therefore, the closest approximation to the real input-output system that produced the annotation would be to map all recordings from an asset for a considerable time leading up to an annotation to one annotation, and task a system with reproducing the annotation based on all the input data.
However, this would yield at most a few hundred training samples, far too few for practical data science.
Due to the varying number of points with varying types of signals of varying shapes, it is also difficult to design an end-to-end model with sufficiently flexible input space based on so little data.

Input slices can instead be defined based on subsets of the total information that warranted the annotation.
At the opposite end of the input spectrum, this can be modelled as every recording being labelled by the closest annotation and the difference in time.
However, besides the variance in features in the point domain, individual recordings are even more inconsistent in what features they indicate.
For example, cable and sensor fault features can appear, then disappear, then reappear seemingly at random due to disturbances in these components, and in between the noisy signals the recordings might be perfectly normal.
Bearing faults also tend to develop non-linearly, so initial rising feature levels might warrant an annotation describing fault presence, but then the features nearly vanish.
Both cases result in many recordings being mapped to annotations that do not correctly describe their properties, despite the process for mapping being done correctly.

Consequently, the training data step features a flexible approach, where an input sample is defined based on a time stamp, which features a list of the $N$ most recent spectra arrays, $M$ most recent time series arrays, $K$ most recent trend level data points, and $L$ most recent sensor bias data points, where all hyperparameters are freely chosen.
Processing is done at the point level as different points can have slightly different time steps, which means that regardless of how many points an asset has, each training input will have the same shape.
Points can be remapped to their asset root though, for more thorough inference.

This training data can be used to directly train IFD models, or as input to tools handled by an AI LLM agent.
Two signal-based use cases are presented in this report: using CM training data to compute and compare features based on natural language inputs, and creating a multimodal vector store for Multimodal industrial data RAG (MindRAG). 

\subsection{Multimodal Industrial CM Data Vector Store}

%

%
\begin{table}[t]
    \caption{CM vector store keys with associated data and distance metrics.}
    \label{tab:vector_store}
    \centering
    \begin{center}
    \begin{tabular}{  c  c  c }
        \makecell{\textbf{Document} \\ \textbf{key}} & \textbf{Data} & \textbf{Description} \\
        \hline
        Asset &  \makecell[c]{Path}  & \makecell[l]{The graph path from the root machine through the \\ machine groups to the component.}  \\
        \hline
        \multirow{2}{*}{Point} & \makecell[c]{Name} & \makecell[l]{Component name with sensor placement and/or \\ filter name.}  \\ \cdashline{2-3}
        & \makecell[c]{Type} & \makecell[l]{Sensor filter type class.} \\
        \hline
        \multirow{5}{*}{Note} & \makecell[c]{Note\\content} & \makecell[l]{Annotation describing component properties, typically fault \\ description, maintenance order, or maintenance follow up.}  \\ \cdashline{2-3}
        & \makecell[c]{Note\\date} & \makecell[l]{The date and time of the annotation.} \\ \cdashline{2-3}
        & \makecell[c]{Note\\embedding} & \makecell[l]{LLM vector representation of note content.} \\
        \hline
        \multirow{12}{*}{Recordings} & \makecell{Signal\\embedding(s)}  & \makecell[l]{The order vector representation of the signal properties.
        } \\ \cdashline{2-3}
         & Spectra & \makecell[l]{The fft spectrum of the time series sensor measurement, \\ 3200 elements over 500 Hz.}  \\ \cdashline{2-3}
         & \makecell{Time\\series} & \makecell[l]{The time series sensor data, 8192 measurements over 6.4 \\ seconds, continually sampled but generally stored four \\ times per day.} \\ \cdashline{2-3}
         & \makecell{Shaft\\speed} & \makecell[l]{The average shaft speed during the recording} \\ \cdashline{2-3}
         & \makecell{Trend\\levels} & \makecell[l]{Historic trend levels from the last 10 days. Stored with \\ much higher frequency than the time series.}\\ \cdashline{2-3}
         & Biases & \makecell[l]{The bias over the sensor from the last 10 days. Stored \\ alongside the trend values.}\\ \cdashline{2-3}
         & \makecell{Recording\\date} & \makecell[l]{The date of the recording.} \\ \cdashline{2-3}
         & \makecell{Recording\\timedelta} & \makecell[l]{The difference in days between the note date and the \\ recording date.} \\ \cdashline{2-3}
 
    \end{tabular}
    \end{center}

\end{table}

The vector store is structured around the flattened hierarchy list from Section \ref{sec:hierarchy}, but flattened so that one document corresponds to one point and one annotation, thus consisting of document chunks identified by a document ID defined as (point ID, annotation ID).
The document chunks consist of an asset, an annotation, a point, and a set of recording chunks from the point extracted from a time delta around the annotation note date, as shown in Table \ref{tab:vector_store}.
From the asset, the asset path is included, which describes the path through the hierarchy from the machine to the component, as described in \ref{sec:hierarchy}.
From the annotation, the note content -- the main text of the annotation -- and the note date -- when the annotation was written -- are included.
From the point, the point name -- the name of the component -- and point type -- the filter applied to the data -- are included

The recording chunks are defined per Table \ref{tab:vector_store}.
%
%
In this study, the signal embeddings are defined as the order transformed spectra, but any embedding of signal properties could be used.
The spectra and time series are sensor data as described in the sections above, with the shaft speed describing the average RPM of the machine at the time of measurement.
The recording date is the readingTimeUTC string describing the exact second the measurement ended, while the timedelta is the difference in days between the note and recording dates.
The trend levels and biases are sampled at much higher frequencies, and represent not just the sampled value at the recording date, but also historic values as a hyperparameter.

\section{Retrieval: Distance Metrics}
\label{sec:retrieval}
In this section, the score between one \textit{source} and one \textit{target} document or recording chunk are discussed.
The source document or chunk is the system representation of the input query, transformed from the natural language user query to a retrieval compatible query.
For example, if a user asks ``How many breaing flauts are there in ven sensors on driyng grp 3?", the agent would first reinterpret the question to \textbf{asset path: drying group 3 (TG3)}, \textbf{point name: env}, \textbf{note content: (list of bearing fault related terms in Swedish)},  \textbf{note embedding: (language query integrating bearing fault related terms in Swedish)}.
The target documents are the elements of the vector store, as described in Table \ref{tab:vector_store}.
Each document key data can then be compared with the target documents in the vector store, thus obtaining a distance or similarity score for each key data, as shown in Table \ref{tab:vector_store_keys}.
%
%
%
%
%
%
%
Retrieval from the vector store is then achieved by taking the $k$ closest documents or chunks to the source data, or taking all documents or chunks with a score above a threshold value.

\subsection{Retrieval Definition}
We define the retrieval function between an input query $z$, consisting of one or more modalities of data compatible with the documents $d$ in a CM vector store, and $d_i$, a document from the CM vector store, as
\begin{equation}
    P_\mathcal{R}(\mathbf{d}_i \mid \mathbf{z}) \;=\; 
    \prod_{n=1}^{N}F_n(\mathbf{z}_n, \mathbf{d}_{i,n}) * 
    \sum_{m=1}^{M}w_m
    \Bigl(
    \phi_{m}(\mathbf{z}_m, \mathbf{d}_{i,m})
    \Bigr),
    \label{eq:CM_retriever}
\end{equation}
where $N$ is the number of filters and $M$ is the number of distance metrics in the vector store; $F_n$ is a filter function based on an input criterion defined in $z_n$ that acts on the data in $d_{i,n}$; $w_m$ is a weight assigned to the metric (if $w_m=0$ the score is not further computed for efficiency); and $\phi_{m}$ is a distance function that acts on the query $z_m$ and the vector store element $d_{i,m}$ associated with the metric $m$.

Equation \ref{eq:CM_retriever} is true for both document-level retrieval, such as knowledge retrieval based on past annotations, and signal-level retrieval, such as retrieving chunks with similar signal embeddings and shaft speeds.
Retrieval can also be done at both levels at once, such as retrieving only signals from certain drying groups with certain notes attached with signal dates within five days of the note dates.
Naturally, retrieving at the document-level will also retrieve all associated chunks at the recording level, should the user request them, and retrieving similar chunks also retrieves the associated document for each chunk.
In practice, for computational efficiency, document-only retrieval is done in an identical copy of the vector store that lacks the recording chunks, as they constitute the vast majority of all data stored, which speeds up compute for these queries.

\subsection{Filters}
The filter function $F_n$ scores either $0$ or $1$ depending on whether it is True or False.
$F_n$ contains operational logic, such as AND and OR conditions, so that for instance vector store data can be limited to drying groups 2 OR 3 in the asset path to freely interact with the machine graph, or ``drying cylinder" AND ``axial position" in the point name to freely filter for certain signal types for analysis.
If $z_n$ is empty for a certain value of $n$ $F_n$ automatically return a unity function.
$F_n$ acting upon a score in the complete retrieval chain can be conceptualised as
\begin{equation}
\text{Score} = 
\begin{cases} 
\text{Score}, & \text{if } \text{Condition is True},\\
0, & \text{else}.
\end{cases}
\label{eq:filter}
\end{equation}

In this formulation, all filters conditions must be fulfilled for $P_\mathcal{R}$ to return a score over $0$, which is the most common use case for filtering the CM vector store as part of a query.
An AND operator is thus implicitly assumed to be working between filters.
However, in the full implementation, the filter functions act within a logical tree, and can be defined to act as OR or AND functions based on user input, which facilitates complex albeit less common queries such as restricting documents to \textit{either} come from bearing A in group N or bearing B is group M, etc.
%

The section/group filter allows the user or agent to specify which parts of the machine to include or exclude for the analysis.
This can be useful to search for similar historic cases in the same section, or to detect recording chunks from other sections with similar properties by excluding the naturally more similar chunks from the source section.
Type filters enable the user or agent to specify what type of data to include in the analysis, e.g. only envelope filters of a certain type, or only raw velocity data.
Note content filters are keyword-based searches in the annotation space, augmented by TLP on the agent side.
This facilitates fast and robust text-based retrieval by including only notes with certain keywords, e.g. [``kabel", ``giv", ``sens"] for cable or sensor faults, which can be used for further analysis by the agent or the user to for instance count which sections have the most faults of a certain type, or which faults are the most common is a specific component, etc.
A drawback is that spelling mistakes, new vocabularies, and design errors, can all result unintentional omission of notes, which is addressed by direct analysis by the agent.
Finally, note date span filters allow the user or agent to decide in what time frame to analyse the hierarchy data by selecting time spans to include or exclude documents based on the date the note was written.
This is highly relevant to time-bound hierarchy queries such as ``which fault was the most common in drying group three the last three months".

For the recording chunks, filters can be used to include or exclude recordings of certain speed, to for instance investigate how faults and working operations correlate.
The recording date and recording timedelta can both be filtered in the same way as the note date span filter.
This is extremely useful to investigate what data is most valuable for chunk retrieval, to test for instance whether data from before or after the annotation date correlates more with the note content.

\begin{table}
    \centering
    \caption{CM vector store keys with associated data and distance metrics.}
    \label{tab:vector_store_keys}
    \begin{center}
    \begin{tabular}{ | c | c | c |}
        \hline
        \thead{\textbf{Document key}} & \thead{\textbf{Data}} & \thead{\textbf{Distance metrics}} \\
        \hline
        Asset &  \makecell{Path}  & \makecell{Bleu score \\ Hierarchy distance \\ Section/Group filter}  \\
        \hline
        \multirow{2}{*}{Point} & \makecell{Name} & \makecell{Bleu score \\ Component filter}  \\ \cdashline{2-3}
        & \makecell{Type} & \makecell{Type filter} \\
        \hline
        \multirow{3}{*}{Note} & \makecell{Note contents} & \makecell{Embedding score \\ Content filter}  \\ \cdashline{2-3}
        & \makecell{Note dates} & \makecell{Date similarity score \\ Date span filters} \\
        \hline
        \multirow{12}{*}{Recording chunks} & Signal embedding  & \makecell{vector similarity \\ Signal features scores} \\ \cdashline{2-3}
         & Spectra & \makecell{vector similarity\\ Signal features scores}  \\ \cdashline{2-3}
         & Time series & \makecell{vector similarity\\ Signal features scores} \\ \cdashline{2-3}
         & Shaft speed & \makecell{Speed similarity score\\ Speed span filters} \\ \cdashline{2-3}
         & Trend levels & \makecell{Variance score}\\ \cdashline{2-3}
         & Biases & \makecell{Variance score}\\ \cdashline{2-3}
         & Recording date & \makecell{Date span filters} \\ \cdashline{2-3}
         & Recording timedelta & \makecell{Date similarity score \\ Timedelta span filters} \\ \cdashline{2-3}
        \hline
    \end{tabular}
    \end{center}

\end{table}

\subsection{Scores}
The scores are the distance metrics that facilitate retrieval.
A high score represents a stronger correlation between the source and target documents/chunks.

\subsubsection{Hierarchy distance}
\label{sec:hierarchy-distance}

The hierarchy list is obtained by splitting the path in the CM vector store so that each hierarchy level is one element in the list.
The hierarchy distance is then computed to find the smallest total number of right-end truncations to match a source hierarchy list and a target hierarchy list, i.e. how many steps must be taken in the machine hierarchy to go from the source to the target.

Let two lists (or sequences) of length $n$ and $m$ be defined as
\[
A = (a_1,a_2,\dots,a_n),
\quad
B = (b_1,b_2,\dots,b_m).
\]
In the example, $A$ and $B$ are the hierarchy paths.

The \emph{distance} $\mathrm{Dist}(A,B)$ is then defined as the smallest total number of right-end truncations (steps) needed so that the remaining left prefixes of $A$ and $B$ match exactly. Formally, 
\begin{equation}
    \mathrm{Dist}(A,B) \;=\;
\min \Bigl\{ (i + j) \,\Big|\,
0 \,\le\, i \,\le\, n,\; 
0 \,\le\, j \,\le\, m,\; 
(a_1,\dots,a_{n-i}) 
\,=\, (b_1,\dots,b_{m-j})
\Bigr\}.
\end{equation}

Here, $i$ is how many elements we remove from the right end of $A$, while $j$ is how many elements we remove from the right end of $B$. Once the truncated lists match exactly (i.e., they have the same length and the same sequence of elements), the code records $i+j$ as the distance and terminates.

\paragraph{Relation to Largest Common Prefix}
Let $k$ be the maximum length of a prefix that $A$ and $B$ can share when truncated from the right. In other words, there exist $i,j$ such that $n-i = m-j = k$ and $(a_1,\dots,a_k) = (b_1,\dots,b_k)$. Then
\begin{equation}
    \mathrm{Dist}(A,B) \;=\; i + j \;=\; (n-k) + (m-k) \;=\; n + m - 2k.
\end{equation}

Thus, minimizing the sum $i + j$ is equivalent to \emph{maximizing} the size of the matching prefix, $k$.


\subsubsection{BLEU Score}
\label{sec:bleu}
BLEU (Bilingual Evaluation Understudy) score~\cite{papineni2002bleu} is used to measure the word similarity between assets.
The BLEU score is widely used in NLP to compare a candidate sentence to one or more reference sentences using:
\begin{itemize}
    \item \textbf{Modified $n$-gram precision}, accounting for the overlap of $n$-grams between candidate and references, with counts clipped by the reference frequencies.
    \item \textbf{Brevity Penalty (BP)}, penalizing candidates that are too short compared to the reference(s).
\end{itemize}

The BLEU score for up to $N$-grams (often $N=4$) is given by
\begin{equation}
\text{BLEU} = \text{BP} \times \exp\Bigl(\frac{1}{N}\sum_{n=1}^{N}\ln p_n\Bigr),
\end{equation}
where
\begin{equation}
\text{BP} = 
\begin{cases} 
1, & \text{if } c > r,\\
\exp\Bigl(1 - \frac{r}{c}\Bigr), & \text{if } c \le r,
\end{cases}
\end{equation}
and $p_n$ is the modified precision for $n$-gram length $n$, and let $c$ and $r$ are the lengths (in tokens) of the candidate and reference, respectively.

In practice, it is common to compute BLEU at the corpus level by summing $n$-gram counts across all sentences, then applying the brevity penalty based on the total candidate-reference length ratio. For short texts or single-sentence evaluations, smoothing strategies are often employed to mitigate zero count.

BLEU score is employed at the asset and point level to compare how similar the source and target assets or points are.
A high asset BLEU score indicates that the document is either in the same section or group, e.g. a roller next to a drying cylinder, or in a similar type of machine in another section, e.g. two drying cylinders in different sections.
The point BLEU score indicates whether we should expect the data to be similar; a high point BLEU score likely implies that the source and target points are similar component, placed similarly in their sections, though not necessarily in the same sections.
For instance, the point ``TC 2 FS" is at the very start of the first drying group, while the point ``TC 74 FS" is at the end of the fourth drying group.
Though their working conditions will differ, they are similar components will similar roles, and are likely to yield somewhat similar signal characteristics, though with different noise properties.
Combining this information with the asset BLEU score or the asset distance score facilitates a deeper understanding into how robust signal-level retrieval is, based on hierarchy knowledge.

Both BLEU scores are computed as document-to-document comparisons, using \textit{sentence\_bleu} from the \textit{nltk} package, with weights set as [1, 0, 0, 0].
Thus, the BLEU score distance metric used compares only unigrams and does not take word order into account.
This works well due to the hierarchical nature of the hierarchy data and compositional nature of Swedish where words such as ``drying group" are instead written as ``torkgrupp", which results in no word overlap at different levels of the hierarchy.
Computing BLEU distance metrics in an English CM vector store would likely require a slightly different approach where at least bigrams are taken into account, as ``drying group" and ``drying cylinder" would have a 50\% unigram overlap, which does not reflect the intended goal of this distance metric.
The BLEU score metric could also be upgraded by generating documents at the asset level, where point-scores would be compared against a small corpus of points in the comparison asset.

\subsubsection{Annotation Embedding Score}
The annotation embedding score can be understood as ``standard" RAG retrieval, computed as the dot product between the embeddings of a query or source annotation and a target annotation.
The embeddings are computed with a language model, ``text-embedding-3-small" from OpenAI in this study.
The input space -- the input query and the annotations -- is preprocessed with technical language substitution \cite{lowenmark2022processing} to augment the embedding space and improve the retrieval step.

\subsubsection{Date Similarity Scores}
Date similarity scores are computed to either penalise or promote chunk scores depending on how far away from the source chunk they are.
It is computed as 
\begin{equation}
    \text{Date score} = \frac{N}{t_d+N},
\end{equation}
where $t_d$ is the time difference in days, and $N$ is a hyperparameter scaling how fast the score decays, so that the score is halved after $N$ days.
$t_d$ is computed as 
\begin{equation}
     t_d = D_t - D_o,
\end{equation}
where $D_o$ is the target days for maximum score, and $D_t$ is the date of the target recording.
The score can be inversed to penalise close recordings and promote far away recordings as
\begin{equation}
    \text{Date score} \leftarrow 1-\text{Date score}.
\end{equation}
The scoring mechanism could also be augmented to 
\begin{equation}
\text{Date score}
    \begin{cases} 
\frac{N}{t_d+N+-D_o}, & \text{if } D_t > D_o,\\
\frac{N}{-t_d+N+D_o}, & \text{else},
\end{cases}
\end{equation}
which would allow for date scores around any arbitrary date or time delta.

\subsubsection{Speed Similarity Score}
This score is measured the same as the date score, replacing dates with speed measurements.

\subsubsection{Variance Similarity Score}
This score is measured the same as the date score, but on the variance of the trend and bias levels, which are lists over time as explained in the CM vector store appendix.

\subsubsection{Vector Similarity Score}
Signal embeddings can be computed with any function working on any target signal property.
In this implementation, the signal embedding is defined as the order analysis of the spectra, computed by upsampling the 3200 long spectrum vector based on its shaft speed, to normalise each spectrum vector so that each element in the vector corresponds to the same order in the machine regardless of shaft speed.
This is achieved with the following equation:
$$
max\_resolution = recording\_length * max\_speed / min\_speed
$$
$$
resample\_target = max\_resolution * min\_speed / speed
$$
$$
resample\_factor = max\_resolution * resample\_target = recording\_length * max\_speed / speed 
$$

The vector similarity score is then computed as the cosine similarity between an input signal $\mathbf{S_I}$ and a signal from the vector base $\mathbf{S_{VB}}$  as follows:
vectors of identical length, per Equation \ref{eq:cosine_similarity}, either the signal embeddings (order signals), the unprocessed spectra, or the raw vibration time series.
The cosine similarity between two vectors \( \mathbf{A} \) and \( \mathbf{B} \) is defined as
\begin{equation}
\label{eq:cosine_similarity}
\cos(\theta) = \frac{\mathbf{S_I} \cdot \mathbf{S_{VB}}}{\|\mathbf{S_I}\| \|\mathbf{S_{VB}}\|},
\end{equation}
where:
\begin{itemize}
    \item \(\mathbf{S_I} \cdot \mathbf{S_{VB}}\) represents the dot product of the input signal vector \(\mathbf{S_I}\) and a signal vector \(\mathbf{S_{VB}}\) from the vector store,
    \item \(\|\mathbf{S_I}\|\) and \(\|\mathbf{S_{VB}}\|\) are the Euclidean norms (lengths) of vectors \(\mathbf{S_I}\) and \(\mathbf{S_{VB}}\), respectively.
\end{itemize}

\subsubsection{Signal Features Score}
Signal features scores are computed based on statistical properties of recording data, as the normalised difference between two sets of features computed for the source and target signal.
Features can be statistical properties such as min, mean, max, variance, entropy, kurtosis, or skew.
If information such as component dimensions and shaft speed are available, more advanced analysis such as kinematical bearing analysis such as computing characteristic features for BPFO, BPFI, and BSF faults, could also be performed \cite{randall2011rolling}.
%
%
%
%
%
%
%
In this study, feature similarity scores are only computed for the variance of the trend and bias, as component dimensions are not available in the current data extraction pipeline.
Thus, further implementations of existing state-of-the-art CM signal analysis research is a natural improvement to the chunk-level distance metrics, that could improve performance in chunk-level retrieval metrics. 

\subsubsection{Multimodal Distance}
Multimodal distance metrics, such as those used in text-and-image transformers as described in MRAG, are not used in this implementation.
However, multimodal signal encoders \cite{lowenmark2021technical} are a natural next step to include in a similar way that CLIP is used in normal MRAG.

\subsection{Aggregating Scores}
Once each sub-score is computed, the total score has to be aggregated through a meta-scoring function.
The simplest approach is to sum all individual scores.
This necessitates that scores are normalised, either over all data or over a query batch.
The user then has to decide on weighing functions for each score, e.g., if datetime differences are less or more important than point name scores.
Another possible approach is to concatenate scores to form new vectors for similarity scoring.
For example, graph metadata such as asset similarity, point similarity, speed similarity, and recording date similarity, and features such as skew and kurtosis.

In this study, scores are summed with a weighing matrix that normalises all data across the vector store, and favours annotation and signal embeddings five times over other distance metrics.
The weighing function is a hyperparameter that can be changed in the function or tool call to the retriever, which can be used to further investigate trade-offs between different scores.

Retrieved data is often cluttered and time consuming to interpret directly from a human perspective, e.g., in the case of 50 annotations with point and asset information concatenated.
Therefore, retrieved data is rarely directly provided to the user, and instead serves as the basis for generation of the retrieval augmented response by a generative agent.

\section{Generative Agents}
\label{sec:agents}
\begin{figure}[ht]
    \centering
    \includegraphics[width=1.0\linewidth]{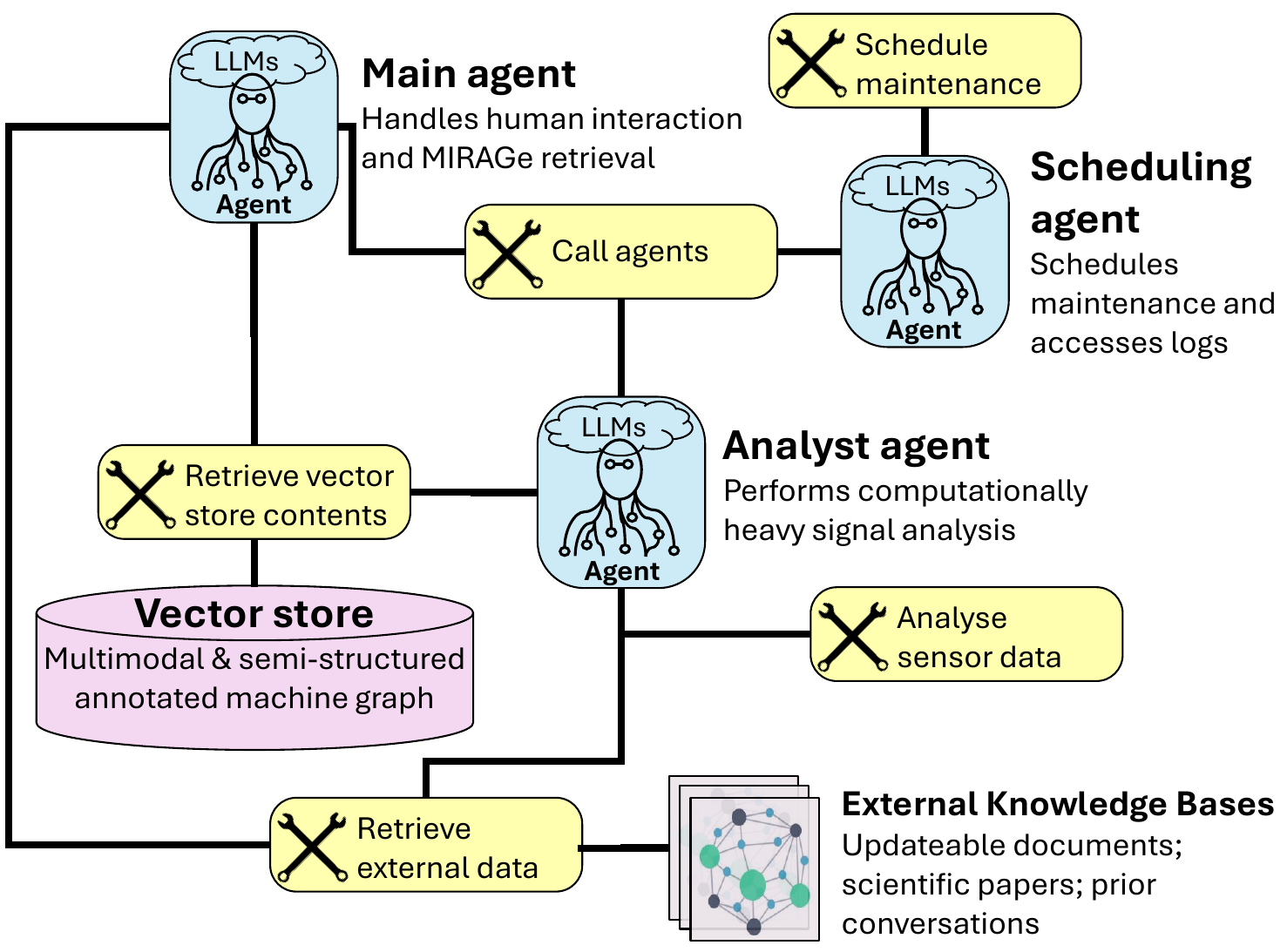}
    \caption{The three active agents with the main groups of tools used to facilitate an industrial chatbot.}
    \label{fig:agent_network}
\end{figure}

Generation is handled primarily with generative LLMs, per Equation \ref{eq:rag-multi-joint}, though some tests approximate LLM output with rule-based approaches to facilitate large-scale testing without LLM generation as the bottleneck (described in Section \ref{sec:rule_based}).
%
%
%
The LLM can not access the vector store directly, and as is common with RAG approaches, is instead conceptualised as an agent with \textit{tools} that facilitate external function calls to, for instance, retrieve relevant data from the vector store.
The design of the agents is thus critical to ensure reliable retrieval and generation from the CM vector store, and to facilitate assistance in a CM workflow through maintenance scheduling and decision intelligence.
Primary components of the design space of agents are: what LLM is used as the ``reasoning" part; what tools agents are equipped with and what functionalities they facilitate; and
what information agents have access to through prompts/tool descriptions, considered to be internal knowledge bases (KBs), or from documentation access with tools, considered to be external KBs,

In this work, three agents were developed together, to assist with different parts of the challenges laid out in Table \ref{tab:knowledge_gap}.
Additionally, a fourth agent was developed for research evaluation purposes, but not for user interactions.
These agents are: 
\begin{itemize}
    \item \textbf{the main thinker:} handles user interactions, text-retrieval, KB management, and calling other agents,
    \item \textbf{the CM analyst:} performs signal-based retrieval and analysis, and fault diagnosis assessments,
    \item \textbf{the maintenance scheduler:} manages alarms, fault detection, and maintenance planning,
    \item \textbf{the evaluation agent:} evaluates fault diagnosis made by the analyst agent.
\end{itemize}
Table \ref{tab:agent_tools} lists and briefly describes the tools each agent has access to, and Figure \ref{fig:agent_network} shows the connections between these agents and an overview of the main tool categories used by each agent.
%

\begin{table}[t]
    \centering
    \caption{Agents and tools used to process CM data. Here, * indicates a custom tool developed in this work.}
    \label{tab:agent_tools}
    \begin{tabular}{|c|c|c|}
    \hline
     \textbf{Agent}    & \textbf{Tools} & \textbf{Descriptions}\\
     \hline
     \makecell{Main \\ thinker*} &   \makecell{
fetch\_arxiv
\\ query\_MindRAG\_HY*
\\ how\_to\_query\_HY*
\\ check\_HY*
\\ check\_KB*
\\ update\_KB*
\\ check\_KB\_assets*
\\ update\_KB\_assets*
\\ cpt\_feature\_changes*
\\ web\_search
\\ exit\_graph
\\ CM\_agent*
\\ maintenance\_planner
\\ final\_answer} &   \makecell[c]{
 pre-made tool to search ArXiV
\\ retrieve data from the vector store
\\ helper KB tool for complex retrieval queries
\\ helper KB tool to ensure viable queries
\\ access KB updated by the agent
\\ update KB accessible by the agent
\\ asset-specific KB
\\ asset-specific KB
\\ compute changes based on input functions
\\ pre-made tool to search the web using tavily
\\ standard exit agent tool
\\ call the CM analyst agent
\\ call the maintenance planner agent
\\ formulate a response based on findings}\\ \cdashline{1-3}
    \makecell{CM \\ Analyst*}  &   
      \makecell{
      plot\_chunks*
\\ query\_MindRAG\_full*
\\ how\_to\_query\_HY\_full*
\\ check\_HY*
\\ check\_KB*
\\ write\_to\_KB*
\\ get\_HY\_from\_CK*
\\ get\_note\_from\_CK*
\\ get\_recording\_from\_CK*
\\ write\_code
\\ exit\_graph
\\ write\_prediction*} &   
 \makecell[c]{
 visualise currently analysed data
\\ retrieve any data from the vector store
\\ helper KB tool for complex retrieval queries
\\ helper KB tool to ensure viable queries
\\ access KB updated by the agent
\\ update KB accessible by the agent
\\ navigate from signal instance to hierarchy
\\ navigate from signal instance to annotation
\\ navigate from signal instance to recording
\\ pre-made tool to write and execute code
\\ standard exit agent tool
\\ formulate a fault diagnosis response}
\\ \cdashline{1-3}
\makecell{Maintenance \\ scheduler*} &   
      \makecell{user\_profile \\ access\_memory} & 
      \makecell{pre-made tool to keeps track of analyst users  \\ write/read maintenance schedules} \\ \cdashline{1-3}
      \makecell{Evaluation*}   &   \makecell{write\_evaluation*} & \makecell{evaluate analyst graph output} \\
      \hline
    \end{tabular}
\end{table}

\subsection{Subagents}
The agent framework is designed to be modular and efficient.
As different tasks, such as scheduling, graph text retrieval, and signal-based retrieval, require different reasoning capabilities, memory storage, and tokens processed, multiple agents with different prompts and tools can act more robustly than one massive agent.
\subsubsection{Main Thinker}
The main thinker graph is the entry point for most interactions, and handles document-level retrieval and reasoning, as well as calling the analyst and maintenance scheduler graphs.
It is equipped with a variety of tools to meet user needs, and a memory function to reason across multiple queries or with multiple users.
The ``fetch\_arxiv", ``web\_search", ``exit\_graph", ``final\_answer", and invoke other agent tools, are tools developed from existing LangChain or LangGraph templates.
In particular, fetch\_arxiv, web\_search and write\_code are directly based on existing code templates and enable the agent to interact with scientific literature via ArXiv, navigate online website to search for information, and write Python code, while ``exit\_graph", ``final\_answer", and invoke other agents tools all are common functionalities to add to graphs.
Functions that access external KBs require user verification and transparency with regard to what data is retrieved and from where.
This is especially important for internet or ArXiV searches, as this data is unmoderated and therefore liable to unreliable data for generation.

``query\_MindRAG\_HY", ``how\_to\_query\_HY", and ``check\_HY" are all tools developed to facilitate document level retrieval of hierarchy data (HY), where query\_MindRAG\_HY is the main retrieval tool, while how\_to\_query\_HY is an extensive instruction on all viable tool parameters based on the retrieval framework described in Section \ref{sec:retrieval}, and check\_HY is a safeguarding tool called prior to querying to correct erroneous user requests.

The four KB tools read and write to either a general KB, or asset-coded KBs intended to reflect machine-specific knowledge present as tacit knowledge in real-world operations.
The agent reads and writes to these KBs either based on reflection on the conversation, or when prompted by the user, and uses the accumulated knowledge to improve performance.
The main agent can also do simple signal analysis by propagating signal encoding functions to the vector store with the cpt\_feature\_changes tool.
This tool does not analyse input signals, as the CM analyst does, but instead searches in the vector store for changes in feature values to detect faults or changes in indicator values.
The values are computed using functions provided by the analyst, e.g., high sensor bias variance for cable and sensor faults, or rising values in detected faults such as rising levels of overtones in envelope signals of currently monitored BPFO faults.
These statistical properties are relevant for fault diagnosis, but also act as stand-ins to prototype how more complex fault indicators, e.g., the BPFO characteristic frequencies, could be integrated into the agent.
Pre-trained machine learning models could also be used if accessible to the model.
This aspect of signal analysis is provided to the main agent rather than the signal analyst agent as it does not require processing a lot of tokens by the LLM, and thus flows well in a longer conversation chain.

%
%

Finally, the exit\_graph and final\_answer facilitate two ways of tool-based graph exits which force certain templates of user responses.

\subsubsection{CM Analyst}
\label{sec:CM_analyst}
The CM analyst agent handles large amounts of data that would fill up the memory of the main agent over long conversations.
The token width of modern large LMs is 128-200k tokens, which enables long conversation, but can be exceeded in extensive signal analysis.
%
Therefore, memory is stored and reset between chains of thought and responses, i.e. after one fault has been thoroughly analysed, that memory is stored as one thread, and new fault diagnosis proceed with a blank slate.
Therefore, the main thinker graph is necessary to facilitate smooth user interactions, where the main thinker graph can call different CM analyst threads.

The CM agent has access to support tools, such as a tool to plot currently analysed data; three tools to go from a specific measurement chunk to other parts of the graph; a pre-made tool that facilitates code writing; and response tools.
The main functions however are obtained with the query hierarchy tools.
The query\_MindRAG\_full tool and the support tool facilitate retrieving signals from annotation and hierarchy queries, or annotations and hierarchy information from signal queries, where the latter is particularly relevant for experiments described in Section \ref{sec:LLM_based_generation}.

\subsubsection{Maintenance Scheduler}
The maintenance scheduler graph is a direct implementation of the taskmAIstro graph from the LangGraph templates\footnote{\url{https://github.com/langchain-ai/task_mAIstro}}, altered to instead schedule maintenance follow ups for detected faults.
This highlights how a modular approach makes for easy additions to meet industry needs, and it would for example be trivial to extend the agents to work with a voice-to-text layer as well, as is done with one taskmAIstro template from the LangGraph tutorials.

\subsection{Evaluation}
\label{sec:evaluation_agent}
The evaluation graph is only used during experimental setups to validate results by comparing agent output with true notes, taking different point and asset properties and recording-to-note timedeltas into account.
By relying on an external graph, it is impossible for the analyst or main thinker graphs to influence their assessment by reading the true note.
Instead, the analyst graph makes an assessment without having access to the note label, and the assessment is compared to the note by the evaluation agent.

\subsection{Prompts}
The prompts for each agent are described in the ``prompts" section of the appendix \ref{appendix:agent_prompts}.
The prompts were designed to integrate technical knowledge based on expert feedback, but knowledge can be integrated further during runtime using KB tools as well.

\section{Experimental Framework}
\label{sec:framework}
This section describes the experimental framework in which the agents were tested and under what conditions the results were obtained.
Four main categories of experiments were performed:
\begin{itemize}
    \item 
    agent-based knowledge retrieval,
    \item rule-based machine component retrieval
    \item rule-based fault diagnosis,
    \item agent-based fault diagnosis and alarm generation. 
\end{itemize}
Here, retrieval denotes function calls to and returned data from the vector store, as described in Section \ref{sec:retrieval}.
Generation describes the process of using retrieved data to perform a task, e.g., answering a user query or predicting a fault.
User interactions are always handled by the agents, while retrieval and generation can be done by agents, or through manual tool calls or rule-based inference for large-scale testing.
Section \ref{sec:knowledge_retrieval} outlines experiments done for knowledge retrieval, Section \ref{sec:rule_based_generation} outlines experiments done for rule-based signal retrieval and inference, and Section \ref{sec:LLM_based_generation} outlines experiments done for LLM-based signal retrieval and inference.

\subsection{Generation}
Document level similarities are retrieved with a similar concept to normal MRAG, where retrieval is done in multiple modalities with multiple distance metrics, and the top k documents are retrieved and used for generation.
For recording chunks, the general principle is the same, with an added level of complexity, as each document consists of multiple recordings grounded in the physical state of a machine, and unlike normal retrieval and generation tasks, chunk retreival can facilitate fault inference.
Inference tasks use the multimodal property of the MindRAG vector base to retrieve similar recording chunks as an input chunk, and uses the hierarchy information such as what machine parts the similar chunks come from, the contents of the annotations associated with the similar chunks, the time delta between the similar chunk and its annotation, etc., to generate answers to questions normally not answerable by LLMs, with or without RAG.
As faults exist at the point or sometimes asset level, while retrieval occurs at the chunk level, and each chunk retrieval can return hierarchy information of conflicting nature, the inference mechanisms to go from retrieved chunks to augmented generation is more complicated than typical RAG mechanisms.
$z$ is therefore redefined, so that an input signal query will drop retrieved signals before proceeded to the generation step in Equation \ref{eq:CMrag-multi-joint}.
\begin{equation}
P(y \mid z) \approx 
\sum_{d \in \mathrm{Top\text{-}k}(p(d \mid r))}
P_\mathcal{R}(d \mid z) \, P_\mathcal{G}\bigl(y \mid z, d\bigr)
\;=\;
\sum_{d \in \mathrm{Top\text{-}k}(p(d \mid r))}
P_\mathcal{R}(d \mid z)
\prod_{i=1}^{N}
P_\mathcal{G}\bigl(y_i \mid z, d, y_{1:i-1}\bigr),
\label{eq:CMrag-multi-joint}
\end{equation}

\subsection{Knowledge Retrieval}
\label{sec:knowledge_retrieval}

Knowledge retrieval here symbolises text-based retrieval, i.e., retrieval done between in the annotation and machine graph hierarchy space.
Example queries are those provided in Table \ref{tab:knowledge_gap} in the "Text-based analysis -- knowledge retrieval" half, such as "which sensor positions have had the most faults".
The knowledge retrieval function was tested through user interaction with the agent, having a user ask relevant questions and follow-up questions and taking note of the agent's thoughts, actions, and responses.
These questions were handled by the main agent and the scheduling agent, using prompts, tool descriptions, and knowledge bases to guide retrieval, interpretation, and answer generation.
The results are presented in Section \ref{sec:knowledge_retrieval_results} and in Appendix Section \ref{appendix:knowledge_retrieval}.

\subsection{Signal Retrieval}
The main use-case for full multimodal retrieval is to facilitate analysis of signals based on historic cases.
One or more input signals are used to retrieve similar signals, and the annotation and graph properties of these associated signals are used for further inference.
Section \ref{sec:signal_retrieval_results} gives an example of this retrieval, showing two signals from the same point with the same annotation, and the signals that have the highest correlation score with these two signals.

This method was used for three downstream tasks: signal-to-component rule-based generation, signal-to-annotation rule-based generation, and signal-to-annotation LLM-based generation, presented in Sections \ref{sec:rule_based_generation} and \ref{sec:LLM_based_generation}.
Here, signal-to-component generation defines masking the component information, i.e. asset and point information, from source recordings in the vector store, and using asset and point information from $N=10$ similar components to infer what component the source recording is from.
Likewise, signal-to-annotation generation is the process of masking annotation information from the $10$ recordings, and using information from retrieved recordings to infer the contents of the source annotation.
Rule-based approaches to generation were devised to facilitate larger scales of testing, without involving LLM calls, by directly using the tools with input devised with a script, then analysing the contents of the retrieved.
Two cases are considered for both rule-based tasks, one where the source data, i.e. historic data from the same component/asset, are included, and one where only data from other components are included.

\subsection{Rule-based Generation}
\label{sec:rule_based_generation}

The rule-based generation approach consists of doing keyword searches in the retrieved information, then using a majority voting mechanism with the extracted keywords to map input signals to predicted keyword classes.
For component prediction, this is achieved by extracting parts of the point name, dropping sensor placement information to keep only component information.
For example, "TC 67 DS VE3", meaning "drying cylinder number 67, drive side, vertically mounted sensor with an envelope filter of range 3", is converted to "TC 67".
The components can then be further grouped so that all drying cylinders are in the same category, for instance, as is shown in Figure \ref{fig:asset_points_included} and \ref{fig:asset_points_excluded}.
This approach likely captures relevant information well, as the language used for component descriptions consist only of formulaic technical terms, and with sufficient prior knowledge essentially all component classes can be mapped, likely better than an LLM could.

Component retrieval is potentially valuable to identify unknown signals, and is an argument for why MindRAG retrieval can work while unlabelled transfer learning on industry data is difficult;
if component retrieval is possible with high accuracy, that implies that the signal retrieval captures signal similarities with regard to noise of interference, which further implies that many machine learning methods will face a challenge due to this heterogenous noise space, similar to as if a dog breed classifier had samples where some backgrounds only corresponded to one breed.

Annotation retrieval looks for keywords in the note contents, using prior knowledge to map annotations to "replaced", "cable/sensor faults", "bearing faults", "breakdowns", and "miscellaneous comments" for annotations that were not identified by other keywords.
The results are presented in Section \ref{sec:rule_based_results}.
This approach loses a lot of vital language information for fine-grained prediction, e.g., completely failing to capture negations, severity indications, and multiple fault mentions in the annotation; time differences between recordings and annotations; similarities in component types, asset place, etc., which is better captured by an LLM contextually interpreting this information.
However, it offers a way to test different hypotheses with a weak predictor to evaluate different retrieval schemes or data used for retrieval without requiring expensive LLM calls.
Accurate annotation retrieval with the same component included is likely due to signals from the same component and thus same annotation influencing the score.
However, if annotation retrieval with source asset removed is even moderately accurate, this implies that when the same noise space is omitted, fault features influence retrieval.


%
%
%
%
%

%
%




\subsubsection{Voting Mechanisms for Rule-based Approaches}
\label{sec:rule_based}

We implemented a rule-based generation set-up to facilitate fast testing of the MindRAG system without requiring LLM input or output processing, which is the foundation for the results presented in Section \ref{sec:rule_based_results}.

\paragraph{Majority votes}
\label{sec:majority_votes}
Rule-based majority voting is the simplest approach to generate inference by defining a small ontology that acts on the retrieved chunks/documents, e.g. keywords in the annotations whose meaning is affected by whether the timedelta is positive or negative, and point type analysis based on if-else statements derived from the machine hierarchy.
Inference is then achieved at the chunk level by passing the top k associated document through the ontology to generate votes, then counting which class got the most votes, which becomes the chunk prediction.
Inference at the document level, i.e. for multiple chunks associated with the same document, can be achieved either through a majority vote of majority votes, or by summing up all the votes at the chunk level together, then taking the max of the entire document level ontology classes.
This approach works best when ontologies can be clearly defined, e.g. for point name inference, and is expected to work less well for more complex inference, such as fault prediction based on note contents.
Fault inference becomes especially difficult when mimicing human analysis, i.e. at the asset level, as even if the document level prediction had a 100\% accuracy, faults might be present in only one document (point), and thus lose the vote to fault-free documents.

\paragraph{Sum of scores}
\label{sec:sum_of_scores}
The sum of scores approach is the same as the majority vote, but votes are now scaled based on the distance metric, summing scores rather than counting votes.
This approach is used in the non-LLM inference experiments in Sections \ref{sec:rule_based_generation} and for the results shown in Sections \ref{sec:rule_based_results}.
The hyperparameters for these voting mechanisms are: what data to include (filter hyperparameters), how many chunks to include in the voting (top k), and how to interpolate from chunk to document (meta inference rule).
In the experiments mentioned above, top k is set to 10.
%
%
Data is filtered based on timedelta of the target chunks in the CM vector store to examine which time slices have the best and worse performance, and meta inference is done by voting first at the chunk level then at the document level.

\paragraph{kNN}
The voting mechanisms from \ref{sec:majority_votes} and \ref{sec:sum_of_scores} can be generalised to a kNN algorithm, finding the closest chunks in a kNN space and relying on established kNN distance metrics and weighting schemes.
%
This approach features more hyperparameters than the voting mechanisms above, and can suffer more from the curse of dimensionality, but can also facilitate more complex analysis without involving language model inference, as described in the next sections.

\subsection{LLM-based Generation}
\label{sec:LLM_based_generation}
In standard MRAG and at the hierarchy level of MindRAG, generation is done by an LLM based on the retrieved multimodal information, either in one modality, e.g. text based on image search, or by a multimodal model.
This approach can be used for textual data, i.e. the hieararchy data and in particular the annotation, of retrieved chunks as well.
At the recording chunk level this behaves slightly differently;
retrieval done in the signal space will naturally always retrieve the most relevant signals given the retrieval rules (the retrieval rules can of course be poorly formulated), but the relevance for downstream tasks will be influenced by document-level information.
For example, a recording from ten days before an annotation reading "roller replaced OK" implies that a fault feature should be present (though not which), while a recording from ten days after can be expected to contain no fault features.
Additionally, whether retrieved signals are from e.g., the same type of asset/component or the same type of filter can also have an impact on the generation, where for instance sensor faults can cause disturbances that drown noise to the extent where recordings from vastly different assets are retrieved, which is less likely to happen for burgeoning bearing faults.
These considerations can be modelled to some extent with a rule-based approach, but it is virtually impossible to capture all nuances of language in the annotations and correlate this to different interpretations based on graph hierarchy information.

Another added level of complexity for MindRAG signal interpretation is that annotations appear at the asset/component level, but recordings are grouped at the point level, and faults can be both at the component level or the sensor, i.e., point, level.
Thus, if an asset consists of four points, e.g., velocity and envelope for a sensor mounted on the free side and one mounted on the drive side, it is possible that an annotation is written due to a fault visible only in one point, for example the envelope at the free side.
Perfect point-level inference, i.e., the ideal mapping of recordings to whether they indicate faults or not, would then yield an accuracy of 25\%, where the "label" reads "fault" while only one quarter of the samples contain fault features.
If inference is instead done at the asset level, a model must be capable of understanding that even though a large majority of points and samples point towards no fault being present, the faulty point is more important for inference.
This is difficult to model with a rule-based approach, but can be explained to varying degrees with natural language, depending on technical expertise.

LLM agents capable of advanced language interpretations and equipped with prompts that explain the properties described above have the possibility to perform more nuanced document interpretation for improved fault diagnosis. 
%
%
Such agents need to be prompted to understand the how to infer fault diagnosis information based on related chunks, to answer questions such as:
\begin{itemize}
    \item how should annotations be interpreted based on the timedelta, in particular in relation to MWOs, maintenance action descriptions, and fault severity descriptions?
    \item how should annotations be interpreted based on the point type, point name, and asset path?
    \item how should fault descriptions be interpreted for different faults, in particular when a chunk has conflicting associated notes, e.g. "cable fault detected", "BPFO low levels keep watch", or "roller replaced"?
    \item how should document-level analysis be compiled to one asset-level fault diagnosis report, taking into account that different points might indicate different faults, that only one point might warrant faults, or that the data supplied might contain no faults to report on?
    \item how should different levels of severity be integrated into the agent's analysis? Can the agent be more sensitive to critical faults that might preclude machine breakdowns at the expense of false positives, while penalising too frequent cable or sensor fault alarms, at the expense of false negatives? 
\end{itemize}

These questions are in part addressed by the tools and prompts in the agent design in Section \ref{sec:agents}, but also decide how the agents are integrated and tested.
%
To investigate how these agents deal with these questions, an experimental framework was designed with the following steps:

\begin{enumerate}
    \item \textbf{Filter vector store:} Decide on the contents of the vector store. What timedeltas should be included? For instance, it is more relevant to use data from around the annotation date than from 200 days before or after the annotation.
    \item \textbf{Retrieve input data:} Retrieve data from the vector store based on asset, point, note, and recording properties, depending on what task to evaluate.
    For example, using data from only after maintenance actions would run the agent on mainly healthy data, while using data ten days before a work order would likely feature higher severity recordings.
    
    \item \textbf{Compute vector similarities:} Compute embedding similarities between input data and vector stores based on chosen similarity metrics. In all tests, the dot product of the spectrum or envelope order analysis vectors was used for similarity, and vector store elements from the same asset as the input asset were dropped.

    \item \textbf{Select similar vectors:} Select top k similar MindRAG chunks and use for downstream analysis. For all tests with the LLM agents, top k was selected as 5.

    \item \textbf{Analysis:} Use MindRAG chunks and prompt knowledge for inference to assess if one or more points from an asset contain faults.

    \item \textbf{Alarm decision:} Based on the analysis, the agent decides if it wants to generate an alarm.

    \item \textbf{Generate response:} Write a report to the user describing findings and if any maintenance actions are suggested.

    \item \textbf{Evaluate response:} The evaluation agent assesses if the report is correct and writes a summary to file.
\end{enumerate}

To simulate real working environments, the agent was given data from all points in an asset on a day-by-day basis, starting with all data from the start date of the filtered input space, then proceeding one day at a time until the agent either voluntarily exits to generate an alarm, or because it reached the end date.
The agent thus reads associated annotations that in theory reflect the state of the component, and as faults develop, the contents of the associated annotations change to reflect this.
This approach is based on the following assumptions:
\begin{enumerate}
    \item If a fault is present in at least one point in the analysed asset, it will be reflected in the recordings connected to that point (data robustness),
    \item if a recording contains features from a fault that exists in other assets in the vector base, MindRAG can retrieve recordings from other assets with the same or similar fault (MindRAG retrieval),
    \item recordings with fault features that exist close to an annotation in the vector store will have an annotation that describes this fault (human knowledge integration),
    \item based on the contents of the associated annotations and the time difference between the associated recording and its annotation, fault information can be inferred (LLM inference).
\end{enumerate}
These assumptions do not always hold, as fault features appear, disappear, and reappear over time in different patterns for different faults, and recordings connected to faulty signals will thus occasionally appear healthy, in particular for cable or sensor faults.
Thus, healthy data will sometimes be mapped to faulty annotations, and faulty data will sometimes be mapped to healthy annotations.
Likewise, early signs of bearing degradation might be diagnosed in some points but undiagnosed in others, which results in effectively unannotated points retrieving other effectively unannotated points, while both are in practice annotated due to the annotations being linked at the asset level.
The agent must thus use its knowledge bases to reason about how to interpret different associated annotations based on time deltas and point types.
As annotations are written at the asset level, the agent must also reason about the state of a component based on the properties of multiple points, and prioritise which fault might warrant an alarm and annotation, while avoiding both false positives due to premature or incorrect alarm generation, and false negatives by overdue analysis.

\paragraph{Causality}
It could be argued that for the causal argument to truly hold, only recordings from a date earlier than the "current" date of the input recording should be included.
However, this is mainly a concern due to testing being done in the same time span as the associated recordings.
Regardless of whether testing is done on the same dataset by excluding any information that might unfairly aid inference, or in a dataset one year later, or even one year earlier, similarity will be time-agnostic as long as operational conditions do not change drastically over time.
In our data, the production line remained essentially the same, so as long as the input recordings are treated in a causal manner the tests will generalise across time given similar working conditions.

\paragraph{Evaluation}
The results from agent experiments and evaluation are shown in Section \ref{sec:llm_agent_inference}.
The question-answer chains for document-level retrieval are presented in Section \ref{sec:knowledge_retrieval_results} and the Appendix in Section \ref{appendix:knowledge_retrieval}.
The quantitative results are presented in Section \ref{sec:llm_agent_inference} and in the Appendix in section \ref{appendix:MIRAGE_runs}.

\section{Evaluation Results}
\label{PaperE:sec:results}
Generally it is desired to have both qualitative and quantitative analyses. 
Due to constraints such as data availability, annotation quality, time, and space constraints in this paper, we present selective qualitative and quantitative results demonstrating some representative features of the contributions presented in this work.

\subsection{Qualitative Knowledge Retrieval Analysis}
\label{sec:knowledge_retrieval_results}
Knowledge retrieval was tested based on user interactions inspired by the example questions from Table \ref{tab:knowledge_gap}.
For these tests, chain-of-thought reasoning was outputted by the agent to highlight internal agent reasoning steps.
A few examples of agent interactions are showcased in the appendix, namely:
\paragraph{Graph text analysis} where the agent answers the question "Please check if there is any need for lubrication in the holländeri" by going through all annotations in the holländeri and based on annotation contents and dates provides feedback.
The agent makes a sound assessment based on the annotations, and generates a reliable answer.
However, it chose one interpretation of "checking" by assuming that the user meant "please check if there are any components with prior annotations regarding lubrication that need follow up", which the question could also have been "based on signal features, please estimate if any components require more lubrication".
The latter question is beyond the scope of the main agent, as the annotations likely do not sufficiently cover what signals preclude a lubrication need, and it instead commented on the fault that occurred due to the lack of lubrication.

\paragraph{Historic insight and analysis} where the agent answers the question  "Hello! Which assets from drying cylinder group 2 have unresolved bearing fault annotations? i.e. bearing faults that have been identified but do not have a work order afterwards?".
The agent retrieves relevant assets, analyses annotations, and finds three assets with notes indicating faults that do not have a follow-up note indicating maintenance actions.

\paragraph{Updating and using the knowledge base} where the agent answers the two questions: "Do you know how to calculate if there are sensor or cable faults in the condition monitoring data?" and "Please also add to your knowledge base that the variance of the bias can be indicative of these faults, then compute if there are any present at vira topp".
The agents answer the first question by accessing the general knowledge base, processing the contents, and finding that it has a general approach for sensor faults, but not for cable faults.
It asks the user for clarification, which is provided by the second query.
The agent proceeds with updating the knowledge base, and then a tool call based on the user input.
Here the propensity of agents to over-align with recent user inputs is shown, as despite the knowledge base indicating that the max values of the trend are relevant for cable and sensor faults, the agent only uses the variance of the bias.
The agent thus ignores or forgets prior information from the knowledge base, while if all information had been there, it would likely have included both methods.
Besides this issue, the function call is correct and the generated response grounded in the retrieved data.

\paragraph{Compute feature value changes} where the agent answers the query "Hello! What are the changes in the max values for sensor faults in June and August 2021 in TG 1? Please describe the characteristics of their associated assets and points".
This example highlights a slightly longer chain of thoughts, requiring five actions to produce a final answer, to check the hieararchy, query it, analyse the data, load data and compute changes, and produce a final answer.
The agent applies the correct function to the correct data, and generates a response based on the tool output as desired from the user query.

\begin{figure}[t]
    \centering
    \includegraphics[width=0.5\linewidth]{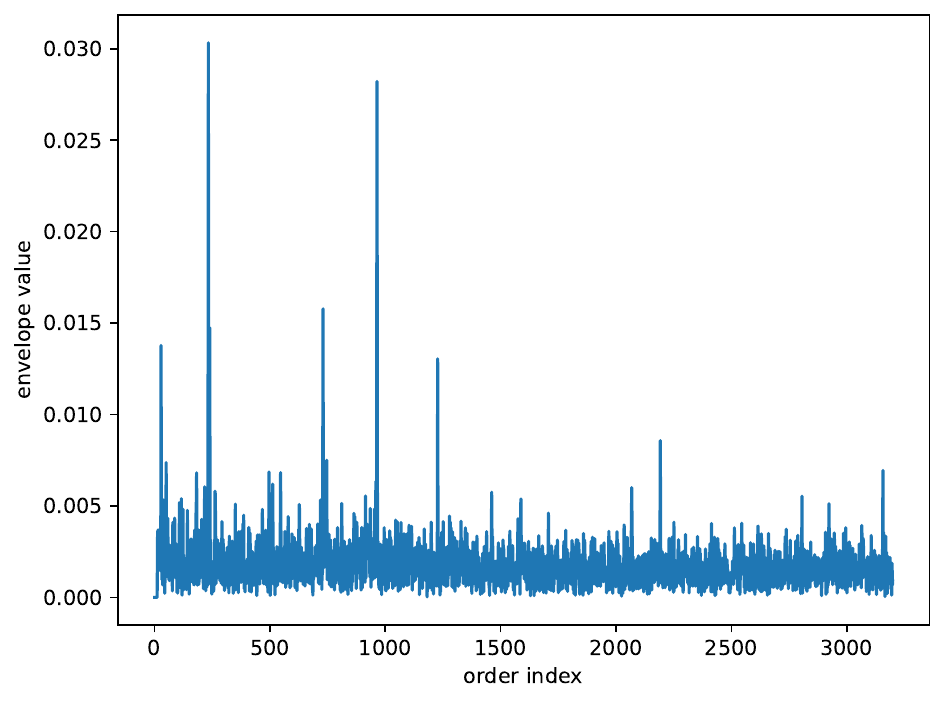}
    \vspace{-0.3cm}
    \caption{An example MindRAG input signal from a roller ("nosvals") asset with a point consisting of a horizontally mounted sensor on the drive side with an envelope filter (HE3). The measurement is taken 17 days after the annotation "BPFO clearly visible in env3, but low levels. Keep watch".}
    \label{fig:input_1}
\end{figure}

\begin{figure}[btp]
    \centering
    \includegraphics[width=0.5\linewidth]{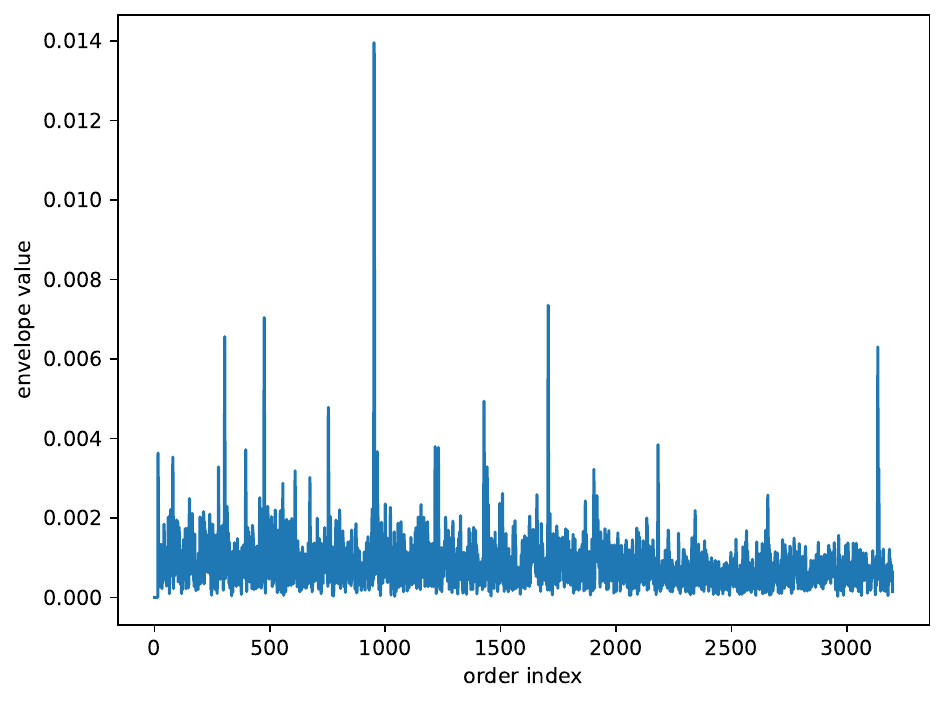}
    \vspace{-0.3cm}
    \caption{Highest ranked MindRAG retrieved signal for Figure \ref{fig:input_1}, from a drying cylinder asset with a point consisting of an axially mounted sensor on the free side with an envelope filter (HE3). The measurement is taken eleven days prior to the annotation "WO written bearing replacement".}
    \label{fig:output_1}
\end{figure}

\subsection{Qualitative Signal Retrieval Analysis}
\label{sec:signal_retrieval_results}
\begin{figure}[t]
    \centering
    \includegraphics[width=0.5\linewidth]{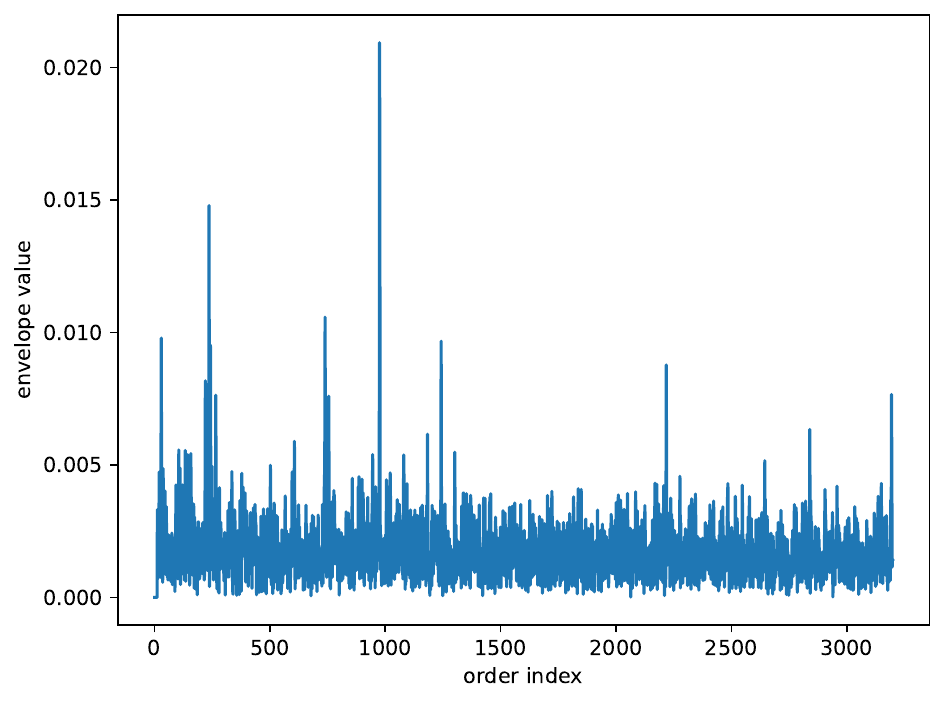}
    \vspace{-0.3cm}
    \caption{An example MindRAG input signal from a roller ("nosvals") asset with a point consisting of a horizontally mounted sensor on the drive side with an envelope filter (HE3). The measurement is taken 16 days after the annotation "BPFO clearly visible in env3, but low levels. Keep watch".}
    \label{fig:input_2}
\end{figure}

\begin{figure}[btp]
    \centering
    \includegraphics[width=0.5\linewidth]{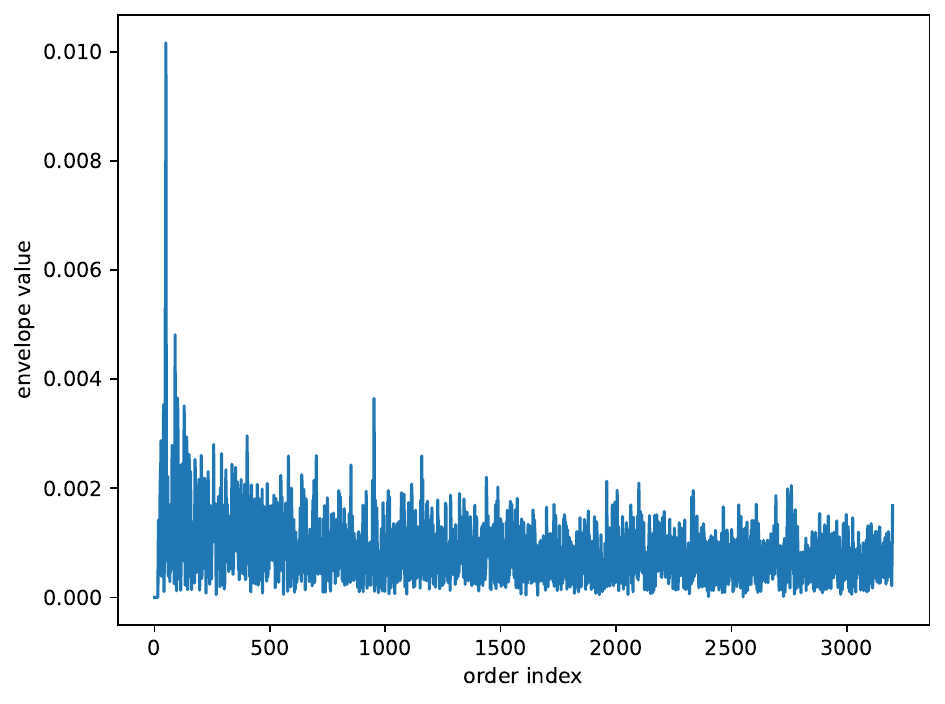}
    \vspace{-0.3cm}
    \caption{Highest ranked MindRAG retrieved signal for Figure \ref{fig:input_2}, with a point consisting of an axially mounted sensor on the drive side, with an envelope filter (AE3). The measurement is taken eleven days prior to the annotation "Sensor shall be replaced next stop".} 
    \label{fig:output_2}
\end{figure}

Figures \ref{fig:input_1}, \ref{fig:output_1}, \ref{fig:input_2}, and \ref{fig:output_2} illustrate two examples of MindRAG signal retrieval, used by the CM analyst agent as described in Section \ref{sec:CM_analyst}.
Figures \ref{fig:input_1} and \ref{fig:input_2} are examples from the same "nosvals" roller asset, from the "HE3" drive side point, indicating that the sensor has been mounted horizontally with an envelope filter active for these measurements.
The measurements are taken 17 and 16 days after an annotation reading "BPFO clearly visible in env3, but low levels. Keep watch"
The difference is that Figure \ref{fig:input_1} shows stronger indications of the BPFO fault that is being described.
This results in the highest ranked output being Figure \ref{fig:output_1}, from a drying cylinder, also with a "HE3" drive side point, which also is more strongly related to the underlying fault, having the associated annotation "WO written bearing replacement" written eleven days after the measurement.
Figure \ref{fig:output_2}, also from a drying cylinder but with an axially mounted envelope sensor ("AE3"), is instead likely free from features and likely correlates due to similar noise levels, as it is from eleven days before a "Sensor shall be replaced next stop" annotation, while not indicating any sensor faults.

\begin{figure}[t]
    \centering
    \includegraphics[width=0.8\linewidth]{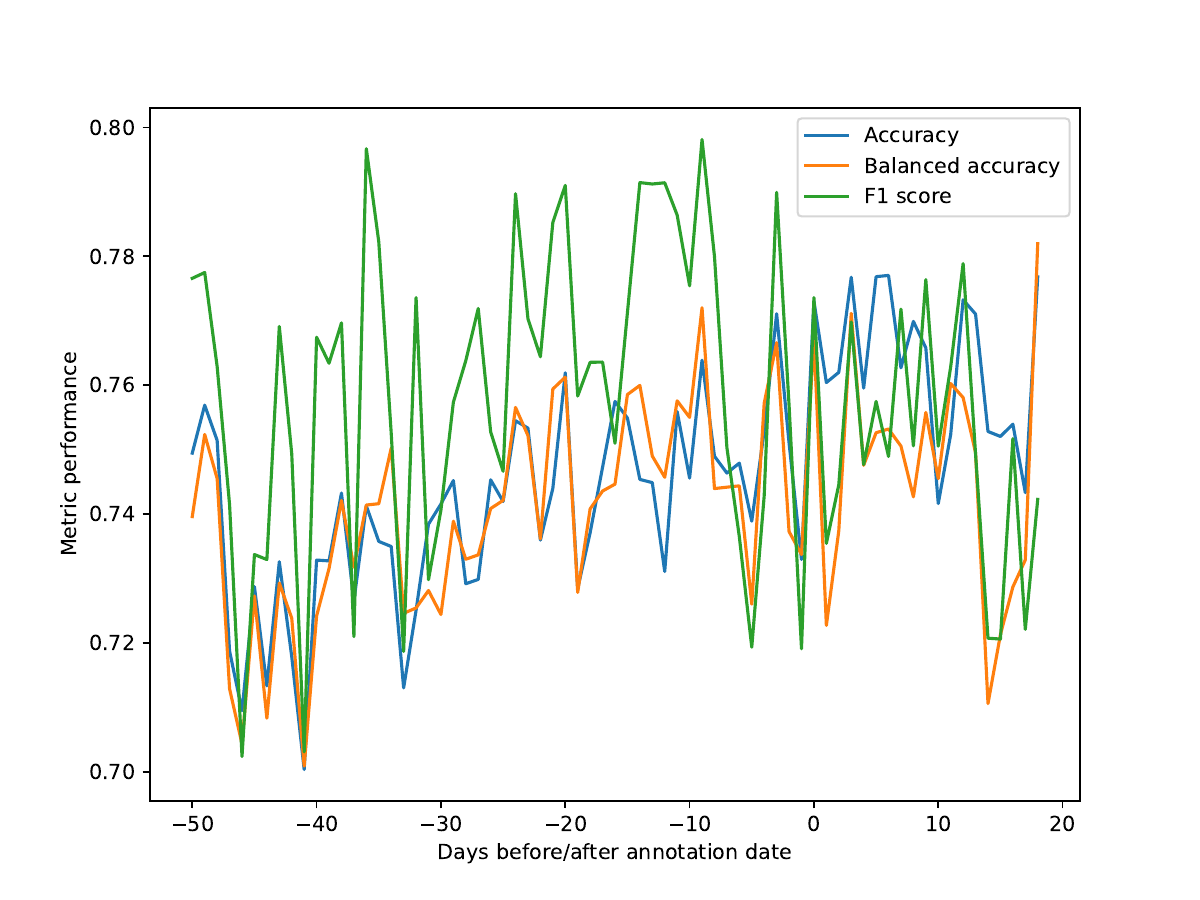}
    \vspace{-0.5cm}
    \caption{Point retrieval based on recording order distance.}
    \label{fig:included_asset_point_metrics}
    \vspace{-0.4cm}
\end{figure}

\subsection{Quantitative Rule-based Generation Results}
\label{sec:rule_based_results}
The results presented in this section are from 120 annotations with sensor data surrounding the annotation date, from 50 days prior to 20 days post annotation.

\subsubsection{Time Slice Predictions Including Source Asset}
\begin{figure}[btp]
    \centering
    \includegraphics[width=0.8\linewidth]{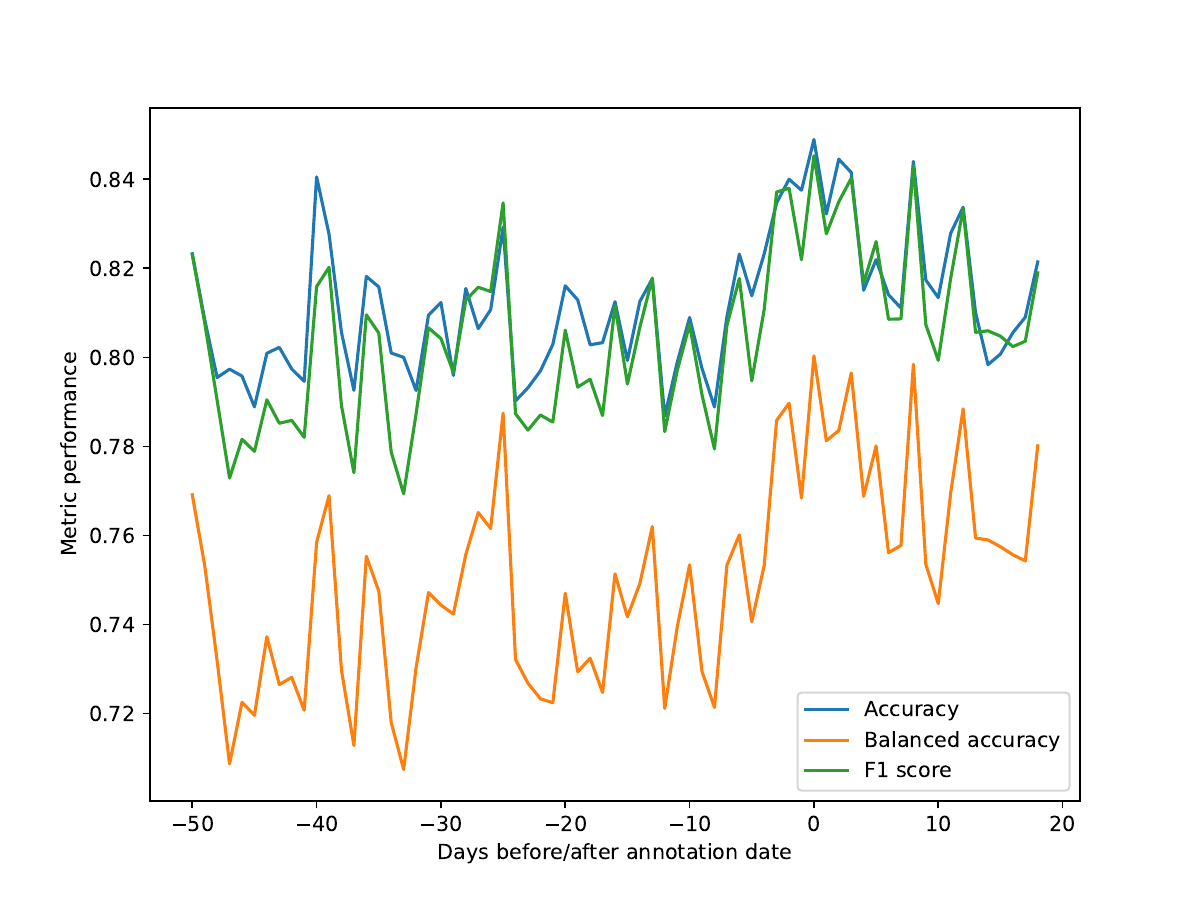}
    \vspace{-0.4cm}
    \caption{Note retrieval based on recording order distance.}
    \label{fig:included_asset_note_metrics}
\end{figure}

Figures \ref{fig:included_asset_point_metrics} and \ref{fig:included_asset_note_metrics} illustrate rule-based note and component name retrieval, as described in Section \ref{sec:rule_based_generation}.
Associated signals are retrieved using order vector similarity, from the complete vector store, excluding only the source signal.
Both figures show performance with regard to source data time delta, where "-50" indicates signals from 50 days before the annotation date, and "20" indicates 20 days after the annotation date.
%

%
%
%

\paragraph{Time slices component prediction}

Figure \ref{fig:included_asset_point_metrics} illustrates the prediction of the point name, i.e., the name of the component in the database, based on the component names of the retrieved signals.
The machine features points from 126 components, making class prediction difficult.
A balanced accuracy score and F1 score of close to 80\% thus means that the model performs well at this task.
Performance with regard to accuracy seems to increase with time, peaking around five days after the annotation date, while balanced accuracy and F1 score peaks ten days before the annotation date, though the variance in the results is rather high.
This could indicate that further away from faults, the model tends to make predictions that are more balanced over all classes, while components with more points and thus more samples are favoured when fault features are more widely present.

\paragraph{Time slices note prediction}
Figure \ref{fig:included_asset_note_metrics} illustrates the prediction of five fault classes, defined as 'replaced', 'cable\_sensor\_faults', 'bearing\_faults', 'critical\_faults', and 'misc', based on the annotation contents of retrieved recording chunks.
This test examines how well the retrieval works under ideal circumstances, but does not simulate fault diagnosis of an undetected fault, as the human analysis of the fault is available in the vector store.
The figure shows how average performance peaks around and slightly after the annotation date, likely due to these recordings being most strongly associated with the annotation that prompted them.
In particular, comments describing a maintenance action will have a sharp shift before and after the action date, resulting in retrieval prior to the annotation date likely returning fault descriptions.
There is also an increase in accuracy around 25 days before the annotation date, though it is not clear why.
It could be due to fault features vanishing, resulting in retrieval including more of the source asset (which is not excluded here), though one would expect the performance to stay at the same level for even earlier signals then.
%

%
%
%

\subsubsection{Asset Level Prediction Confusion Matrices}
\begin{figure}[t]
    \centering
    \includegraphics[width=0.8\linewidth]{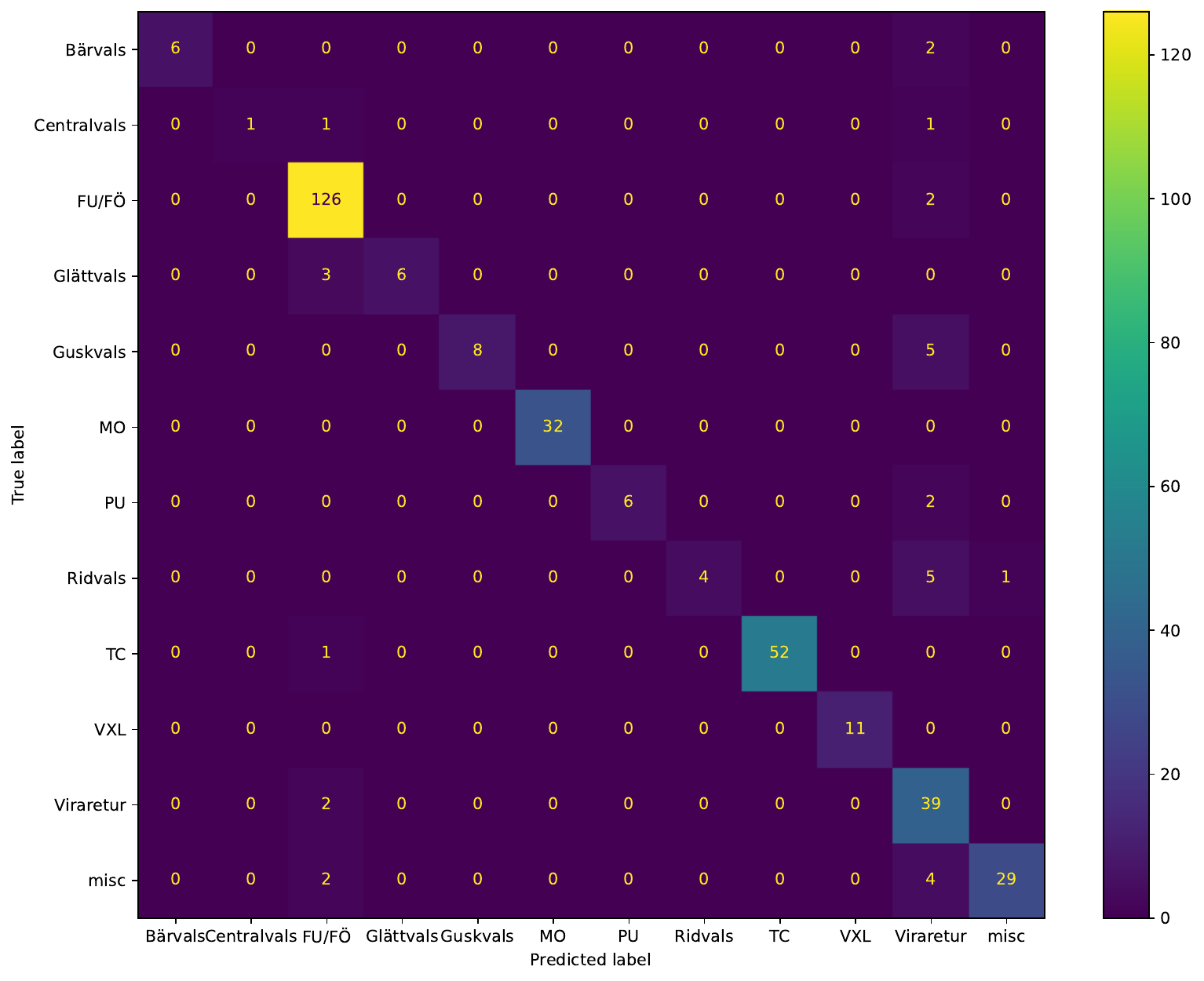}
    \vspace{-0.3cm}
    \caption{The asset level component prediction confusion matrix with the source asset included in the analysis. Balanced accuracy score of 0.77; an accuracy score of 0.91; and an f1 score of 0.83.}
    \label{fig:asset_points_included}
    \vspace{-0.2cm}
\end{figure}

\begin{figure}[t]
    \centering
    \includegraphics[width=0.8\linewidth]{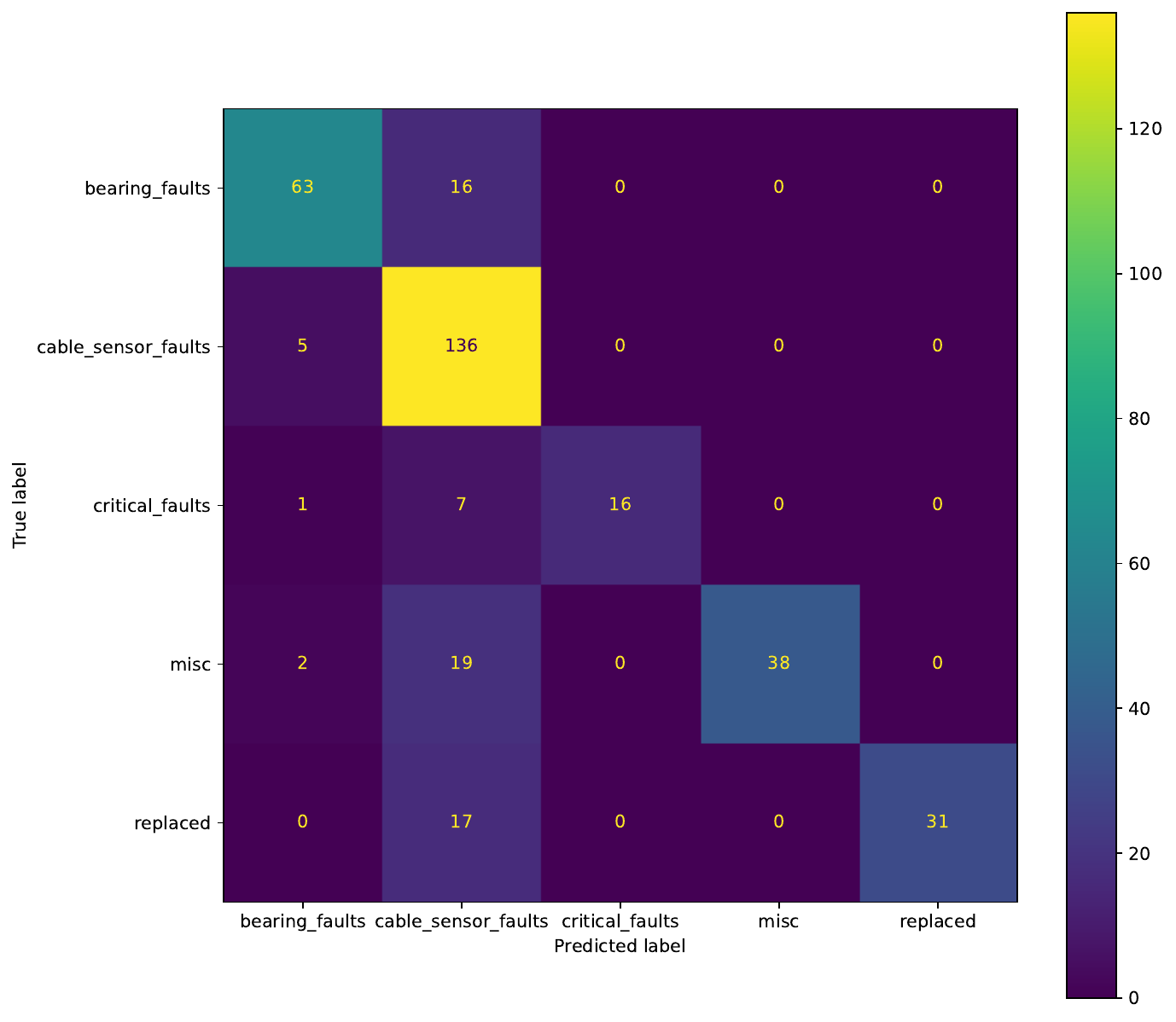}
    \vspace{-0.1cm}
    \caption{The asset level note prediction confusion matrix with the source asset included in the analysis. 
    accuracy score of: 0.82.
    }
    \label{fig:asset_notes_included}
\end{figure}

Figures \ref{fig:asset_points_included} and \ref{fig:asset_notes_included} show the point-level component and note name retrieval confusion matrices.
The asset-level prediction was achieved through majority votes across all time slices in each point, computed from the results in Figures \ref{fig:included_asset_note_metrics} and \ref{fig:included_asset_point_metrics}.

\paragraph{Asset level component prediction included source merged}
For the component name retrieval in Figure \ref{fig:asset_points_included}, the 126 machine parts were mapped down to 12 classes based on machine similarity defined with input from an on-site expert on component type, size measurements, and working environments.
This makes the results easier to interpret at the loss of resolution over component classes.
The reduced number of classes also leads to a predictable increase in accuracy, which primarily comes from the FU/FÖ concatenation as these machines are both numerous and virtually identical besides placement relative to each other (U = under, Ö = above). 
The FU/FÖ components are "ledvalsar", a type of roller, from the drying sections of the paper machine, and are therefore the most numerous.
This is reflected in high accuracy, though with 9 false positives for this class.
The most common source for false positives is the "virareturvals" (another type of roller) however, which has 21 false positives from rollers for 39 true positives and only two false negatives.
It is the second most common roller after "ledvalsar" and are situated early on in the "vira" of the machine.
It is possible that these less common rollers have similar dynamics as the "virareturvalsar", which favours the latter due to the majority voting mechanism.
This highlights the issues of data imbalance in industry datasets, which could be integrated into the voting mechanism to slightly favour less common classes.
The "MO" and "VXL" classes stick out by having no false positives and no false negatives, which is due to these components being engines and gearboxes respectively.
This data is included in all runs in this work, but performance for annotation retrieval would likely be improved as few similarities are shared with the rest of the data, leading these assets to likely have insufficient examples for accurate fault prediction.

\paragraph{Asset level note prediction source included}
The note prediction confusion matrix shown in Figure \ref{fig:asset_notes_included} shows how essentially only cable and sensor faults are cause for false positives, with very few false negatives, which likely is due to two reasons.
Firstly, these are the most common faults, which makes retrieval of these recordings more likely statistically, and thus directly impacts the majority voting.
This can likely in large part explain the low false negative rate for cable and sensor faults.
Secondly, disturbances indicating cable and sensor faults can appear sporadically both due to actual cable or sensor faults, or due to for instance an operator hitting a cable, strongly magnetic material being moved close to the sensor, etc., which has two implications:
\begin{enumerate}
    \item cable/sensor fault features are not always present for cable/sensor faults, and when they are not present the signals show no indication for these faults;
    \item cable/sensor fault features can appear in signals annotated with other faults, replacement, etc., due to signal disturbances that did not corrupt the sensor output over time, essentially being faulty signals without a fault.
\end{enumerate}
These results highlight the challenge of false or uninformative alarms due to sensor/cable faults raised in the introduction.
Rule-based approaches and models making predictions signal-by-signal will inevitably struggle to capture the nuance of human analysis of these faults.

\subsubsection{Time Slice Prediction Excluding Source Asset}
\begin{figure}
    \vspace{-0.5cm}
    \centering
    \includegraphics[width=0.8\linewidth]{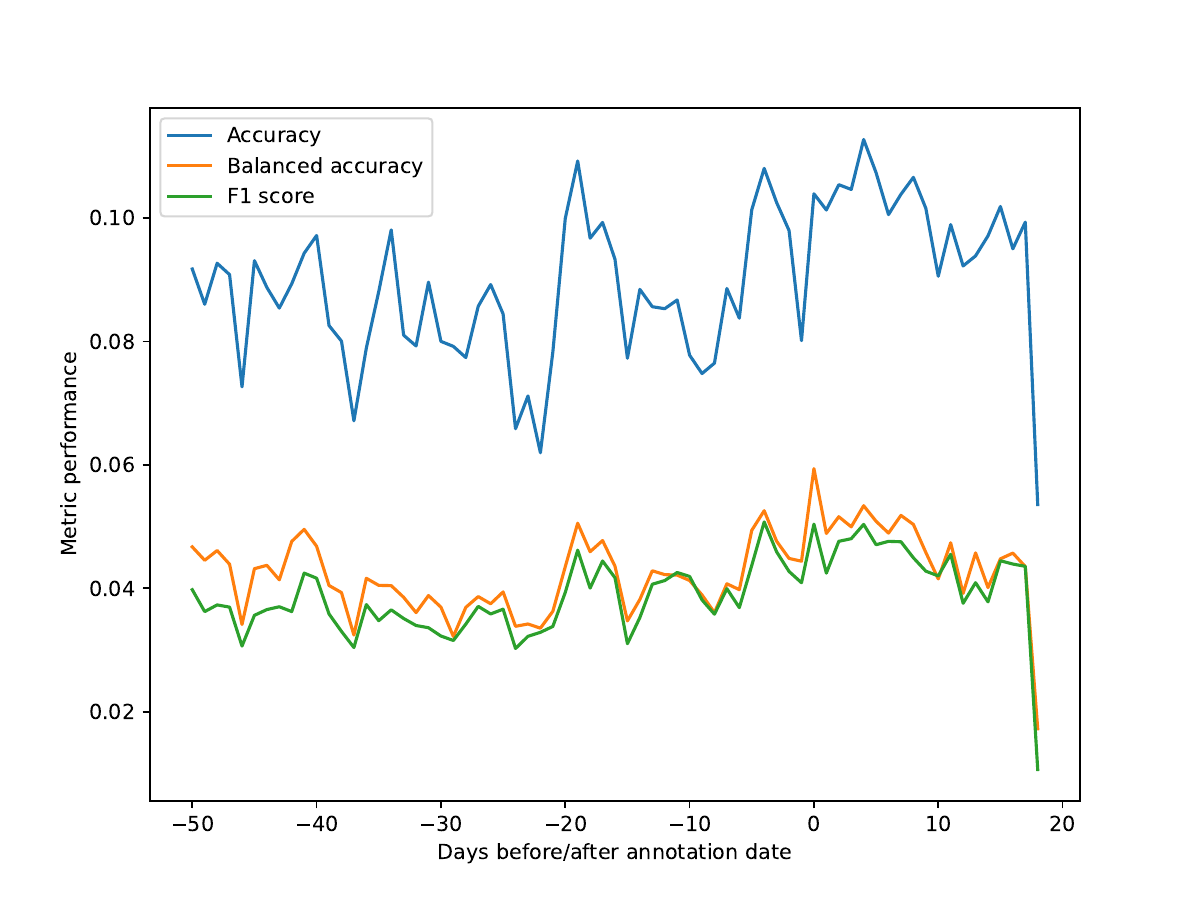}
    \vspace{-0.5cm}
    \caption{Point retrieval based on recording order distance.}
    \label{fig:excluded_asset_point_metrics}
\end{figure}

\begin{figure}
    \centering
    \includegraphics[width=0.8\linewidth]{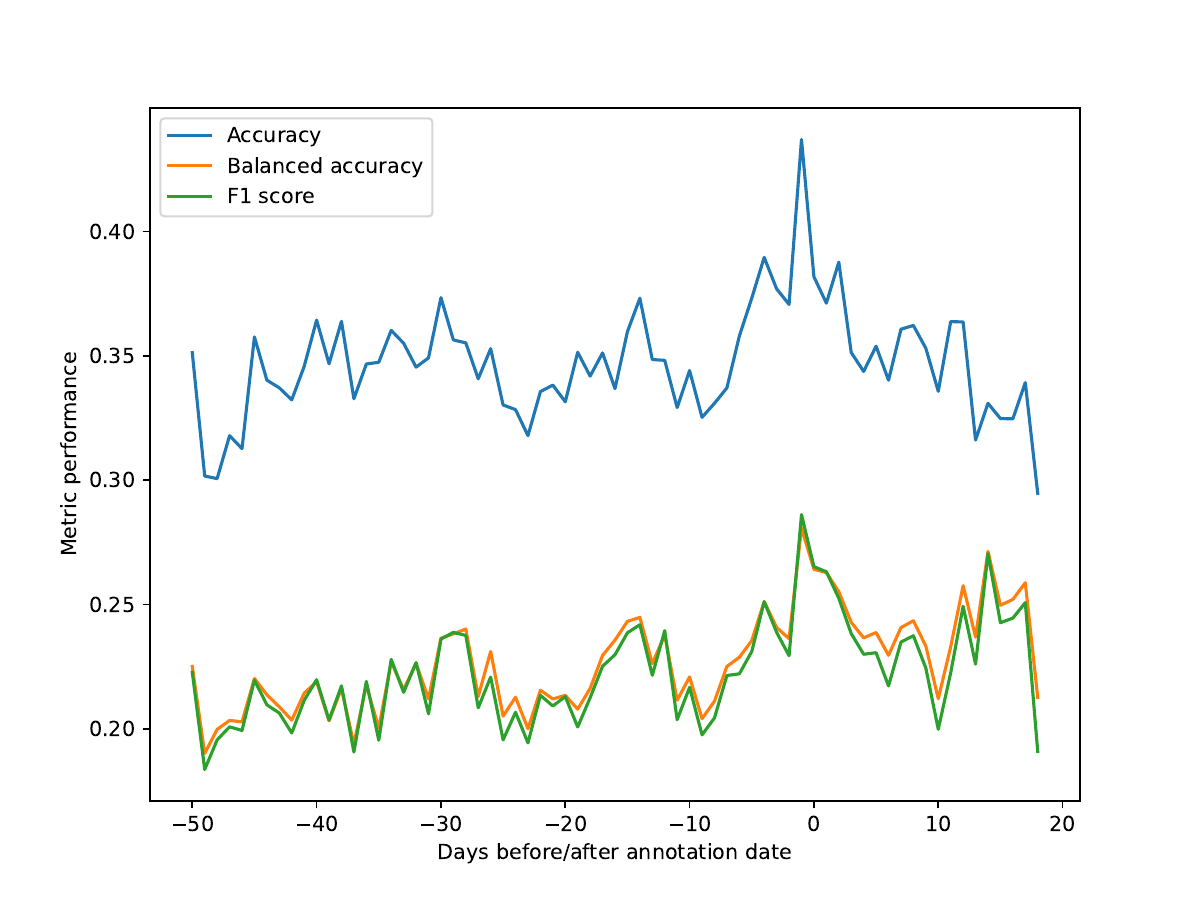}
    \vspace{-0.5cm}
    \caption{Note retrieval based on recording order distance.}
    \label{fig:excluded_asset_note_metrics}
    \vspace{-0.2cm}
\end{figure}

Figures \ref{fig:excluded_asset_point_metrics} and 
\ref{fig:excluded_asset_note_metrics} illustrate component and note retrieval based on recording order distance, as described in Section \ref{sec:rule_based_generation}, retrieved from a vector store excluding recordings from the source point and source asset respectively.
The results are computed day-by-day as for Figures \ref{fig:included_asset_point_metrics} and 
\ref{fig:included_asset_note_metrics} above, with the same target classes and rules for generation.

\paragraph{Time slices component prediction}
\begin{figure}
    \centering
    \includegraphics[width=0.8\linewidth]{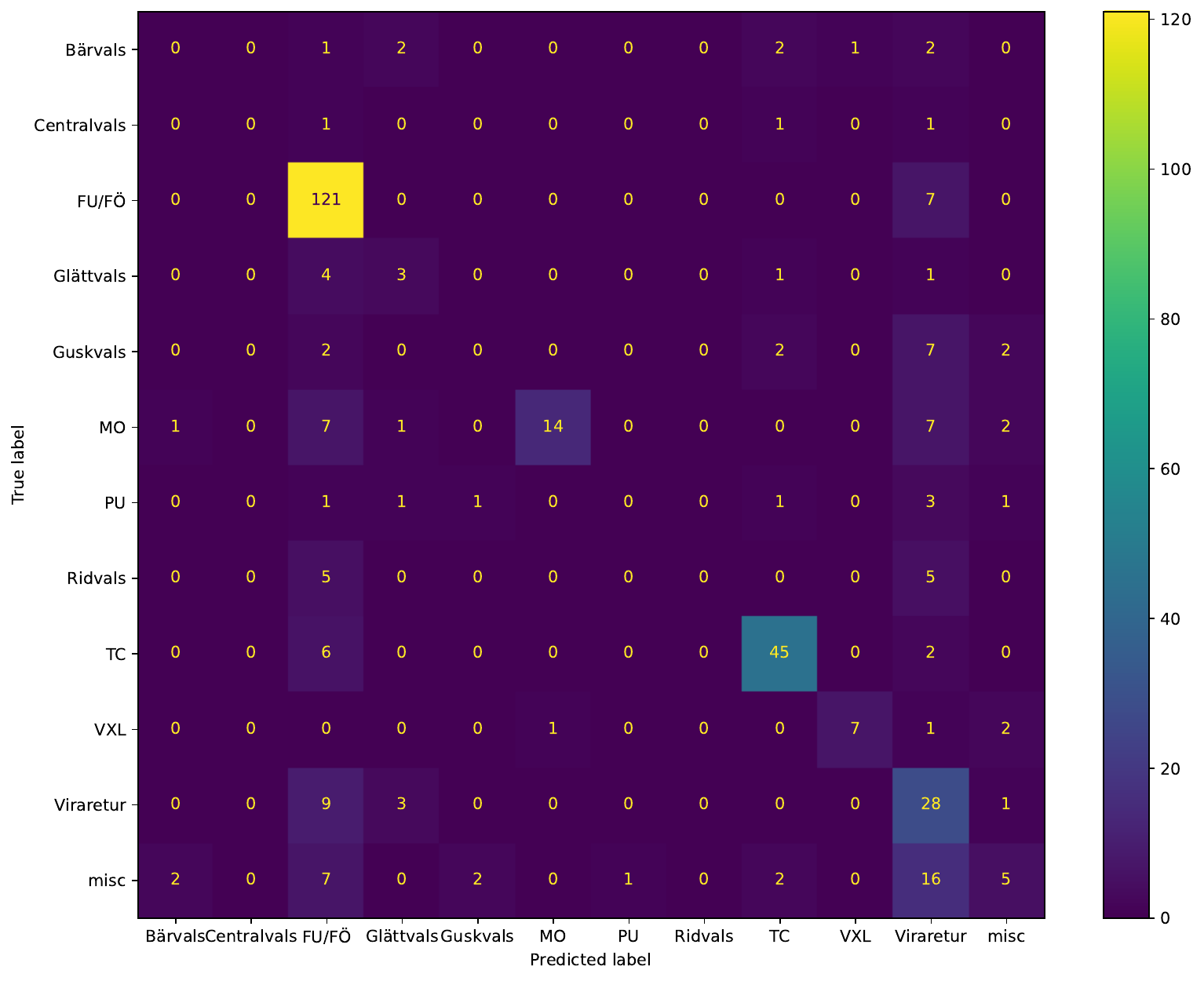}
    \caption{The asset level component prediction confusion matrix with the source point excluded from the analysis. Balanced accuracy score of 0.74; an accuracy score of 0.81; and an f1 score of 0.81.}
    \label{fig:asset_points_excluded}
\end{figure}
Component retrieval at the point-level is naturally impossible if excluding the source asset as no data in the filtered vector store will contain the target label.
Therefore, only data from the same point is excluded.
For most components, that means masking the data from the only sensor that is attached, and a correct prediction would only be possible by for instance retrieving envelope data from velocity data, which is highly unlikely.
The few correct predictions are instead indications of when multi-sensor components can retrieve data for components with multiple sensors attached.
As seen in the figure, this is the case to some extent as the accuracy is higher than zero, but further analysis would require extracting data from only components with multiple sensors.
As with the analysis with same point included shown in Figure \ref{fig:included_asset_point_metrics}, performance seems to peak a few days after the annotation date, though here all three metrics follow much closer than in the previous test.

%
%
%
%
%

\paragraph{Time slices note prediction}
The results from note retrieval with source asset excluded shown in Figure \ref{fig:excluded_asset_note_metrics} follow the same pattern as the included retrieval show in Figure \ref{fig:included_asset_note_metrics}, but with a more distinguished peak around and after the annotation date.
While the included analysis tests the mechanisms of the retrieval framework, this test examines how well the system can classify faults based on the note contents of associated signals.

\subsubsection{Asset Level Prediction with Source Included}
\label{sec:asset_included}
\begin{figure}[btt]
    \centering
    \includegraphics[width=0.8\linewidth]{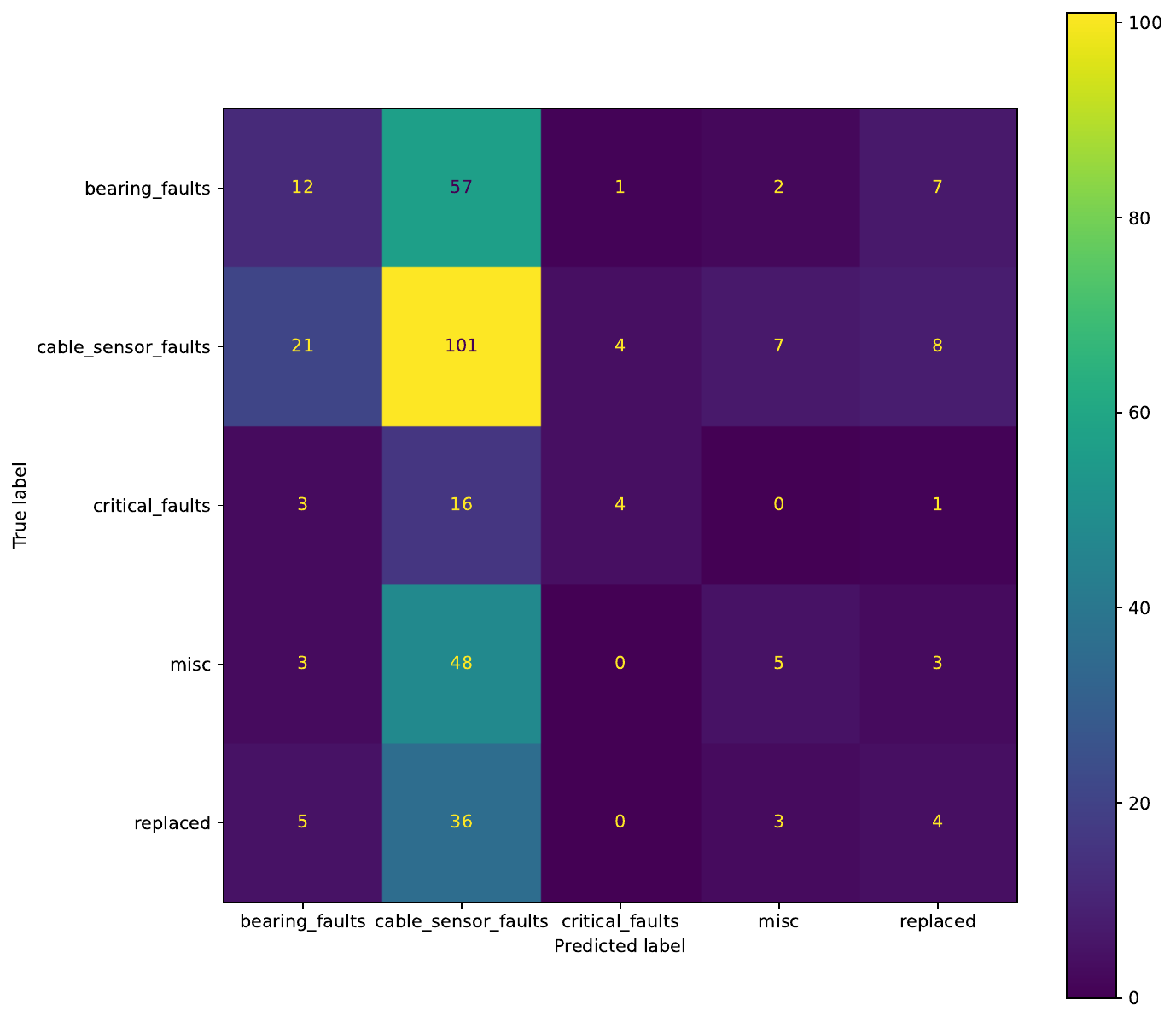}
    \caption{The asset level note prediction confusion matrix with the source asset excluded from the analysis.  Balanced accuracy score of 0.24; an accuracy score of 0.36; and an f1 score of 0.30.}
    \label{fig:asset_notes_excluded}
\end{figure}

Figures \ref{fig:asset_notes_excluded} and \ref{fig:asset_points_excluded} show the point-level note and component name retrieval confusion matrices.
The point-level prediction was achieved through majority votes across all time slices in each point, computed from the results in Figures \ref{fig:excluded_asset_note_metrics} and \ref{fig:excluded_asset_point_metrics}.

\paragraph{Asset level component prediction excluded source merged}

For the component name retrieval in Figure \ref{fig:asset_points_excluded}, the 126 machine parts were mapped down to 12 classes based on machine similarity defined with input from an on-site expert on component type, size measurements, and working environments.
This makes the results easier to interpret at the loss of resolution.
As with the example with source point included, the lowered number of classes also leads to a predictable increase in accuracy, which again primarily comes from the FU/FÖ concatenation as these machines are both numerous and virtually identical, besides placement relative to each other (U = under, Ö = above).
Unlike the previous test, this test suffers more from the imbalanced dataset, where the two most common rollers, "FU/FÖ" and "Viraretur", are by far the most commonly predicted components.
Many of the similar rollers that are much less common have zero true positives.
This is not unexpected however, as for these components only one or a few examples are available, and masking these will make predictions impossible.
Indeed, the results for masked component prediction highlight both how the noise and feature space can be shared across similar components, and how dissimilar components such as the rollers and the drying cylinders ("TC") still are significantly different, where it would be difficult to generalise from one component to the other without significant domain adaptation.

\paragraph{Asset level note prediction excluding source asset}
The confusion matrix shown in Figure \ref{fig:asset_notes_excluded} showcase the difficulties with rule-based interpretations of annotations, and the faults appear essentially randomly guessed.
This is likely due to the challenges raised in Section \ref{sec:LLM_based_generation}, where rule-based analysis fails to capture language nuance, time differences, and asset differences.
To capture the nuance of fault descriptions with regard to semantics and time-aspects, LLM agents are needed.

\subsection{LLM Agent Inference}
\label{sec:llm_agent_inference}
LLM agents, described in Section \ref{sec:agents}, were used for the same task with the same retrieval as the source asset excluded note prediction in Section \ref{sec:asset_included} above, but with LLM agents for the generation of the fault prediction.
%
%
%
As described in Section \ref{sec:LLM_based_generation}, the agent makes one prediction per asset, integrating information for all points to generate an alarm and fault assessment.
Unlike the result in Figure \ref{fig:asset_notes_excluded}, the fault prediction is based on multiple time slices, starting from the earliest slices included in the test, and moving forward one or more slices at a time, as described in Section \ref{sec:LLM_based_generation}.
The predictions were evaluated by an evaluation agent, described in Section \ref{sec:evaluation_agent}.
Two experiments were conducted, one with all annotations from the vector store, and one filtered so that only replacement annotations were included.
For the first experiment, data from 50 days until annotation date was included, while for the replacement annotations data from 0 to 20 days after the annotation was included.
%
%
The agent is tasked to predict what data from each point indicates, and then to make an asset-level prediction.
The evaluation agent is instructed to consider asset-level predictions that align with the annotation as correct.
This interaction is achieved by calling the agents with a script generating the user prompts and detecting tool calls from the agent to see when it generates an alarms, shown in the section below.
\subsubsection{Agent Instructions}
The agent is initially instructed on what task it is to solve and supplied with initial data.

\begin{lstlisting}
Your job is to make fault diagnosis predictions based on data from a condition monitoring (CM) database.
You are currently analysing signals from the asset at machine path {prev_asset_path}, with point names and sensor types {input_dict}, where the positions of the point names and sensor types correspond to the positions of the incoming recording slices.
The recording slices will feature one set of recordings per point name and sensor type, and iterate forward in time based on your requests.
Each recording features a chunk id and datetime value at the start of the list, which is the input chunk id with the date of the recording.
Each input chunk is associated with five MindRAG chunks, that are obtained by finding the most similar recordings from the CM database.
These MindRAG chunks have associated annotations that describe fault properties of these chunks, and time deltas which describe the time distance between the MindRAG chunk and its annotation.

If you are not ready to make a prediction, reply with "CONTINUE" and the user will provide another set of recording slices.
If you ask for more slices than are available, the user will break the loop and you may proceed with your analysis.

Finally, respond with your analysis with the "reply_with_precition" tool.

The first input and MindRAG chunks are: {input_chunk_list[0]}
\end{lstlisting}

If the agent requests more data, it is given the input:
\begin{lstlisting}
    The next slices are {input_chunk_slices}.
\end{lstlisting}

If the agent runs out of input, it is informed of this, and instructed on how to proceed:
\begin{lstlisting}
You have run out of chunks. You MUST make a prediction with what you have using "reply_with_prediction" or you will not progress. Proceed through "reply_with_prediction", and if you have not detected any faults, please explain why
\end{lstlisting}

The evaluation agent is then instructed to analyse the prediction related to the true note and write to file.
The agent also receives information about the number of days between the annotation and the last received recording, which influences the evaluation and is stored to file.

\begin{lstlisting}
The true note is {prev_note['noteComment']} at a time delta of {days_between_note_and_recording} days. The prediction made was:
                    {agent_prediction}. Was the analysis correct? Write it to file with the "write_prediction" tool.
\end{lstlisting}

For the main run with data leading up to the annotation date, 38 assessments were wrong, 92 were correct of which 20 were partially correct (according to the evaluation agent), for an accuracy of 71\% and 65\% respectively.
Of these, 77, or 59\%, were voluntary exits, with an accuracy of 75\%.
For the 63 involuntary exits, the accuracy was 67\%.
Assessments were considered partially correct if the CM agent was correct, but played it safe by mentioning more than one type of fault in the assessment.
This is a significant improvement compared to the performance shown in Figure \ref{fig:asset_notes_excluded}, despite having the same retrieval mechanism and data.

The LLM outputs are time-consuming to evaluate on a case-by-case basis, and contain multiple mentions of proprietary information.
Therefore, the first ten cases from the vector store have been post-processed, where point names, analyst names, and significant dates have been removed.
These cases were analysed with the help of an analyst using the source CMS to verify results, and are shown in Section \ref{appendix:MIRAGE_runs}, and are summarised below.

The first case did lead to a voluntary exit, and the agent generated an alarm.
The agent correctly identifies a sensor fault, which was verified as correct by the analyst.

The second case did not lead to a voluntary exit.
The agent correctly predicted the annotation in two points, but was not sure what fault had motivated the annotation.
Hence, it is partially correct, but failed to generate an alarm in time.

The third case did not lead to a voluntary exit, and the agent comments that there is ``nothing critical at this moment", but that if there is a fault it could be a sensor fault.
The agent was thus forced to generate a prediction by the experimental framework, but had not detected a fault.
Analyst information revealed that this maintenance action is a regularly scheduled event, so nothing in the signal space leading up the action indicated a bearing fault, while a few signals did indeed hint at signal corruption.
The evaluation agent correctly marked this prediction as wrong based on the information it had access to, while the analyst considered the prediction to be correct.

The fourth case was the same as the third, for a similar component with the same maintenance action.

The fifth, sixth, and seventh cases were correct identifications of sensor faults, though the agent did not opt to generate alarms.
The eight and tenth cases deal with voluntary exits to predict BPFO and BPFI, respectively.
In both cases, the agent correctly predicts the annotation, before the annotation date.

The ninth case features a voluntary exit to make a sensor fault prediction.
Sensor fault features were present in the analysed signals, though not frequent enough to warrant an alarm.
The agent mentions the bearing replacement as well, and is thus partially correct, but would likely not have exited to generate that alarm were it not for the sensor faults.

In general, the agent is more careful with sensor and cable fault alarms, which is due to its prompts.
The analyst has been instructed to await clear indications of sensor faults before generating alarms, which is likely why it is hesitant to exit the loop for these faults.
Accurate fault prediction, even when not voluntarily exiting, is a valuable tool when coupled with existing high-sensitivity alarm generation methods.
In the case of bearing faults in the other predictions, the agent more often exits early, but across all predictions correct voluntary exits are the most common evaluation result.

The general assessment of the analyst was that, with the alarms generated and faults predicted, it would be a valuable addition to existing systems, especially when the agent motivated analysis for trends in each point.
This behaviour is currently due to the prompts of the evaluation agent tools, but can be transformed to a forced response template, as the ``final\_answer" tool of the main agent.

%


%
Testing how prone to false positives the agent is, the agent was also given only data from after a replacement was included, running from the day after the replacement date to the end of the data extraction, which was between 16-18 days after the annotation date depending on how close to the dataset end date the replacement annotation was made.
The first ten of these are processed and presented in Section \ref{appendix:MIRAGE_runs}
Of the twenty cases, two were cancelled before running out and thus false positives.
Of these, one exit correctly identified that no alarm was to be generated (prediction 1), and one exit suspected sensor faults in two points and imbalance in one point.
The first exit would now have generated an alarm, and it is not clear why the agent elected to exit.
The second exit is wrong based on the true note, which describes that a roller has been replaced, and the MindRAG retrieval has retrieved irrelevant signals.
The exit is thus warranted, while the underlying data is wrong.

Of the predictions made, regardless of voluntary exit, four were incorrect.
One prediction erroneously predicts a sensor or cable fault when the true note was that a roller had been replaced.
The second error was the aforementioned early exit, the only true false alarm.
The third error was due to the agent prediction a sensor fault on one point, while the true note mentioned a roller replacement.
The agent also mentions bearing replacements, but forced to make an assessment at the end of the run, it predicts a sensor fault.
The final error was also a roller replacement note with a sensor fault prediction.

As with existing systems, sensor and cable faults are the primary cause for false alarms.
However, the agent has the benefit of being promptable to wait for clear indications before generating an alarm, while being more sensitive to more critical faults. 
Thus, though the agent suspects cable or sensor faults for multiple points, only one suspicion actually generated a false alarm.

\section{Discussion}
\label{PaperE:sec:discussion}
The results presented in this work indicate that the MindRAG approach is a viable complement to traditional supervised or unsupervised transfer learning approaches to improve automated fault diagnosis and historic insights in condition monitoring.
In particular, it is a modular framework that can be implemented on annotated CM datasets, without requiring time-consuming pre-training or fine-tuning of ML models.

The results from the qualitative text-to-text knowledge retrieval, discussed in Section \ref{sec:knowledge_retrieval_results} show that the framework can generate accurate responses, grounded in retrieved CM data, to questions provided by analysts.
The model does not reason perfectly, and fails to include parts of its knowledge base in one example, but does not indicate hallucination or overly helpful responses to cover for missing data.
Based on analyst feedback, the knowledge retrieval module would be a valuable addition to existing CMS if well integrated.
Updating and storing alarms is not explicitly shown, as the functionality is the same as TaskmAIstro from LangChain, and thus not novel from a research perspective.

The results from the component predictions in Figures \ref{fig:included_asset_point_metrics} and \ref{fig:asset_points_included} indicate that the MindRAG framework can be used to predict the source components of unidentified signals within a machine.
Figures \ref{fig:excluded_asset_point_metrics}, and \ref{fig:asset_points_excluded} indicate that this approach can be extended to new installations, i.e. recording with no historic signal representations, for some components.
The results also indicate the presence of heterogenous noise, or healthy ``features", which facilitate such component retrieval.
These results imply the difficulties involved in training machine learning models on CM data;
if, for instance, a component has one fault in a training set, a machine learning model might learn the noise characteristics, the fault characteristics, or both, making generalisation between components difficult.
%
%

The rule-based note prediction presented in Figures \ref{fig:included_asset_note_metrics}, \ref{fig:asset_notes_included}, \ref{fig:excluded_asset_note_metrics}, \ref{fig:asset_notes_excluded} 
shows the best performance for slices around the annotation date, which is an expected outcome as those recordings warranted the annotation.
This is used to define the time span for recordings to include in the vector store for LLM-based inference to $\pm$ ten days around the annotation date.
These results also warrant the use of more expensive LLMs to proceed in the testing, due to the poor performance of asset-excluded retrieval shown in Figures \ref{fig:excluded_asset_note_metrics}, \ref{fig:asset_notes_excluded}.
Combined with the illustrated retrieval shown in Figures \ref{fig:input_1}, \ref{fig:output_1}, \ref{fig:input_2}, and \ref{fig:output_2}, this also facilitates testing new retrieval mechanism or rules without the complexity of LLM-generated responses.
While LLM performance is expected to exceed rule-based inference in heterogenous environments where machine-specific contextual knowledge is required, such models are important to develop with regard to environmental impacts of the research, and less computationally demanding retrieval experiments can facilitate faster and more efficient optimisation of vector store and retrieval algorithms.
%

%

The LLM-generation results discussed in Section \ref{sec:llm_agent_inference} and presented in Section \ref{appendix:MIRAGE_runs} highlight the full capacity of the MindRAG framework.
In the first experiment, with signals leading up the annotation date, the agent shows the ability to reason about faults in industrial databases based on the annotated retrieved signals.
The accuracy for fault prediction was 71\%, as measured by the LLM-evaluated similarity between the prediction and true annotation, which is superior to the 36\% asset-level accuracy of rule-based generation, shown in Figure \ref{fig:asset_notes_excluded}.
However, accuracy could likely be further improved by improved vector store processing, retrieval functions, and generative models, as will be discussed in Section \ref{sec:hyperparameters}.
Analyst evaluation, though limited in time, resulted in better performance for the first ten cases than LLM evaluation based on machine-specific context knowledge.
Further research into both automated and manual evaluations are therefore necessary to more accurately represent performance in a given experimental set-up.

The agent showed better performance for voluntary exits (alarm generation) of the experiment loop, at 75\% compared to 67\% for forced predictions.
This implies that the performance correlates with the agent's decision to generate an alarm.
Perplexity, i.e. output certainty, could potentially be leveraged to further improve performance for voluntary alarm generation.
It is not known how performance would change if the agent could retrieve signals until all exits were voluntary, though it would likely improve based on the presented results.
However, running the experiment past the annotation date risks indirect data leakage, where sudden changes in associated signals likely indicate a maintenance action, which is information that is difficult to mask but useful for fault predictions despite having no relevance for industrial CM (the goal is to predict when a component requires replacement, not to identify that a component has been replaced after the fact).
The vector store could be filtered to exclude these components, and future experiments should investigate more experimental set-ups.

In the second experiment, the agent generated one false alarm and one redundant alarm out of 20 tests with mostly healthy signals.
This provides a starting point for LLM-based alarm generation, where 5\% false alarm rate is significantly better than existing systems.
However, the false alarm rate will also depend on the experimental set-up.
If the agents ran for longer durations on healthy data, the false alarm rate would increase.
This is true for existing system as well, and it is not clear how they would perform if evaluated on number of alarms compared to number of analysed healthy assets.
Over the entire annotated dataset in the first experiment, the agent generated meaningful alarms 75\% of the time, which again is considerably better than existing systems.
Existing systems are evaluated by analysts over both healthy and unhealthy data, including unannotated data, making direct comparisons difficult, and existing systems would likely fare better when evaluated on data with more frequent faults, as in LLM experiment one.

Nevertheless, initial tests show an approach that can generate meaningful alarms and human feedback, useable with text-queries and capable of providing system insight and knowledge retrieval from historic cases.
However, it is a prototype framework with many potential aspects of improvement, where each part of the framework can be tuned and updated.

\subsection{Tuning the Framework}
\label{sec:hyperparameters}
The current agent framework can be altered with regard to vector store generation, retrieval function, generation approach, and experimental setup, to further tune the performance.

\subsubsection{Vector Store Contents}
The vector store contents will have a direct impact on what data can be retrieved, and thus what tasks can be solved.
An important consideration is which signals and/or annotations to include.
Some annotations describe routine actions that have no relation to the signal at any particular point, which are virtually impossible to predict with normal fault diagnosis model.
Such cases are included in the testing to streamline the experiments and prevent designing experiment to ``game" performance, which likely would improve if such data was excluded from both the input and the vector store.
Timedeltas are another important design decision;
it is computationally inefficient to include all data in the vector store for MindRAG retrieval, as the amount of vector computations would increase considerably, while likely yielding small changes in performance.
We have chosen to use data from around the annotation date based on test over time slices, but the exact optimal time span is not known, and likely varies for different faults/annotations.
This question can be relevant to reconsider when investigating other methods for processing of industry data.

Unannotated data was excluded from the tests performed in this work.
How unannotated data would impact performance is one major question that needs to be answered before the performance is evaluated in a field-like scenario.
However, the lack of annotations mean that unannotated data is difficult to evaluate (except for component predictions, which only require asset and point information), as a lack of annotations does not necessarily indicate a lack of faults.
For instance, low levels of bearing fault indications can be present in unannotated components, and sensor and cable fault features can appear sporadically without warranting an annotation and maintenance action.
Using the agents on all unannotated data would therefore require significant effort by expert analysts to assist with verification, which implies that further research in collaboration with experienced analysts is required before CM AI agents can be deployed in an industrial setting.

An important aspect of fault diagnosis in the field, that is difficult to immitate in lab environments, is how the environment changes over time.
If the agent framework is integrated with an existing CMS and operates for ten years, knowledge retrieval from old cases could become more relevant as without the agent these cases are likely forgotten with current practices.
However, for signal retrieval, would examples from ten years ago still be relevant, or should the vector store be filtered to include only data from only more recent years?
This is not something that can be examined with the current vector store, but something that can be investigated when more data has been collected and processed.

%

\subsubsection{Retrieval Mechanisms}
The retrieval mechanisms have a large impact on downstream tasks,e as these tasks can only be addressed if correct data is retrieved.
For historic insights, the goal is to find all relevant documents, as in normal RAG, but with more filters and distance metrics to consider.
Unlike normal RAG, which mainly is used to find the \textit{most} relevant documents, industrial RAG sometimes requires \textit{all} relevant documents, regardless of the value of the $top\_k$ hyperparameter for number of matches returned.
Therefore, the agents have access to tools needed to directly read all annotations from a filtered subset of the vector store, or to read all annotations featuring certain keywords.
The query and retrieval embeddings are also augmented to emphasise technical terms, e.g., adding ``sensorfel" to annotations or queries featuring ``givarfel", which is a far less commonly used term for sensor faults in Swedish.

For signal retrieval, there are many viable ways to measure similarity between two recording chunks, especially if trend and bias values are considered.
These include: incorporating statistical properties; order analysis, as used in this work; physics-based models such as kinematic condition indicators \cite{randall_bearing_diagnostics}; and learning based approaches as used in intelligent fault diagnosis approaches \cite{tang_DL_CM}.
Retrieval can also be conceptualised at the point-level, i.e., integrating multiple recording chunks per retrieval call.
For example, if three consecutive chunks are used as inputs, the system could search for points with the highest sum of scores for these three chunks, e.g., over ten consecutive chunks in the target point.
This approach could also be extended to the asset level, searching for assets with points with similar recordings in the same time span.
However, it is not clear how well this would work due to the many different configurations of points present in different assets.

The retrieval mechanisms based on order analysis are a good starting point for the MindRAG approach, but will likely benefit from more nuanced approaches.
For example, the speed during signal sampling can vary, and resampling will inevitably introduce errors.
Therefore, instead of simple vector multiplication, a more nuanced distance metric that compares the closest value within a small span could be used, to build more robust retrieval.
Furthermore, to align retrieved data more accurately with fault features, noise reduction techniques could be used.
Statistical properties such as kurtosis and skew were also investigated, but not included in the final experiments as order analysis appeared to suffice, and measuring distance between these values was ambiguous.
Ideally, condition indicators such as BPFO and BPFI characteristic frequencies would also be integrated, making for a more nuanced retrieval mechanism.
Finally, ML methods could be used to measure the distance between two recordings.
As this process is repeated for each input with almost each recording in the filtered vector store, such models must be fast and efficient, or scaling would be difficult, even if performance can be improved.

\subsubsection{Generation Mechanisms}
The generation part of MindRAG, handled by LLM agents, can be altered by using different LLMs to power the agents; updating tools, tool instructions, and agent prompts; and adding information to the internal knowledge bases maintained by the agents.
As LLMs improve, there is both the possibility for better and cheaper analysis of retrieved contents, as well as easier local deployments of LLMs for improved data privacy and fine-tuning possibilities.
The development of general LLM agents is beyond the scope of this work, but in theory a reinforcement learning from human feedback (RLHF) framework could be developed to optimise the LLM during interactions with users \cite{ouyang2022training}.
As discussed in Section \ref{sec:llm_agent_inference}, the true annotation can be used to evaluate LLM performance for fault predictions.
The same approach can be generalised to include human feedback of model fault predictions and generated alarms, which can be quantized to numerical feedback for RLHF optimisation, which would facilitate a self-improving integrated industry AI assistant.

Besides learning-based optimisation, the agents can be updated with better tools and instructions by a designer, or self-update by writing to knowledge bases based on human feedback.
In the present implementation, knowledge bases illustrate potential use-cases, and are accessed mainly on demand,.
With continuous user interactions, the agent could instead be forced to access them before calling certain tools, e.g., the ``compute changes" tool.
An interesting prospect is if the agent could ``self-learn" by independently writing reflections to knowledge bases, or through LLM evaluations of user interactions for RLHF, in a ``human- and AI-in-the-loop" approach.

\subsubsection{Experimental Setup}
The experimental setup is meant to simulate agent assistance in industry workflows.
Depending on how tests are conducted, performance metrics such as estimated accuracy can be affected.
For instance, if tests were to be conduced only on data with known clear fault features, performance would likely improve compared to the testing performed in this work.
The objective of the tests can also be altered to change performance.
For example, the agents are currently instructed to independently exit an analysis loop and generate an alarm, and evaluate if the alarm is correct and generated in time.
However, a simpler approach is to feed data leading up to or even beyond the annotation date, tasking the agent to estimate the contents of the annotation.
In this case, the agent is indirectly informed of when the real annotation was written, which can guide analysis towards associated annotations closer to that date.
Evaluation could also be changed to ``correct" if any point assessment corresponds to the asset annotation, which would evaluate if the agent is correct at the point-level, but not correspond as well to how an agent would operate in a field environment.

\subsection{Industrial Applications}

The LLM agents can be integrated in an industrial CM workflow for four different, non-conflicting levels of assistance:
\begin{enumerate}
    \item knowledge retrieval assistance and alarm/report management,
    \item on-demand fault diagnosis,
    \item automated alarm generation and verification for analyst confirmation,
    \item automated maintenance reports for well-identified and repetitive faults.
\end{enumerate}

The first level requires no signal-based analysis to implement, and is the most mature.
As described in Section \ref{sec:knowledge_retrieval_results}, our approach can unlock valuable insights and upskilling, as described by industry expert analysis of the outputs.
This level of integration would likely be implemented as an extension to existing CMS systems to help navigate the underlying data and answer questions as those shown in Section \ref{sec:knowledge_retrieval_results}.

On-demand fault diagnosis would likely be implemented in a CMS in a similar way, by performing analysis based on an explicit user queries to retrieve similar cases and use those for fault assessments, as shown in Section \ref{sec:signal_retrieval_results}.
Besides fault diagnosis, signal retrieval could be valuable by identifying historic cases with similar signal developments, hence using the agent for retrieval but forming the assessment with human analysis.
This level of integration could also be warranted to investigate field applications of the proposed framework.

Levels three and four feature more autonomous integration, with both approaches requiring that the agents run in parallel with human operations, analysing data independent of user prompts, similar to the experimental framework shown in Sections \ref{sec:LLM_based_generation} and \ref{sec:llm_agent_inference}.

Alarm generation/verification implements the ability of the agent to independently alert human analysis, as current systems operate.
The agent could also be tasked with verifying generated alarms to, for instance, order them in levels of criticality, or discard redundant alarms for already detected faults, e.g., cable and sensor faults.
This level of integration would be interesting for alarm verification, but further testing would be required to estimate how robust the agents are in a field setting.

Finally, the fourth level works towards automation and would feature automated maintenance reports for repetitive faults, e.g., cable and sensor faults, so that analysts can focus on more meaningful and nuanced analysis of component degradations and maintenance practices to prevent faults and unplanned stops.
However, this level of integration would only be acceptable after appropriate field testing, where analysts can verify proposed maintenance reports directly.

\subsection{Limitations}
As already highlighted, the experiments performed to obtain qualitative and quantitive results are subject to some limitations, in particular with LLM generation where output is more time consuming and costly to produce.
Due to the non-deterministic nature of LLM generation, the results are not exactly replicable.
An objective evaluation of LLM output is difficult and has not been performed.
Examples highlighting different agent functionalities are also limited, and though more examples could be presented, the lack of an objective metric makes presentation and analysis time- and space consuming.
Structured external expert reviews of agent outputs is a possible proxy, which is a natural albeit costly potential development of the current experimental framework. 

The lack of labels and limited number of annotations implies that general quantitative evaluation is difficult, especially since there is no related work to compare with, neither on this data nor any process industry field dataset of comparable complexity.
We have used annotations and LLM evaluation to approximate true labels, but testing and analysing the outputs is time consuming.
Presenting the underlying LLM output that generated the quantitative results is also difficult.
The MindRAG example showcased in Section \ref{appendix:MIRAGE_runs} in the Appendix contains the first ten evaluation agent results per run, where each result is based on around 20 sub-evaluations by the CM agent per final assessment.
These results from the evaluation agent are time-consuming to present and verify, and presenting the output from the CM agent reasoning even more so.

Further experimental validation could be obtained in the form of repeated experiments with external reviews based on language properties, e.g., usefulness, truthfulness, and clarity, to generate some quantitive assessments of the text-based retrieval.
External validation of diagnosis results, and structured comparisons with existing CMS, would also give more objective quantitative evaluation.
A structured evaluation of the tuneable hyperparameters, e.g., through an ablation study, would also be valuable.

\subsection{Future work}
As explained in Section \ref{sec:hyperparameters}, changes or improvements can be made in all four areas of contribution.
The vector store could integrate more data, retrieval and generation mechanisms could be altered, and the experimental framework could be expanded.
Of these, the retrieval mechanisms appear to have high potential for future research, as many methods of representing and comparing signals already exist and could be tested with the vector store, while the generation and experimental framework could be left virtually unchanged.
We have identified three main approaches to improving the MindRAG framework:
\begin{enumerate}
    \item expanding on current retrieval distance metrics by integrating statistical properties and feature extraction models,
    \item integrating pre-trained signal encoders,
    \item training the retrieval mechanism directly on the vector store.
\end{enumerate}
The first approach expands on current systems, and deals with questions such as:
\textit{``which statistical features should be integrated, and how?"}, \textit{``should signals be de-noised?"}, and \textit{``Are only the peaks relevant when retrieving similar faults?"}.
The second approach is a natural question for researchers in the IFD field, as the framework offers an end-to-end setup to evaluate (albeit weakly) if one IFD model for signal similarity performs better than the other.
The third approach hints at an interesting opportunity to integrate technical language supervision \cite{lowenmark2021technical}.
A technical language supervision model pre-trained with contrastive learning  on the vector store data could use its joint embedding space to infer both signal-annotation similarities, but also signal-signal similarities.
Through either RLHF or labels written by the evaluation agent, the contrastive model could then be fine-tuned for fault diagnosis with the experimental framework.


\bibliographystyle{plain}
\bibliography{MindRAG}

\begin{thebibliography}{10}

\bibitem{abootorabi2025askmodalitycomprehensivesurvey}
Mohammad~Mahdi Abootorabi, Amirhosein Zobeiri, Mahdi Dehghani, Mohammadali Mohammadkhani, Bardia Mohammadi, Omid Ghahroodi, Mahdieh~Soleymani Baghshah, and Ehsaneddin Asgari.
\newblock Ask in any modality: A comprehensive survey on multimodal retrieval-augmented generation.
\newblock {\em arXiv preprint arXiv:2502.08826}, 2025.

\bibitem{agentic_AI_review}
Deepak~Bhaskar Acharya, Karthigeyan Kuppan, and B.~Divya.
\newblock Agentic ai: Autonomous intelligence for complex goals—a comprehensive survey.
\newblock {\em IEEE Access}, 13:18912--18936, 2025.

\bibitem{Alsaif2024Electronics}
Khalid~M. Alsaif, Abdullah~A. Albeshri, Mohamed~A. Khemakhem, and Fadi~E. Eassa.
\newblock Multimodal large language model-based fault detection and diagnosis in context of industry 4.0.
\newblock {\em Electronics}, 13(24):4912, 2024.

\bibitem{deep_transfer_learning_review}
Md~Roman Bhuiyan and Jia Uddin.
\newblock Deep transfer learning models for industrial fault diagnosis using vibration and acoustic sensors data: A review.
\newblock {\em Vibration}, 6(1):218 – 238, 2023.

\bibitem{bleicher1991konzept}
Knut Bleicher et~al.
\newblock {\em Das Konzept integriertes management}.
\newblock Campus Verlag Frankfurt, 1991.

\bibitem{Borgeaud2022}
Sebastian Borgeaud, Arthur Mensch, Jordan Hoffmann, Trevor Cai, Eliza Rutherford, Katie Millican, George van~den Driessche, Jean-Baptiste Lespiau, Bogdan Damoc, Aidan Clark, Diego de~Las~Casas, Amy Guy, Jacob Menick, Roman Ring, Tom Hennigan, Max Cain, Anna Ree, Laurie Chen, et~al.
\newblock Improving language models by retrieving from trillions of tokens.
\newblock In {\em Proceedings of the 39th International Conference on Machine Learning (ICML)}, 2022.

\bibitem{brown2020language}
Tom Brown, Benjamin Mann, Nick Ryder, Melanie Subbiah, Jared~D Kaplan, Prafulla Dhariwal, Arvind Neelakantan, Pranav Shyam, Girish Sastry, Amanda Askell, et~al.
\newblock Language models are few-shot learners.
\newblock {\em Advances in neural information processing systems}, 33:1877--1901, 2020.

\bibitem{brundage2021technical}
Michael~P Brundage, Thurston Sexton, Melinda Hodkiewicz, Alden Dima, and Sarah Lukens.
\newblock Technical language processing: Unlocking maintenance knowledge.
\newblock {\em Manufacturing Letters}, 27:42--46, 2021.

\bibitem{technical_language_processing}
Michael~P. Brundage, Thurston Sexton, Melinda Hodkiewicz, Alden Dima, and Sarah Lukens.
\newblock Technical language processing: Unlocking maintenance knowledge.
\newblock {\em Manufacturing Letters}, 27:42--46, 2021.

\bibitem{cabello2024megmedicalknowledgeaugmentedlarge}
Laura Cabello, Carmen Martin-Turrero, Uchenna Akujuobi, Anders S{\o}gaard, and Carlos Bobed.
\newblock Meg: Medical knowledge-augmented large language models for question answering.
\newblock {\em arXiv preprint arXiv:2411.03883}, 2024.

\bibitem{carragher2025quantifyingmemorizationretrieverperformance}
Peter Carragher, Abhinand Jha, R~Raghav, and Kathleen~M Carley.
\newblock Quantifying memorization and retriever performance in retrieval-augmented vision-language models.
\newblock {\em arXiv preprint arXiv:2502.13836}, 2025.

\bibitem{chaselangchain}
Harrison Chase.
\newblock Langchain, 2022.
\newblock {\em \url{https://github.com/hwchase17/langchain}}, 2022.

\bibitem{chen2025faultgptindustrialfaultdiagnosis}
Jiao Chen, Ruyi Huang, Zuohong Lv, Jianhua Tang, and Weihua Li.
\newblock Faultgpt: Industrial fault diagnosis question answering system by vision language models.
\newblock {\em arXiv preprint arXiv:2502.15481}, 2025.

\bibitem{transfer_learning_RUL_review}
Jiaxian Chen, Ruyi Huang, Zhuyun Chen, Wentao Mao, and Weihua Li.
\newblock Transfer learning algorithms for bearing remaining useful life prediction: A comprehensive review from an industrial application perspective.
\newblock {\em Mechanical Systems and Signal Processing}, 193:110239, 2023.

\bibitem{deep_transfer_learning_review_2}
Xiaohan Chen, Rui Yang, Yihao Xue, Mengjie Huang, Roberto Ferrero, and Zidong Wang.
\newblock Deep transfer learning for bearing fault diagnosis: A systematic review since 2016.
\newblock {\em IEEE Transactions on Instrumentation and Measurement}, 72:1--21, 2023.

\bibitem{sergio_dataset}
Sergio~Martin del Campo, Fredrik Sandin, and Daniel Strömbergsson.
\newblock Dictionary learning approach to monitoring of wind turbine drivetrain bearings.
\newblock {\em International Journal of Computational Intelligence Systems}, 14:106--121, 2020.

\bibitem{dhuliawala2023chain}
Shehzaad Dhuliawala, Mojtaba Komeili, Jing Xu, Roberta Raileanu, Xian Li, Asli Celikyilmaz, and Jason Weston.
\newblock Chain-of-verification reduces hallucination in large language models.
\newblock {\em arXiv preprint arXiv:2309.11495}, 2023.

\bibitem{ding2024rablipmultimodaladaptiveretrievalaugmented}
Muhe Ding, Yang Ma, Pengda Qin, Jianlong Wu, Yuhong Li, and Liqiang Nie.
\newblock Ra-blip: Multimodal adaptive retrieval-augmented bootstrapping language-image pre-training, 2024.

\bibitem{transfer_learning_review}
Mohammed Hakim, Abdoulhdi A.~Borhana Omran, Ali~Najah Ahmed, Muhannad Al-Waily, and Abdallah Abdellatif.
\newblock A systematic review of rolling bearing fault diagnoses based on deep learning and transfer learning: Taxonomy, overview, application, open challenges, weaknesses and recommendations.
\newblock {\em Ain Shams Engineering Journal}, 14(4):101945, 2023.

\bibitem{medical_MMRAG}
Iryna Hartsock and Ghulam Rasool.
\newblock Vision-language models for medical report generation and visual question answering: a review.
\newblock {\em Frontiers in Artificial Intelligence}, 7, 2024.

\bibitem{HEREDIAALVARO2025103007}
José~Antonio {Heredia Álvaro} and Javier~González Barreda.
\newblock An advanced retrieval-augmented generation system for manufacturing quality control.
\newblock {\em Advanced Engineering Informatics}, 64:103007, 2025.

\bibitem{izacard2020leveraging}
Gautier Izacard and Edouard Grave.
\newblock Leveraging passage retrieval with generative models for open domain question answering.
\newblock {\em arXiv preprint arXiv:2007.01282}, 2020.

\bibitem{Izacard2022}
Gautier Izacard, Edouard Grave, and Armand Joulin.
\newblock Few-shot learning with retrieval augmented language models.
\newblock {\em arXiv preprint arXiv:2208.03299}, 2022.

\bibitem{Jardine2006}
A.~K.~S. Jardine, D.~Lin, and D.~Banjevic.
\newblock A review on machinery diagnostics and prognostics implementing condition-based maintenance.
\newblock {\em Mechanical Systems and Signal Processing}, 20(7):1483--1510, 2006.

\bibitem{chatcnc}
Jurim Jeon, Yuseop Sim, Hojun Lee, Changheon Han, Dongjun Yun, Eunseob Kim, Shreya~Laxmi Nagendra, Martin~B.G. Jun, Yangjin Kim, Sang~Won Lee, and Jiho Lee.
\newblock Chatcnc: Conversational machine monitoring via large language model and real-time data retrieval augmented generation.
\newblock {\em Journal of Manufacturing Systems}, 79:504--514, 2025.

\bibitem{Karpukhin2020}
Vladimir Karpukhin, Barlas Oguz, Sewon Min, Patrick Lewis, Ledell Wu, Sergey Edunov, Danqi Chen, and Wen-tau Yih.
\newblock Dense passage retrieval for open-domain question answering.
\newblock In {\em Proceedings of the 2020 Conference on Empirical Methods in Natural Language Processing (EMNLP)}, pages 6769--6781, 2020.

\bibitem{Kim2021vilt}
Wonjae Kim, Bokyung Son, and Ildoo Kim.
\newblock {ViLT}: Vision-and-language transformer without convolution or region supervision.
\newblock In {\em International Conference on Machine Learning}, pages 5583--5594, 2021.

\bibitem{kwa2025measuring}
Thomas Kwa, Ben West, Joel Becker, Amy Deng, Katharyn Garcia, Max Hasin, Sami Jawhar, Megan Kinniment, Nate Rush, Sydney Von~Arx, et~al.
\newblock Measuring ai ability to complete long tasks.
\newblock {\em arXiv preprint arXiv:2503.14499}, 2025.

\bibitem{roadmap}
Yaguo Lei, Bin Yang, Xinwei Jiang, Feng Jia, Naipeng Li, and Asoke~K. Nandi.
\newblock Applications of machine learning to machine fault diagnosis: A review and roadmap.
\newblock {\em Mechanical Systems and Signal Processing}, 138:106587, 2020.

\bibitem{Lewis2020}
Patrick Lewis, Ethan Perez, Aleksandra Piktus, Fabio Petroni, Vladimir Karpukhin, Naman Goyal, Urvashi Khandelwal, Angela Fan, Vishrav Chaudhary, Francisco Guzm{\'a}n, Tim Rockt{\"a}schel, Marie-Francine Moens, and Veselin Stoyanov.
\newblock Retrieval-augmented generation for knowledge-intensive {NLP} tasks.
\newblock In {\em Advances in Neural Information Processing Systems (NeurIPS)}, pages 9459--9474, 2020.

\bibitem{li2025simpleeffectiverolesgraphs}
Mufei Li, Siqi Miao, and Pan Li.
\newblock Simple is effective: The roles of graphs and large language models in knowledge-graph-based retrieval-augmented generation, 2025.

\bibitem{long2023adaptcontextsretrievalaugmenteddomain}
Quanyu Long, Wenya Wang, and Sinno~Jialin Pan.
\newblock Adapt in contexts: Retrieval-augmented domain adaptation via in-context learning, 2023.

\bibitem{lowenmark2023dataset}
Karl L{\"o}wenmark, Fredrik Sandin, Marcus Liwicki, and Stephan Schnabel.
\newblock Dataset with condition monitoring vibration data annotated with technical language, from paper machine industries in northern sweden.
\newblock {\em Svensk nationell datatj{\"a}nst (SND)}, 2023.

\bibitem{lowenmark2022processing}
Karl L{\"o}wenmark, Cees Taal, Joakim Nivre, Marcus Liwicki, and Fredrik Sandin.
\newblock Processing of condition monitoring annotations with bert and technical language substitution: A case study.
\newblock In {\em 7th European Conference of the Prognostics and Health Management Society 2022 (PHME22), July 6-8 2022, Turin, Italy}, volume~7, pages 306--314. PHM Society, 2022.

\bibitem{lowenmark2023labelling}
Karl L{\"o}wenmark, Cees Taal, Amit Vurgaft, Joakim Nivre, Marcus Liwicki, and Fredrik Sandin.
\newblock Labelling of annotated condition monitoring data through technical language processing.
\newblock In {\em 15th Annual Conference of the Prognostics and Health Management Society, PHM 2023. Salt Lake City, USA. 28 October 2023 through 2 November 2023}, volume~15. Prognostics and Health Management Society, 2023.

\bibitem{lowenmark2021technical}
Karl Löwenmark, Cees Taal, Stephan Schnabel, Marcus Liwicki, and Fredrik Sandin.
\newblock Technical language supervision for intelligent fault diagnosis in process industry.
\newblock {\em International Journal of Prognostics and Health Management}, 13, 2022.

\bibitem{KG_LLM_FD}
Yunfei Ma, Shuai Zheng, Zheng Yang, Hongcheng Pan, and Jun~Hong and.
\newblock A knowledge-graph enhanced large language model-based fault diagnostic reasoning and maintenance decision support pipeline towards industry 5.0.
\newblock {\em International Journal of Production Research}, 0(0):1--22, 2025.

\bibitem{ouyang2022training}
Long Ouyang, Jeff Wu, Xu~Jiang, Diogo Almeida, Carroll Wainwright, et~al.
\newblock Training language models to follow instructions with human feedback.
\newblock In {\em Advances in Neural Information Processing Systems}, volume~35, 2022.

\bibitem{pan2024unifying}
Shirui Pan, Linhao Luo, Yufei Wang, Chen Chen, Jiapu Wang, and Xindong Wu.
\newblock Unifying large language models and knowledge graphs: A roadmap.
\newblock {\em IEEE Transactions on Knowledge and Data Engineering}, 36(7):3580--3599, 2024.

\bibitem{pandya2023automating}
Keivalya Pandya and Mehfuza Holia.
\newblock Automating customer service using langchain: Building custom open-source gpt chatbot for organizations.
\newblock {\em arXiv preprint arXiv:2310.05421}, 2023.

\bibitem{papineni2002bleu}
Kishore Papineni, Salim Roukos, Todd Ward, and Wei-Jing Zhu.
\newblock Bleu: a method for automatic evaluation of machine translation.
\newblock In {\em Proceedings of the 40th Annual Meeting of the Association for Computational Linguistics (ACL)}, pages 311--318, 2002.

\bibitem{peng2024graphretrievalaugmentedgenerationsurvey}
Boci Peng, Yun Zhu, Yongchao Liu, Xiaohe Bo, Haizhou Shi, Chuntao Hong, Yan Zhang, and Siliang Tang.
\newblock Graph retrieval-augmented generation: A survey.
\newblock {\em arXiv preprint arXiv:2408.08921}, 2024.

\bibitem{Radford2021clip}
Alec Radford, Jong~Wook Kim, Chris Hallacy, Aditya Ramesh, Gabriel Goh, Sandhini Agarwal, Girish Sastry, Amanda Askell, Pamela Mishkin, Jack Clark, and Ilya Sutskever.
\newblock Learning transferable visual models from natural language supervision.
\newblock In {\em International Conference on Machine Learning}, pages 8748--8763, 2021.

\bibitem{randall2011rolling}
Robert~B Randall and Jerome Antoni.
\newblock Rolling element bearing diagnostics—a tutorial.
\newblock {\em Mechanical systems and signal processing}, 25(2):485--520, 2011.

\bibitem{randall_bearing_diagnostics}
Robert~B. Randall and Jérôme Antoni.
\newblock Rolling element bearing diagnostics—a tutorial.
\newblock {\em Mechanical Systems and Signal Processing}, 25(2):485--520, 2011.

\bibitem{rao2024ravenmultitaskretrievalaugmented}
Varun~Nagaraj Rao, Siddharth Choudhary, Aditya Deshpande, Ravi~Kumar Satzoda, and Srikar Appalaraju.
\newblock Raven: Multitask retrieval augmented vision-language learning.
\newblock {\em arXiv preprint arXiv:2406.19150}, 2024.

\bibitem{sardana2023beyond}
Nikhil Sardana, Jacob Portes, Sasha Doubov, and Jonathan Frankle.
\newblock Beyond chinchilla-optimal: Accounting for inference in language model scaling laws.
\newblock {\em arXiv preprint arXiv:2401.00448}, 2023.

\bibitem{schwab2017fourth}
Klaus Schwab.
\newblock {\em The fourth industrial revolution}.
\newblock Currency, 2017.

\bibitem{snell2024scalingllmtesttimecompute}
Charlie Snell, Jaehoon Lee, Kelvin Xu, and Aviral Kumar.
\newblock Scaling llm test-time compute optimally can be more effective than scaling model parameters.
\newblock {\em arXiv preprint arXiv:2408.03314}, 2024.

\bibitem{tang_DL_CM}
Shengnan Tang, Shouqi Yuan, and Yong Zhu.
\newblock Deep learning-based intelligent fault diagnosis methods toward rotating machinery.
\newblock {\em IEEE Access}, 8:9335--9346, 2020.

\bibitem{thus2024exploring}
Dominik Th{\"u}s, Sarah Malone, and Roland Br{\"u}nken.
\newblock Exploring generative ai in higher education: a rag system to enhance student engagement with scientific literature.
\newblock {\em Frontiers in Psychology}, 15:1474892, 2024.

\bibitem{tiboni2022review}
Monica Tiboni, Carlo Remino, Roberto Bussola, and Cinzia Amici.
\newblock A review on vibration-based condition monitoring of rotating machinery.
\newblock {\em Applied Sciences}, 12(3):972, 2022.

\bibitem{vaswani2017attention}
Ashish Vaswani, Noam Shazeer, Niki Parmar, Jakob Uszkoreit, Llion Jones, Aidan~N Gomez, Łukasz Kaiser, and Illia Polosukhin.
\newblock Attention is all you need.
\newblock {\em Advances in neural information processing systems}, 30, 2017.

\bibitem{wei2022chain}
Jason Wei, Xuezhi Wang, Dale Schuurmans, Maarten Bosma, Ed~Chi, Quoc Le, and Denny Zhou.
\newblock Chain of thought prompting elicits reasoning in large language models.
\newblock {\em Advances in Neural Information Processing Systems}, 35:24824--24837, 2022.

\bibitem{wu2024inference}
Yangzhen Wu, Zhiqing Sun, Shanda Li, Sean Welleck, and Yiming Yang.
\newblock Inference scaling laws: An empirical analysis of compute-optimal inference for problem-solving with language models.
\newblock {\em arXiv preprint arXiv:2408.00724}, 2024.

\bibitem{yan2017industrial}
Jihong Yan, Yue Meng, Lei Lu, and Lin Li.
\newblock Industrial big data in an industry 4.0 environment: Challenges, schemes, and applications for predictive maintenance.
\newblock {\em IEEE access}, 5:23484--23491, 2017.

\bibitem{yao2023react}
Shunyu Yao, Jeffrey Yu, Juncheng Shi, Michele Luo, Karthik Narasimhan, and Christopher Rè.
\newblock React: Synergizing reasoning and acting in language models.
\newblock {\em arXiv preprint arXiv:2210.03629}, 2023.

\bibitem{yu2025visragvisionbasedretrievalaugmentedgeneration}
Shi Yu, Chaoyue Tang, Bokai Xu, Junbo Cui, Junhao Ran, Yukun Yan, Zhenghao Liu, Shuo Wang, Xu~Han, Zhiyuan Liu, et~al.
\newblock Visrag: Vision-based retrieval-augmented generation on multi-modality documents.
\newblock {\em arXiv preprint arXiv:2410.10594}, 2024.

\bibitem{yuan2024ragdrivergeneralisabledrivingexplanations}
Jianhao Yuan, Shuyang Sun, Daniel Omeiza, Bo~Zhao, Paul Newman, Lars Kunze, and Matthew Gadd.
\newblock Rag-driver: Generalisable driving explanations with retrieval-augmented in-context learning in multi-modal large language model.
\newblock {\em arXiv preprint arXiv:2402.10828}, 2024.

\bibitem{zhang2025llm}
Qi~Zhang, Chao Xu, Jie Li, Yicheng Sun, Jinsong Bao, and Dan Zhang.
\newblock Llm-tsfd: An industrial time series human-in-the-loop fault diagnosis method based on a large language model.
\newblock {\em Expert Systems with Applications}, 264:125861, 2025.

\bibitem{zhang2024visionsearchassistantempower}
Zhixin Zhang, Yiyuan Zhang, Xiaohan Ding, and Xiangyu Yue.
\newblock Vision search assistant: Empower vision-language models as multimodal search engines.
\newblock {\em arXiv preprint arXiv:2410.21220}, 2024.

\bibitem{zhao2023retrievingmultimodalinformationaugmented}
Ruochen Zhao, Hailin Chen, Weishi Wang, Fangkai Jiao, Xuan~Long Do, Chengwei Qin, Bosheng Ding, Xiaobao Guo, Minzhi Li, Xingxuan Li, et~al.
\newblock Retrieving multimodal information for augmented generation: A survey.
\newblock {\em arXiv preprint arXiv:2303.10868}, 2023.

\bibitem{zhao2024unifashionunifiedvisionlanguagemodel}
Xiangyu Zhao, Yuehan Zhang, Wenlong Zhang, and Xiao-Ming Wu.
\newblock Unifashion: A unified vision-language model for multimodal fashion retrieval and generation.
\newblock {\em arXiv preprint arXiv:2408.11305}, 2024.

\end{thebibliography}

\section{Appendix: Agent Prompts}
\label{appendix:agent_prompts}
\subsubsection{Main Agent Prompt}
The main agent prompt is rather short, as most relevant information can be stored in the tools, such as how to make hierarchy MindRAG calls, or how to call the other agents.
This agent does not require complex analysis itself, and acts as a conversational agent layer between the complex methods and the user.

\begin{lstlisting}
    """You are a condition monitoring assistant.
Given the user's query you must decide what to do with it based on your list of tools and internal knowledge.

If you see that a tool has been used (in the scratchpad) with a particular query, do NOT use that same tool with the same query again.

You should aim to collect information from a diverse range of sources before providing the answer to the user.
Once you have collected plenty of information to answer the user's question (stored in the scratchpad) use the final_answer tool.

You have access to a condition monitoring dataset in Swedish with data from 2020 and 2021. Any queries to this dataset should be done in Swedish, so that e.g. "sensor" becomes "givare",
"roller" becomes "vals", etc. If you are unsure regarding a translation, ask the user to clarify or use the query_hierarchy tool and read all relevant data to form your opinion.

If you need to do any type of complex CM analysis, such as retrieving and analysing chunks, pass the question along with relevant information to the CM_agent node that you are connected to.
You can also call the CM_scheduler agent to maintain a list of detected faults with TODOs, such as when to follow up, or what conditions warrant further analysis.
"""
\end{lstlisting}

\subsubsection{CM Agent Prompt}
This agent requires extensive prompting on how to perform CM fault diagnosis analysis.
This is best stored in the main prompt, rather than in specific tools, and the quality of this prompt likely determines the capacity of the agent to properly analyse hierarchy data and the nuance of fault descriptions with regard to timedelta and component data.

\begin{lstlisting}
    
"""You are a condition monitoring decision making assistant.
Your task is to make machine component fault assessments about assets based on a series of recording chunks that have been previously analysed by a human expert in the form of an annotation/note.
You will receive data in time steps surrounding the note date, and your goal is to as quickly as possible identify relevant information such as faults based on your input, and support the human by providing your analysis in the form
of a fault prediction alarm. The sooner you can make an alarm prediction, the easier it will be for the human to schedule effective and resource efficient maintenance. Waiting too long for certain faults such as 
"glapp" or "hylsslapp" can even result in critical breakdowns and serious damages.
However, making a wrong prediction because you didn't gather enough data will waste a lot of time and make the human lose trust in your capabilities.
If you do not make a prediction once all data has been supplied, and a real fault was present, that is treated as a failed prediction. However, if the real annotation describes no fault, waiting was the
correct call. You might also be analysing data long before a note, or after a note featuring replacement information (healthy data), in which case your prediction should be that no faults are present.
However, notifying the user of healthy data, even if correct, would waste their time. Even worse is if you notify the user of a fault when no fault is present (false positive), in which case you forced the 
user to do a follow up analysis in vain. Therefore make a prediction once you are fairly certain that a fault is present, but if you detect a fault, don't delay too long as the fault might develop.
Thus, you must prioritise swift analysis, but not at the cost of accuracy or false positives. Data that indicates minor faults such as cable or sensor ("givare" in Swedish) faults, or low levels of bearing faults, can be
analysed for longer, while critical breakdowns ("haveri") are more urgent. Use the contents of the data to make sound judgement. If you are prompted to make a prediction, but you have not detected sufficient
fault levels to warrant quitting voluntarily, please make clear the state of your current analysis, that you believe that certain faults could be present, but that you do not think they warrant human intervention yet.
Also describe whether you would've generated an alarm at this point or not, being mindful of the risk associated with false alarms and the risk of neglecting real fault developments.

All data is in Swedish.

Your input will be the asset path (describes where in the machine the component is), point name (describes the name of the sensor attached to the machine component. One machine (asset) can have multiple points),
sensor type (describes whether the sensor stores raw frequency data ('rms') or envelope filtered data (''Peak' or 'PtP')), and most importantly recording chunks to analyse and similar chunks from the dataset 
that were retrieved with MindRAG chunk similarity from the condition monitoring vector store.
These MindRAG chunks all have associated annotations describing properties such as faults, and you can use these annotations from similar recordings to infer information about the asset where 
the input chunks are from. The MindRAG chunks also feature information about which asset and point they're from, and the time delta between the chunk and its associated annotation.
Input will be provided in slices, with one slice featuring one input chunk with associated MindRAG chunks per point connected to the same asset. Thus, an component with three points will feature three
input chunks with five MindRAG chunks each per slice.
Slices will be provided one at a time per your request, until you're ready to make a assessment. You should always read at least ten slices before proceeding to the assessment stage.

The input chunks and MindRAG chunks both consist of chunk_ids, unique strings identifying one recording;
chunk_tds, which are time deltas describing the number of days to the annotation date (negative is before, positive is after);
asset(machine) paths, point(sensor on component) names, and annotation timedelta and content, which all describe properties of the sensor where a recording chunks comes from.
The input chunks all come from the same asset, point, and annotation, while the MindRAG chunks can come from anywhere in the machine.
Compare information on where in the machinery the recordings are from as one part of your analysis.

Use the associated annotations and the time_deltas to make an assessment of what's happening with the machine. If, for instance, a component initially is associated with bearing faults (
BPFO, BPFI, "lager"-associated words), or notes describing a maintenance action ("bytt", "utbytt", etc.), but then switches to notes describing a cable ("kabel" or sensor ("givare" in Swedish)
fault at the very end, it's likely that no serious faults were present early, but a malfunctioning sensor or cable closer to the note date motivated the annotation.

The associated MindRAG chunk notes are based on recording similarities. Since each annotation has a plethora of associated recordings, not all recordings will reflect the content of the annotation.
Many faults do not appear consistently in the recordings; for example, a cable or sensor might malfunction occasionally, while being healthy or even having another, slowly developing fault such as BPFO,
in between the signal malfunctions. Therefore, a recording associated with a cable or sensor annotation can both be due to clear fault indicators, or due to similarities in the latent noise space between
two fault-free signals, etc.

Furthermore, data is gathered from both pre and post annotation date. Therefore, maintenance action annotations can be associated both with fault features if sampled prior to action (negative timedelta),
and fault free measurements when sampled post replacement. A cable or sensor replacement can be followed by low level bearing fault indications as well, due to these faults developing slowly and often 
being noticed first after reaching certain levels of development. Likewise, data gathered from many days or weeks prior to the annotation date can be before the fault started to develop, which should also
be taken into account. Be mindful of the time deltas of the associated MindRAG chunks; if, for instance, many chunks describe a maintenance action e.g. "... bytt" and the time delta is positive, i.e. after
annotation date, the MindRAG recording chunks are likely relatively free from fault features. Input chunks that are after the annotation (positive timedelta) and associated with such MindRAG chunks are thus
likely free from faults, and likely associated with an annotation describing a replacement, since they're after the replacement.

Likewise, MindRAG chunks associated with a maintenance action but with a negative timedelta, i.e. before the annotation, can contain fault features of the fault that warranted the replacement, e.g.
cable or sensor fault features for a cable or sensor replaced comment, and bearing fault features if a bearing was replaced, etc. Chunks from long before a fault annotation, e.g. 50 days (-50 time delta)
before a "BPFO laaga nivaaer", can also be mostly free from fault features, and contain at most low levels of whatever signal features motivated the annotation.

It is your role to consider both the contents of the annotations and the time deltas of the recordings to make a sound assessment of what has happened that motivated the annotation. Base your initial 
assessment on the content of the MindRAG chunks, and the time deltas to interpret the MindRAG chunk annotations. Each true note that you are trying to predict will feature only one type of fault, so if you 
are unsure about which prediction to make, e.g. due to varying MindRAG annotation information, or the MindRAG annotations have timedeltas that indicate that no fault is currently present, ask the user for more slices.
Furthermore, cable and sensor faults are less harmful for operations, so if you suspect cable or sensor faults, wait until at least ten slices confirm your suspicion (still not proceeding until a total of 
twenty slices have been provided by the user).

An important fact about slices is that not every point in the slice will necessarily predict a true fault. Some points are different measurement types from the same sensor, while other points are different sensors.
If multiple points with different names have the same point type, e.g. of six points, two are "PtP" while one is named DS-something and the other FS-something, then these are likely different sensors. If
for instance three points are present, each with a different measurement type, this is likely the same sensor with different filters. A common abbreviation for envelope filters is "AE", "VE", or "HE", where
"AE" stands for axial envelope, "VE" for vertical envelope, and "HE" for horizontal envelope. Likewise, raw frequency data often comes from points named "AV", "VV", or "HV", where "AE" stands for axial velocity, 
"VE" for vertical velocity, and "HE" for horizontal velocity. The "A", "V", and "H", all describe where (in which direction) on the component the sensor is mounted, which will affect how the signals look. In general, if a component
has more than one mounting direction, it's likely that each direction is one sensor, so the component has multiple sensors. Sensors can also be mounted on both the "free side" (FS) and "drive side" (DS),
which is another way to identify whether a component has one or multiple sensors. FS and DS measurements measure the same feature from different ends, so it's likely that they will correlate to some extent
in detection component faults such as bearing faults. A sensor malfunction should happen only at one of the two sensors however, which can be a way to infer what's happening based on point properties.
Consequently, it is possible that the fault you are tasked to predict is present only in one point, or in one sensor out of multiple sensors. When making the fault assessment, describe what you think is
happening at each point, and include which point you think is the one that warranted the annotation.

A good way to detect if a fault is occurring is if the associated MindRAG chunks switch from one type to another. If, after twenty slices, one point has changed associated MindRAG annotations, 
while the others are static, it is likely that this point warranted the real annotation. However, if, for instance, most points indicate a type of bearing fault, but one point or set of points changes 
to cable or sensor faults, please also mention that there might be an underlying bearing fault besides the appearance of a cable or sensor fault. If a point instead goes from cable or sensor faults to 
bearing faults, it is possible that the cable or sensor was fixed, and that the measurements now indicate low levels of another fault that was previously missed due to the faulty signals.

When you provide your final analysis, you should both mention your assessments of each point and a prediction of the contents of the real note. If one point indicates BPFO and the other a cable fault,
describe both, but use your knowledge and the input data to decide whether the real note described a cable or BPFO fault, and clarify your prediction in your final analysis.

Once you are ready to make an assessment, exit the graph with the "reply_with_prediction" tool and mention all your takeaways in your prediction. If raw signals are available, the user might supply
these for follow up analysis. If this is the case, use the "check_knowledge_base" tool to learn how to analyse raw signals, and reply with a new prediction with "reply_with_prediction" where you explain
your analysis before analysing the signals, after analysing the signals, and how the signals updated your analysis.
"""
\end{lstlisting}

\subsubsection{Evaluation Agent Prompt}
The evaluation agent's prompt determines how the performance of the CM agent is analysed, and is useful for research rather than runtime implementations of a MindRAG agent.
However, the evaluation could be used for an automated RLHF system, which would warrant further research into this agent.

\begin{lstlisting}
    """
You are a condition monitoring expert tasked with evaluating the output of another condition monitoring expert LLM based on the LLM fault diagnosis prediction and the true analysis done by a human,
as defined by the contents of an annotation.
You will receive the output of the other LLM, and the real annotation, and your task is to analyse these and write your analysis to file using the "write_evaluation" tool.
Your analysis should feature the following:
'fault_diagnosis_prediction', which is the prediction generated by the other LLM.
'note', which is the true note supplied by the human user.
'evaluation', your evaluation of whether the prediction and the note align. The evaluation should be written as "korrekt", "delvis korrekt", or "felaktig".
A prediction is correct when the contents of the main prediction mostly align with the contents of the note. For example, if the prediction states that a bearing fault of type BPFI is present, but the true
note states BPFO, this can be treated as correct as these faults are very similar in the dataset, and vice versa. Cable and sensor faults are virtually indistinguishable and a cable fault prediction is 
correct if the note states sensor ('givare' in Swedish) fault, and vice versa. The prediction is also correct if the note features a maintenance action, e.g. that a bearing has been replaced, and the 
prediction meantions a bearing fault that warrants a replacement, or vice versa. The same goes for any type of maintenance action and fault, e.g. cable or sensor faults and cable or sensor replacements.
A prediction is partially correct when a point prediction is correct but not prioritised, e.g. if one point indicated a bearing fault and another point indicates a sensor fault, and the prediction
highlights the bearing fault while the true note was the sensor fault. It is also partially correct if it identifies a relevant fault but fails to describe all main contents of the note, e.g. detecting a
BPFO but without any mentiosn of scheduling a maintenance action or a maintenance action taking place.
A prediction is incorrect when no point predictions corresponded to the note contents.

'reflection', which elaborates on the differences between the prediction and the note, and speculates on reasons why the analysis was right or wrong.
'point_name', the list of point names of the analysed input chunks supplied by the user. These should not be the MindRAG chunks, but the input point names.
'path_name', the asset machine path of the analysed input chunks supplied by the user.
'score', a more complex evaluation metric that can be either true positive, false positive, true negative, or false negative.
The score a true positive when the prediction is correct or partially correct and the true note describes a fault of maintenance action. It is a false positive if the agent predicts the wrong fault. It is a false negative
if the true note describes a fault but the agent predicted no fault, waited until the annotation date before making a prediction. It is a true negative if the agent correctly identifies that no faults 
are present, or if the agent waited until the annotation date before making a prediction when the note describes no faults.
Furthermore, if the recordings are after a "replaced" ("bytt") comment (i.e. positive time delta), and the prediction system predicts healthy or doesn't make a prediction,
it is a true negative as we can assume that no faults were present after replacement. If the system makes a fault prediction as primary prediction in this case, it is thus a false positive.
'note class', an estimation (by you) of what type of fault or action the note pertained to.
'pred class', an estimation (by you) of what type of fault or action the prediction pertained to.
"""
\end{lstlisting}

\section{Appendix: Knowledge Retrieval Examples}
\label{appendix:knowledge_retrieval}
\subsection{Graph Text Analysis}
\subsubsection{Query}
Please check if there is any need for lubrication in the holländeri

\subsubsection{Answer}
Thought: To check if there is a need for lubrication in the "holländeri", I would need to check the annotations for any notes that mention lubrication requirements. First, I need to confirm that "holländeri" is a valid part of the machine hierarchy.

Action: Check Hierarchy
Action Input: ['holländeri']
Observation: True
Thought:Since "holländeri" is a valid part of the machine hierarchy, I can proceed to query the hierarchy for notes related to lubrication in this area.

Action: Query Hierarchy
Action Input: [[['Path', ['holländeri']], [], ['noteComment', ['lubrication', 'smörjning']]], [[], [], ['noteComment', 'noteDateUTC']]]
Observation: []
Thought:The query returned no results, which means there are no annotations mentioning lubrication requirements for the "holländeri". This could mean that there is no immediate need for lubrication, or that it has not been annotated in the notes. However, to provide a more comprehensive answer, I should check for any recent maintenance actions or faults that could imply a need for lubrication.

\textit{Comment: No annotations either matched the filter nor were close enough in the RAG space. Therefore the agent proceeds with a general note search, which is applicable since the holländeri group is unlikely to contain too much data for an LLM to directly analyse.}

Action: Query Hierarchy
Action Input: [[['Path', ['holländeri']], [], []], [[], [], ['noteComment', 'noteDateUTC']]]

Observation: The agent observes all notes in the "holländeri"
Thought:The annotations for the "holländeri" include various maintenance actions and fault detections, such as bearing replacements and increased levels in motors. There is a note mentioning lubrication directly ("500 gram eftersmörjt."), which translates to "500 grams lubricated afterwards." This note is dated 2020-12-29, which is quite some time before the current date of 2021-08-10. Given that there are no recent notes about lubrication and considering the time elapsed since the last lubrication note, it might be reasonable to suggest that a lubrication check could be due. However, without specific lubrication intervals or a clear indication of current lubrication status, I cannot definitively calculate the need for lubrication.

Final Answer: Based on the available annotations, there is no recent note explicitly stating a need for lubrication in the "holländeri". However, since the last lubrication note dates back to 2020-12-29, it may be prudent to check the lubrication status of the "holländeri" to ensure proper maintenance. It is recommended to consult the maintenance schedule or perform a physical inspection to accurately assess the need for lubrication.

\subsection{Historic Insight and Analysis}
\subsubsection{Query}
Hello! Which assets from drying cylinder group 2 have unresolved bearing fault annotations? i.e. bearing faults that
have been identified but do not have a work order afterwards?

\subsubsection{Answer}

Thought: To answer this question, I need to query the hierarchy to find assets from drying cylinder group 2 with bearing fault annotations and then check if there are any follow-up work orders. I will start by querying the hierarchy for assets under drying cylinder group 2 with notes containing bearing fault keywords like "BPFO", "BPFI", or "lager" and then analyze the annotations to see if there are any unresolved faults.

Action: Query Hierarchy
Action Input: [[['Path', ['torkgrupp 2']], [], ['noteComment', ['BPFO', 'BPFI', 'lager']]], [['ID', 'Name'], [], ['noteComment','noteDateUTC','idNode']]]

Observation: The agent observes a large amount of annotations, points, and assets.

Thought:The observation contains a lot of data with various note comments and dates. I need to analyze the note comments to identify bearing faults and then check if there is a follow-up work order (AO) after the fault detection. I will look for keywords indicating a bearing fault (like "BPFO", "BPFI", "lagerskada") and then check for any mention of a work order or part replacement (like "AO", "utbytt") that would indicate the fault has been resolved. If there is a bearing fault mentioned without a subsequent work order, that asset has an unresolved bearing fault.

Action: Analyze language datas
Action Input: The note comments and dates from the observation to identify unresolved bearing faults.
Observation: The note comments and dates from the observation to identify unresolved bearing faults.
Thought:After analyzing the note comments and dates, I can identify which assets have unresolved bearing faults by looking for notes that mention bearing faults without a subsequent work order or part replacement. I will look for the latest note mentioning a bearing fault and then check if there is a later note that mentions a work order (AO) or part replacement (like "utbytt"). If there is no such note, the fault is unresolved.

Final Answer: Based on the analysis of the note comments and dates, the following assets from drying cylinder group 2 have unresolved bearing fault annotations:

1. Asset ID 5808, Name '09626,30 2:a Tork Drift' - The latest note on 2020-03-16 mentions a bearing fault ("Lagerskada växel in. Håll koll."), and there is no subsequent work order or part replacement note.
2. Asset ID 1554, Name '09626,219 FÖ 210 FS' - The latest note on 2020-03-25 mentions a bearing fault ("BPFO i env. Låga värden, avvakta."), and there is no subsequent work order or part replacement note.
3. Asset ID 1625, Name '09626,229 FU 205 FS' - The note on 2018-08-27 mentions a bearing fault ("BPFO i Env3 avvakta mm/s"), and there is no subsequent work order or part replacement note.
4. Asset ID 1628, Name '09626,232 FU 208 FS' - The note on 2018-09-17 mentions a bearing fault ("BPFO Låga nivåer i Env"), and there is no subsequent work order or part replacement note.
5. Asset ID 1630, Name '09626,234 FU 210 FS' - The note on 2020-12-14 mentions a bearing fault ("BPFO låga nivåer och ej stigande trend så det är nog ganska lungt. Håll koll."), and there is no subsequent work order or part replacement note.
6. Asset ID 1685, Name '09626,236 FU 212 FS' - The latest note on 2021-04-09 mentions a bearing fault ("BPFO syns men väldigt låga nivåer, håll koll."), and there is no subsequent work order or part replacement note.

These assets have notes indicating bearing faults without any clear follow-up work orders, suggesting that the faults may still be unresolved.

\subsubsection{Query 2}
Please compute the changes in spectrum average over the last three days in these assets, and return the three biggest changes.

\subsubsection{Answer}
Thought: To compute the changes in spectrum average over the last three days for the specified assets, I need to use the "Load Data and Compute Changes" tool. I will need to provide the asset IDs, the time period for the analysis, and specify the spectrum method as 'np.average'. Since today is 2021-08-10, the last three days would be from 2021-08-07 to 2021-08-10.

Action: Load Data and Compute Changes

Action Input: [[['ID', [5808, 1554, 1625, 1628, 1630, 1685]], [], []], [['2021-08-07T00:00:00', '2021-08-10T00:00:00'], False, 3], [[], ['np.average'], [], []]]

Observation: \textit{The agent observes retrieved data.}

Thought:I now have the data for the three assets with the biggest changes in spectrum average over the last three days. The assets are 'FU 210 FS VV', 'FU 208 FS VV', and 'STATIV IN AV', corresponding to asset names '09626,234 FU 210 FS', '09626,232 FU 208 FS', and '09626,30 2:a Tork Drift'.

Final Answer: The three assets from drying cylinder group 2 with the biggest changes in spectrum average over the last three days are:

1. '09626,234 FU 210 FS' (Asset ID 1630)
2. '09626,232 FU 208 FS' (Asset ID 1628)
3. '09626,30 2:a Tork Drift' (Asset ID 5808)

These assets showed the highest changes in the spectrum average, which could indicate a progression in the bearing faults that were previously identified.

\subsection{Updating and Using Knowledge Bases}
\subsubsection{Query 1}
Do you know how to calculate if there are sensor or cable faults in the condition monitoring data?

\subsubsection{Answer 1}
Thought: The user is asking for a method to calculate sensor or cable faults in the condition monitoring data. I need to check my knowledge base to see if I have instructions on how to perform this analysis.

Action: Check Knowledge Base
Action Input: How do I calculate sensor or cable faults in condition monitoring data?
Observation: ['Sensor faults can be analysed by computing the change over the trend levels using np.max']
Thought: The knowledge base has provided a method for analyzing sensor faults by computing the change over the trend levels using the numpy method np.max. However, it does not mention anything about cable faults specifically. I should clarify with the user if they want to proceed with this method for sensor faults and ask if they have any specific instructions for cable faults.

Final Answer: The knowledge base indicates that sensor faults can be analyzed by computing the change over the trend levels using the numpy method np.max. However, it does not provide specific instructions for analyzing cable faults. Would you like to proceed with this method for sensor faults, and do you have any specific instructions for analyzing cable faults?

\subsubsection{Query 2}
Please also add to your knowledge base that the variance of the bias can be indicative of these faults, then compute if there are any present at vira topp.

\subsubsection{Answer 2}

Thought: The user has provided new information regarding the analysis of sensor and cable faults, which is that the variance of the bias can be indicative of these faults. I need to write this information to the knowledge base before proceeding with the analysis at "vira topp". I also need to check if "vira topp" is a valid part of the hierarchy.

Action: Write to Knowledge Base
Action Input: Add to the knowledge base that the variance of the bias (using numpy method np.var) can be indicative of sensor and cable faults in condition monitoring data.
Observation: None
Thought:I have updated the knowledge base with the new information regarding sensor and cable faults. Now I need to verify if "vira topp" is a valid part of the hierarchy before proceeding with the analysis.

Action: Check Hierarchy
Action Input: ['vira topp']

\textbf{Observation:} (all assets in the graph)

Thought:The hierarchy check indicates that "vira topp" is indeed a valid part of the hierarchy, but it should be noted as "vira topp " with a space at the end to match the hierarchy naming convention. Now that I have confirmed "vira topp" is a valid part of the hierarchy and updated the knowledge base with the new information regarding sensor and cable faults, I can proceed to query the hierarchy for notes related to sensor and cable faults at "vira topp" and then compute changes based on the variance of the bias as indicated by the user.

Action: Query Hierarchy

Action Input: [[['Path', ['vira topp ']], [], ['noteComment', ['kabelfel', 'givarfel', 'kabeln utbytt', 'givaren utbytt']]], [[], [], ['idNode', 'noteComment', 'noteDateUTC']]]

Observation: \textit{Retrieved data}
Thought:The query returned several notes related to "vira topp". Some notes mention cable replacement ("Kabeln utbytt.") and sensor issues ("Givaren skall bytas nästa stopp.", "kabelfel,"). There is also a note about a bearing failure, but it is not directly related to sensor or cable faults. The dates of the notes range from 2018 to 2021, with the most recent note being on the same date as today, 2021-08-10. Now, I need to load the data and compute changes for the asset IDs returned, focusing on the variance of the bias to identify sensor and cable faults as per the user's instructions.

Action: Load Data and Compute Changes

Action Input: [[['ID', [6651]], [], []], [['2021-08-10T00:00:00', '2021-08-10T23:59:59'], [0, 800], False], [[], [], [], ['np.var']]]

Observation: \textit{Retrieved data}

Thought: The data analysis has returned information for asset ID 6651, which corresponds to "64440 21:A Virareturvals DS". The notes associated with this asset include various comments about maintenance actions and observations, such as a bearing failure, cable replacement, and plans to replace a sensor. The variance of the bias was computed for the specified date range, which is the same day as today, and within the specified machine shaft speed range (0 to 800 RPM). The results include notes that mention cable and sensor issues, which align with the user's request to identify sensor and cable faults. The notes indicate that there have been recent actions and plans regarding these faults, which suggests that the system has identified and is addressing the issues.

Final Answer: The condition monitoring data for "vira topp" has been analyzed, and the variance of the bias was computed for asset ID 6651. The analysis identified notes that mention sensor and cable faults, and there have been recent maintenance actions and plans to address these issues. The system appears to be actively managing the faults as per the notes on the specified asset.

\subsection{Computing Feature Changes}
\subsubsection{User Query}
Hello! What are the changes in the max values for sensor faults in June and August 2021 in TG 1? Please describe the characteristics of their associated assets and points

\subsubsection{Action 1}
Thought: To answer this question, I need to identify sensor faults in TG 1 (torkgrupp 1) and then load the data and compute the changes in max values for the specified time period. I will need to query the hierarchy to find the relevant assets and points associated with sensor faults, and then use the Load Data and Compute Changes tool to calculate the changes in max values. First, I should check if "TG 1" is a valid part of the hierarchy.

\textbf{Action:} Check Hierarchy

\textbf{Action Input:} ['TG 1']

\textbf{Observation:} (All assets in the machine)

\textit{Comment: The Check Hierarchy tool is coded to return True if the action input is a valid hierarchy path, and to return the hierarchy tree for analysis otherwise. The user tricked the agent since there is no "TG 1", as it is split into 1A and 1B. The tool thus return the hierarchy (the agent is aware of this through the tool description)}

\subsubsection{Action 2}
\textbf{Thought:} The correct hierarchy path for TG 1 is "torkgrupp 1a" and "torkgrupp 1b". I will need to query the hierarchy for sensor faults in these areas and then load the data and compute the changes in max values for the specified time period of June and August 2021.

\textbf{Action:} Query Hierarchy

\textbf{Action Input:} [[['Path', ['torkgrupp 1a', 'torkgrupp 1b']], [], ['noteComment', ['givarfel', 'givaren utbytt']]], [['ID', 'Name', 'Path'], ['ID', 'Name', 'DetectionName'], ['noteComment', 'noteDateUTC']]]

\textbf{Observation:} (Data corresponding to the query)

\textit{Comment: The agent does a vector store search filtering for the correct assets, using all points, with the note space search centred around sensor faults where retrieval is done based on keyword + embedding similarity. The agent requests the asset ID, name and path; point ID, name, and sensor type (DetectionName); and the content and date of the annotation.}

\subsubsection{Action 3}
\textbf{Thought:} The hierarchy query returned several assets and points with notes indicating sensor faults and replacements. However, the notes also include other faults and actions, such as bearing replacements and cable changes. To answer the user's question, I need to focus on sensor faults and their max value changes between June and August 2021. I will filter out the notes that are not related to sensor faults and then proceed to load the data and compute the changes for the remaining relevant points.

\textbf{Action:} Analyze language datas

\textbf{Action Input:} ['Givaren utbytt.', 'AO 599547 skrivet på kabelbyte.', 'Givare utbytt.', 'Givaren skall bytas.', 'Givaren utbytt.', 'Givaren skall bytas nästa stopp.', 'Givare trasig ?? /HH', 'Givaren utbytt.', 'Givaren skall bytas nästa stopp.', 'Lager utbytt.', 'AO 570551 skrivet på lagerbyte.', 'BPFO i ENV med sidband på Hållare', 'Bytt givare', 'AO 606799 skrivet på kabelbyte.', 'Givaren utbytt.', 'Givaren skall bytas nästa stopp.', 'Givaren utbytt.', 'Givaren skall bytas nästa stopp.']

\textbf{Observation:} (None -- no tools was used)

\textit{Comment: The agent analyses the retrieved data to verify whether the retrieval was fully relevant}

\subsubsection{Action 4}
\textbf{Thought:} The language analysis indicates that several notes are related to sensor replacements ("Givaren utbytt", "Givare utbytt", "Bytt givare") and intentions to replace sensors ("Givaren skall bytas", "Givaren skall bytas nästa stopp"). There are also notes related to cable changes ("AO 599547 skrivet på kabelbyte", "AO 606799 skrivet på kabelbyte") and bearing replacements ("Lager utbytt", "AO 570551 skrivet på lagerbyte"), which are not relevant to the user's question about sensor faults. I will exclude these unrelated notes and focus on the sensor replacement notes for the assets and points identified in the hierarchy query. Now, I need to load the data and compute the changes in max values for the sensor faults between June and August 2021 for the relevant points.

\textbf{Action: } Load Data and Compute Changes

\textbf{Action Input:} [[['ID', [1910, 1927, 1948, 5754]], ['ID', [1911, 1912, 1913, 1938, 1939, 1940, 1957, 1958, 1959, 5768, 5769, 5770]], []], [['2021-06-01T00:00:00', '2021-08-31T23:59:59'], [0, 800], False], [['np.max'], [], [], []]]

\textbf{Observation:} (Data corresponding to the tool call.)

\textit{Comment: The agent makes a correct tool call for data from the associated notes with the instruction to compute differences over time between the specified dates from the original query using np.max as the applied function. The other function boxes are left empty as they were not specified. The agent filters for speed between o and 800 due to tool instructions, though this could also be left blank.}

\subsubsection{Action 5}
\textbf{Thought:} The Load Data and Compute Changes tool has returned the max value changes for the specified points and assets associated with sensor faults in TG 1 for June and August 2021. The points with the highest changes in max values are 'FÖ A5 FS VV', 'FU B1 FS VE3', and 'FÖ B5 FS VV'. These points are associated with the assets '09625,43 FÖ A5 Rikt FS', '09625,72 FU B1 FS', and '09625,65 FÖ B5 FS', respectively. The notes associated with these points confirm sensor replacements and plans for sensor replacements, which align with the user's request for sensor faults. There are no notes indicating a work order after the detection of the sensor fault within the specified time frame, which means the data analysis is consistent with the user's query. Now I can provide the final answer to the user.

\textbf{Final Answer:} The changes in the max values for sensor faults in June and August 2021 in TG 1 are as follows:

1. Point 'FÖ A5 FS VV' associated with asset '09625,43 FÖ A5 Rikt FS' showed a change in max values. The sensor was replaced as indicated by the note on 2021-04-15.

2. Point 'FU B1 FS VE3' associated with asset '09625,72 FU B1 FS' also showed a change in max values. The sensor was replaced, and there was a note about a possibly faulty sensor on 2021-08-19.

3. Point 'FÖ B5 FS VV' associated with asset '09625,65 FÖ B5 FS' showed a change in max values. The sensor was replaced as indicated by the note on 2021-08-10.

These points and assets are part of TG 1 and are associated with sensor faults and replacements within the specified time period. No work orders were noted after the detection of the sensor faults.

\textit{Comment: The agent returns the three highest values as this is the default setting. For a better user experience the agent could have asked the user to specify the number of max values, instead of relying on the default value from the tool description. The analysis was correct, and the agent uses graph knowledge to structure the output and provide relevant metadata such as closest relevant note.}

\section{Appendix: MIRAGe Runs}
\label{appendix:MIRAGE_runs}

\subsection{Evaluation With Data Before Annotations}

\subsubsection{Prediction 1}
\begin{lstlisting}[language={[LaTeX]TeX}]
Prediction: 
Efter att ha analyserat de tillgängliga skivorna har jag observerat följande trender:

1. **[Point]**: MindRAG-chunkarna för denna punkt är konsekvent associerade med anteckningar om utbyte av glättvalsar. Detta tyder på att inspelningarna sannolikt är efter underhåll och kanske inte indikerar nuvarande fel.

2. **[Point]**: MindRAG-chunkarna för denna punkt är associerade med anteckningar om att BPFO (Ball Pass Frequency Outer) börjar dyka upp. Detta kan indikera tidiga tecken på lagerfel.

3. **[Point]**: MindRAG-chunkarna för denna punkt är kopplade till anteckningar om sensorbyte, vilket indikerar potentiella sensorproblem. Detta kan innebära att sensorn fungerar felaktigt eller ger opålitliga data.

Givet den blandade naturen av data, med potentiella tidiga lagerfelindikationer och sensorproblem, är min förutsägelse att det kan finnas ett underliggande lagerfel som börjar utvecklas, men att det också finns sensorproblem som kan ha påverkat mätningarna. Det är troligt att anteckningen handlar om sensorproblem, men det kan också finnas en underliggande lagerfråga som bör övervakas.

Annotation: 
Höjd nivå på [Point]. Tror det är något märfel (givarproblem) för det blir världens skidbacke vid "live" mätning.

Evaluations: 
korrekt

Reflections: 
Prediktionen identifierade korrekt att det fanns sensorproblem, vilket stämmer överens med anteckningen som nämner ett givarproblem. Prediktionen nämnde också ett potentiellt lagerfel, vilket inte nämndes i anteckningen, men huvudfokus på sensorproblem gör att prediktionen kan anses korrekt. Det är möjligt att prediktionen tog hänsyn till fler datapunkter än vad som var relevant för den specifika anteckningen.

Time delta: 
8

\end{lstlisting}

\subsubsection{Prediction 2}
\begin{lstlisting}[language={[LaTeX]TeX}]
Prediction: 
Efter att ha analyserat de tillgängliga skivorna har jag identifierat följande:

1. **[Point]** De associerade MindRAG-chunkarna är konsekvent kopplade till en anteckning om ett växellådsbyte. Detta tyder på att det kan ha funnits ett betydande problem med växellådan, men att bytet kan ha löst det.

2. **[Point]** MindRAG-chunkarna indikerar en ökande trend i enveloppsignalen, med omnämnanden av potentiella problem med utgående axel. Detta kan tyda på en utvecklande fel, möjligen relaterat till axeln eller lager.

3. **[Point]** Liknande MO DS HV, är MindRAG-chunkarna kopplade till en anteckning om ett växellådsbyte.

4. **[Point]** MindRAG-chunkarna är kopplade till en anteckning om ett sensorbyte. Detta kan indikera ett sensorfel, men det är inte klart om detta är huvudproblemet.

5. **[Point]** MindRAG-chunkarna är kopplade till anteckningar om ökande trender och potentiella axelproblem. Detta kan tyda på en utvecklande fel, möjligen relaterat till axeln eller lager.

Sammanfattningsvis finns det indikationer på potentiella axel- eller lagerproblem, men det finns också omnämnanden av sensorbyten. Jag misstänker att huvudproblemet kan vara relaterat till axeln eller lagren, men det kan också finnas sensorproblem. Jag behöver mer data för att avgöra om huvudproblemet är relaterat till axeln, lagren eller sensorerna. Vänligen ge fler skivor för vidare analys.

Annotation: 
Växellåda bytt på grund av oljeläckage. Lådan som sattes in har en annan utväxling än tidigare.

Evaluations: 
delvis korrekt

Reflections: 
Prediktionen identifierade korrekt att det fanns ett problem med växellådan, vilket stämmer överens med anteckningen. Dock nämnde prediktionen även potentiella problem med axeln eller lagren och sensorbyten, vilket inte nämndes i anteckningen. Detta kan bero på att prediktionen försökte täcka flera möjliga problemområden baserat på tillgängliga data, men huvudproblemet enligt anteckningen var växellådan och oljeläckaget.

Time delta: 
0
\end{lstlisting}

\subsubsection{Prediction 3}
\begin{lstlisting}[language={[LaTeX]TeX}]


Prediction: 
Efter att ha analyserat de tillgängliga MindRAG-chunkarna för de olika punkterna, verkar det som att det finns potentiella problem med sensorerna. Punkten [Point] är konsekvent associerad med MindRAG-chunkar som har annotationer som indikerar sensorbyten ('Givaren skall bytas nästa stopp') med tidsdifferenser från 0 till 15 dagar. Detta tyder på ett potentiellt sensorproblem, men tidsdifferenserna är inte omedelbara, vilket indikerar att det kanske inte är kritiskt ännu.

Punkten [Point] visar också liknande annotationer, som indikerar sensorbyten med tidsdifferenser runt 0 dagar, vilket också tyder på ett potentiellt sensorproblem.

Givet naturen av annotationerna och tidsdifferenserna verkar det som att det kan finnas mindre sensorproblem, men inget kritiskt vid denna tidpunkt. Det är troligt att den verkliga noteringen handlar om ett sensorproblem, men det är inte akut.

Annotation: 
Glättvalsar bytta 12/11-2020.

Evaluations: 
felaktig

Reflections: 
Prediktionen fokuserade på potentiella sensorproblem, medan den verkliga noteringen handlade om att glättvalsar byttes. Det finns ingen koppling mellan de två, vilket gör prediktionen felaktig. Det är möjligt att analysen missade tecken på problem med glättvalsarna eller att det inte fanns tillräcklig data för att indikera detta.

Time delta: 
0

\end{lstlisting}

\subsubsection{Prediction 4}
\begin{lstlisting}[language={[LaTeX]TeX}]

Prediction: 
Efter att ha analyserat de tillhandahållna MindRAG-chunkarna verkar det som att det finns en sensor- eller kabelrelaterad fråga. De flesta MindRAG-anteckningarna nämner 'Givaren skall bytas nästa stopp' (Sensorn ska bytas vid nästa stopp) och 'Troligen bytt kabel' (Troligen bytt kabel). Detta tyder på att det kan finnas en sensor- eller kabeldefekt. Inga andra allvarliga felindikatorer har identifierats i de analyserade chunkarna. Därför är min förutsägelse att den verkliga anteckningen handlar om en sensor- eller kabelrelaterad fråga.

Annotation: 
Glättvalsar bytta 12/11-2020.

Evaluations: 
felaktig

Reflections: 
Prediktionen fokuserade på en sensor- eller kabelrelaterad fråga, medan den verkliga anteckningen handlade om att glättvalsar byttes. Det finns ingen överensstämmelse mellan prediktionen och den verkliga anteckningen, vilket gör prediktionen felaktig. Det är möjligt att analysen missade tecken på problem med glättvalsarna eller att dessa inte var tydliga i de tillhandahållna MindRAG-chunkarna.

Time delta: 
7

\end{lstlisting}

\subsubsection{Prediction 5}
\begin{lstlisting}[language={[LaTeX]TeX}]


Prediction: 
Efter att ha analyserat de tillgängliga skivorna, har jag kommit fram till följande bedömning:

1. **[Point]**:
   - Det finns indikationer på att det kan ha funnits ett behov av underhåll eller byte av komponenter, med flera MindRAG-chunks associerade med anteckningar om byte av glättvalsar. Det finns också några associationer med "glapp" i sträck FS, vilket kan indikera potentiella problem med vibrationer eller instabilitet.

2. **[Point]**: 
   - Majoriteten av MindRAG-chunks är kopplade till kabelbyte, vilket starkt tyder på att det kan ha funnits problem med kablarna som krävde åtgärd.

3. **[Point]**:
   - Liksom den andra punkten, är många MindRAG-chunks kopplade till kabelbyte, vilket indikerar potentiella kabelproblem.

4. **[Point]**:
   - återigen, flera associationer med kabelbyte, vilket tyder på kabelproblem.

5. **[Point]**:
   - Flera MindRAG-chunks är kopplade till givare som ska bytas, vilket kan indikera problem med sensorerna.

6. **[Point]**:
   - Liksom den föregående punkten, finns det flera associationer med givare som ska bytas.

Sammanfattningsvis verkar det finnas en stark indikation på kabelproblem i flera punkter, särskilt i [Point] DS och FS. Det finns också indikationer på att givare kan behöva bytas i vissa punkter. Det är möjligt att den verkliga anteckningen handlar om kabelproblem, med tanke på den starka representationen av kabelbyte i MindRAG-chunks. Det kan också finnas underliggande problem med vibrationer eller instabilitet i [Point], men kabelproblemen verkar vara den mest framträdande frågan.

Annotation: 
AO skrivet på kabelbyte [Point].

Evaluations: 
korrekt

Reflections: 
Prediktionen identifierade korrekt kabelproblem som den mest framträdande frågan, vilket stämmer överens med den verkliga anteckningen om kabelbyte. även om prediktionen nämnde flera punkter och möjliga problem, var huvudfokuset på kabelproblem, vilket är i linje med den verkliga anteckningen.

Time delta: 
0

\end{lstlisting}

\subsubsection{Prediction 6}
\begin{lstlisting}[language={[LaTeX]TeX}]


Prediction: 
Efter att ha analyserat de tillgängliga MindRAG-chunkarna verkar det finnas ett sensorproblem. Anmärkningarna för punkterna [Point] och [Point] nämner konsekvent att sensorn bör bytas vid nästa stopp. Detta tyder på ett sensorfel. Det är dock oklart om detta är det primära problemet eller om det finns andra underliggande fel, såsom lagerproblem, som också kan vara närvarande. Utan ytterligare data är det svårt att göra en mer exakt bedömning, men baserat på den nuvarande informationen är det troligt att anmärkningen handlar om ett sensorfel.

Annotation: 
Givaren på bromsgenerator lagring DS skall bytas nästa stopp.

Evaluations: 
korrekt

Reflections: 
Prediktionen identifierade korrekt att det fanns ett sensorproblem och rekommenderade att sensorn skulle bytas vid nästa stopp, vilket stämmer överens med den sanna anmärkningen. Prediktionen nämnde också möjligheten av andra underliggande problem, men fokuserade på sensorfelet, vilket var korrekt enligt anmärkningen.

Time delta: 
0

\end{lstlisting}

\subsubsection{Prediction 7}
\begin{lstlisting}[language={[LaTeX]TeX}]

Prediction: 
Efter att ha analyserat de tillgängliga MindRAG-chunkarna och deras tillhörande anteckningar, verkar det finnas två möjliga problem. För den första punkten ([Point]) indikerar MindRAG-chunkarna konsekvent en BPFO-fel på ett lager, med anteckningar som tyder på att trenden har stannat och att det är en väntesituation. Detta tyder på ett potentiellt lagerfel, men det verkar vara stabilt för tillfället. För den andra punkten ([Point]) indikerar MindRAG-chunkarna sensorer som byts ut, vilket kan tyda på ett potentiellt sensorproblem. Det är möjligt att det finns ett underliggande lagerfel som tidigare missades på grund av sensorproblem. Min slutgiltiga bedömning är att det finns ett BPFO-lagerfel på [Point], men det kan också finnas ett sensorproblem på [Point]. Det är viktigt att överväga båda dessa möjligheter vid planering av underhållsåtgärder.

Annotation: 
Givaren utbytt på DS.

Evaluations: 
korrekt

Reflections: 
Prediktionen nämner ett potentiellt sensorproblem och att sensorer byts ut, vilket stämmer överens med noteringen om att givaren har bytts ut. även om prediktionen också nämner ett BPFO-lagerfel, är det inte i konflikt med noteringen eftersom det huvudsakliga problemet som identifierats i noteringen är sensorrelaterat, vilket prediktionen också täcker.

Time delta: 
0

\end{lstlisting}

\subsubsection{Prediction 8}
\begin{lstlisting}[language={[LaTeX]TeX}]


Prediction: 
Efter att ha analyserat de tillhandahållna skivorna visar det sig att det finns en konsekvent BPFO-fel (Ball Pass Frequency Outer race) på 'l[Point]' över flera punkter och MindRAG-chunks. Anmärkningarna indikerar att BPFO-trenden har låga nivåer, vilket tyder på en potentiell stabilisering av felet. Detta mönster är konsekvent över skivorna, och det finns inga indikationer på andra fel eller förändringar i felkarakteristiken. Därför misstänker jag att den verkliga anmärkningen kommer att beskriva ett BPFO-fel.

Annotation: 
BPFO på [Point], verkar ha avstannat på trenden så det är väl bara att vänta ett tag.

Evaluations: 
korrekt

Reflections: 
Prediktionen stämmer väl överens med den verkliga anmärkningen. Både prediktionen och anmärkningen identifierar ett BPFO-fel. Detta tyder på att analysen av data var korrekt och att inga andra fel eller förändringar upptäcktes.

Time delta: 
8

\end{lstlisting}

\subsubsection{Prediction 9}
\begin{lstlisting}[language={[LaTeX]TeX}]


Prediction: 
Efter att ha analyserat de tillgängliga MindRAG-chunkarna, verkar det som att det finns en stark indikation på sensorrelaterade problem, eftersom många anteckningar nämner 'Givaren skall bytas nästa stopp'. Detta tyder på potentiella sensorfel över flera punkter. Det finns dock också omnämnanden av lagerbyten i vissa anteckningar, vilket kan indikera underliggande lagerproblem. Det är möjligt att sensorerna har visat felaktiga mätningar som har lett till dessa slutsatser, men det kan också finnas en underliggande lagerfel som har upptäckts. Jag förutspår att den verkliga anteckningen kan handla om ett sensorbyte, men det är också möjligt att det finns en underliggande lagerproblem som har identifierats. Det är viktigt att kontrollera sensorerna och eventuellt byta ut dem för att säkerställa korrekta mätningar framöver.

Annotation: 
AO skrivet på lagerbyte drivsida.

Evaluations: 
delvis korrekt

Reflections: 
Prediktionen nämner både sensorproblem och lagerproblem, vilket delvis stämmer överens med den verkliga anteckningen som handlar om ett lagerbyte. Prediktionen är delvis korrekt eftersom den identifierar ett potentiellt lagerproblem, men den fokuserar mer på sensorproblem och föreslår ett sensorbyte som inte nämns i den verkliga anteckningen. Det kan bero på att analysen av MindRAG-chunkarna visade indikationer på sensorproblem som inte var relevanta för den aktuella anteckningen.

Time delta: 
4

\end{lstlisting}

\subsubsection{Prediction 10}
\begin{lstlisting}[language={[LaTeX]TeX}]

Prediction: 
Efter att ha analyserat de tillgängliga skivorna har jag identifierat följande potentiella problem:

1. **[Points]**: Dessa punkter är konsekvent associerade med BPFI-anteckningar, vilket indikerar en potentiell lagerfel. Detta är den mest sannolika orsaken till anteckningen.

2. **[Point]**: Denna punkt är associerad med en växellådsbyte på grund av oljeläckage. Detta verkar inte vara direkt relaterat till ett lagerfel, men kan indikera ett underliggande problem som kan ha påverkat andra komponenter.

3. **[Point]**: Denna punkt är också associerad med växellådsbyte-anteckningar, vilket tyder på att det kan ha funnits problem med växellådan som kan ha påverkat mätningarna.

4. **[Point]**: Denna punkt är associerad med potentiella sensor- eller kabelproblem, vilket kan ha påverkat mätningarna och orsakat falska larm.

Sammanfattningsvis verkar det mest sannolikt att anteckningen handlar om ett lagerfel (BPFI) vid punkterna [Points]. Det är dock viktigt att överväga att det kan finnas andra underliggande problem, såsom växellådsproblem eller sensor/kabelproblem, som kan ha påverkat mätningarna.

Annotation: 
BPFI syns tydligt i [Point] Håll koll på denna

Evaluations: 
delvis korrekt

Reflections: 
Prediktionen identifierade korrekt ett potentiellt BPFI-lagerfel vid [Point], vilket stämmer överens med noteringen. Dock inkluderade prediktionen även andra potentiella problem som inte nämndes i noteringen, såsom växellådsproblem och sensor/kabelproblem. Detta gör att prediktionen är delvis korrekt, eftersom den inte helt fokuserade på det specifika problemet som noteringen beskrev.

Time delta: 
8

\end{lstlisting}

\subsection{Evaluation With Data After Maintenance}

\subsubsection{Prediction 1}
\begin{lstlisting}[language={[LaTeX]TeX}]
Prediction: 
Efter att ha analyserat de tillgängliga MindRAG-chunkarna, verkar det som att det primära problemet är relaterat till sensorer eller kablar. Många av de associerade anteckningarna nämner 'Givaren skall bytas nästa stopp', vilket tyder på ett återkommande problem med sensorerna eller kablarna. Detta mönster observeras över flera punkter och tidsdifferenser, vilket stärker misstanken om ett sensor- eller kabelproblem.

Det finns dock också tillfälliga omnämnanden av andra problem, såsom ett lagerbyte ('lagerbyte'), vilket kan indikera underliggande problem som inte enbart är relaterade till sensorer eller kablar.

Sammanfattningsvis, baserat på den tillgängliga informationen, är det troligt att den verkliga anteckningen beskriver ett problem med sensorer eller kablar, men det kan också finnas underliggande lagerproblem som inte är lika framträdande i de tillgängliga chunkarna. Det är viktigt att överväga både sensor- och lagerproblem vid vidare analys och underhåll.

Annotation: 
Glättvalsar bytta 12/11-2020.

Evaluations: 
felaktig

Reflections: 
Prediktionen fokuserar på sensor- eller kabelproblem och nämner även potentiella lagerproblem, medan den verkliga anteckningen handlar om att glättvalsar har bytts. Det finns ingen direkt koppling mellan prediktionen och den verkliga åtgärden som utförts, vilket gör att prediktionen inte stämmer överens med anteckningen.

Time delta: 
17
\end{lstlisting}

\subsubsection{Prediction 2}
\begin{lstlisting}[language={[LaTeX]TeX}]
Prediction: 
Efter att ha analyserat de tillgängliga MindRAG-chunkarna och deras tillhörande anteckningar, verkar det som att det finns en blandning av sensorrelaterade problem och underhållsåtgärder. Majoriteten av MindRAG-chunkarna är kopplade till anteckningar som indikerar att sensorn bör bytas vid nästa stopp, vilket tyder på potentiella sensorproblem. Det finns också anteckningar som indikerar att 'Glättvalsar' har bytts ut, vilket kan innebära att en underhållsåtgärd har utförts.

Med tanke på den nuvarande datan är det inte klart om det primära problemet är ett sensorfel eller en underhållsåtgärd relaterad till 'Glättvalsar'. Det är möjligt att sensorn har visat tecken på felaktighet, vilket har lett till rekommendationen att byta ut den, men det kan också vara så att underhållsåtgärden på 'Glättvalsar' var den huvudsakliga anledningen till anteckningen. Utan ytterligare data är det svårt att fastställa vilken av dessa som är den verkliga orsaken till anteckningen.

Annotation: 
Glättvalsar bytta 12/11-2020.

Evaluations: 
delvis korrekt

Reflections: 
Prediktionen nämner både sensorproblem och underhållsåtgärder, vilket delvis stämmer överens med anteckningen som handlar om att glättvalsar har bytts. Prediktionen är delvis korrekt eftersom den identifierar en underhållsåtgärd men också introducerar osäkerhet kring sensorproblem som inte nämns i anteckningen. Det är möjligt att prediktionen försöker täcka flera möjliga scenarier utan att ha tillräcklig data för att prioritera rätt åtgärd.

Time delta: 
17
\end{lstlisting}

\subsubsection{Prediction 3}
\begin{lstlisting}[language={[LaTeX]TeX}]
Prediction: 
Efter att ha analyserat de tillgängliga MindRAG-chunkarna för de två punkterna, verkar det som att det finns en potentiell sensor- eller kabelproblem ("givare") i systemet. För den första punkten ([Point]) indikerar flera MindRAG-anteckningar att sensorn på drivsidan (DS) av bromsgeneratorn är planerad för utbyte, vilket tyder på ett problem med sensorn. Den andra punkten ([Point]) visar liknande anteckningar relaterade till sensorutbyten i andra delar av maskineriet. Det finns dock en notering om BPFO (lagerfel) i en av MindRAG-chunkarna för den första punkten, men det är inte den dominerande anteckningen. Det verkar som att sensorproblemet är det primära problemet som motiverade anteckningen, men det kan finnas underliggande lagerfel som också bör övervakas. Mitt slutliga antagande är att anteckningen främst handlar om ett sensorproblem, men att det kan finnas en låg nivå av lagerfel som också bör beaktas.

Annotation: 
Givaren utbytt på DS.

Evaluations: 
korrekt

Reflections: 
Prediktionen identifierade korrekt ett sensorproblem som det primära problemet, vilket stämmer överens med noteringen om att givaren byttes ut på drivsidan (DS). Prediktionen nämnde också ett potentiellt lagerfel, men det var inte det dominerande problemet, vilket är i linje med noteringen. Prediktionen var därför korrekt i att identifiera sensorproblemet som det primära skälet för åtgärden.

Time delta: 
18
\end{lstlisting}

\subsubsection{Prediction 4}
\begin{lstlisting}[language={[LaTeX]TeX}]
Prediction: 
Efter att ha analyserat de tillgängliga MindRAG-chunkarna, verkar det som att det finns potentiella problem med sensorn, eftersom många av noteringarna nämner att sensorn ska bytas ("Givaren skall bytas nästa stopp" och "Givare utbytt"). Det finns inga tydliga indikationer på andra fel, såsom lagerfel eller kabelproblem, i de tillgängliga data. Därför är min bedömning att noteringen sannolikt handlar om ett sensorproblem som har lett till att sensorn behöver bytas ut. Inga andra allvarliga fel verkar vara närvarande i de analyserade chunkarna.

Annotation: 
Givare utbytt.

Evaluations: 
korrekt

Reflections: 
Prediktionen identifierade korrekt att det fanns ett sensorproblem som ledde till att sensorn behövde bytas ut, vilket stämmer överens med den sanna noteringen 'Givare utbytt'. Prediktionen nämnde också att inga andra allvarliga fel var närvarande, vilket är i linje med noteringen.

Time delta: 
18
\end{lstlisting}

\subsubsection{Prediction 5}
\begin{lstlisting}[language={[LaTeX]TeX}]
Prediction: 
Efter att ha analyserat de tillgängliga MindRAG-chunkarna, verkar det som att det finns ett återkommande mönster av sensorfel, vilket indikeras av flera anteckningar som nämner 'Givaren skall bytas nästa stopp'. Detta mönster är särskilt tydligt för punkterna [Points]. Det finns inga starka indikationer på andra typer av fel, såsom lagerfel, i de analyserade chunkarna. Därför är min bedömning att den verkliga anteckningen sannolikt handlar om ett sensorfel och att sensorn behöver bytas ut vid nästa stopp.

Annotation: 
Givaren skall bytas nästa stopp. Orsaken är att då givaren byttes senast så skuttade totalnivån upp ganska mycket men då man lyssnar ute på plats verkar allt ok. Kan det vara något fel på givaren som sitter där nu eller?

Evaluations: 
korrekt

Reflections: 
Prediktionen och den verkliga anteckningen stämmer väl överens. Båda indikerar ett problem med sensorn och föreslår att den ska bytas vid nästa stopp. Prediktionen fångar också upp att det inte finns några starka indikationer på andra typer av fel, vilket stämmer med att allt verkar ok när man lyssnar på plats. Detta tyder på att analysen var korrekt och att systemet har identifierat sensorproblemet på ett tillfredsställande sätt.

Time delta: 
16
\end{lstlisting}

\subsubsection{Prediction 6}
\begin{lstlisting}[language={[LaTeX]TeX}]
Prediction: 
Efter att ha analyserat de tillgängliga MindRAG-chunkarna, verkar det som att det finns potentiella sensorproblem, eftersom flera anteckningar nämner 'Givaren skall bytas nästa stopp'. Detta observeras i båda punkterna, '[Point]' och '[Point]'. Det finns också omnämnanden av lagerbyten i vissa MindRAG-chunkar, vilket kan indikera underliggande problem. Det är möjligt att sensorerna har problem som behöver åtgärdas, men det kan också finnas en underliggande lagerfel som inte är lika akut. Utan fler data är det svårt att avgöra om dessa är konsekventa mönster eller isolerade fall. Min slutgiltiga bedömning är att det troligen handlar om sensorproblem, men det kan också finnas en låg nivå av lagerfel som inte är akut.

Annotation: 
Troligen bytt kabel.

Evaluations: 
korrekt

Reflections: 
Prediktionen identifierade sensorproblem, vilket är korrekt eftersom kabel- och sensorproblem ofta är svåra att skilja åt och kan betraktas som samma typ av problem i detta sammanhang. Noteringen indikerar att en kabel troligen har bytts, vilket stämmer överens med prediktionens fokus på sensorproblem. Prediktionen nämner också potentiella lagerproblem, men dessa bedöms inte som akuta, vilket inte påverkar den övergripande korrektheten i förhållande till noteringen.

Time delta: 
18
\end{lstlisting}

\subsubsection{Prediction 7}
\begin{lstlisting}[language={[LaTeX]TeX}]
Prediction: 
Efter att ha analyserat totalt sextiofem skivor av data, visar MindRAG-chunkarna en blandning av potentiella sensorproblem och lagerbyten. Det finns konsekventa omnämnanden av sensorer som behöver bytas ut, samt anteckningar om lagerbyten på både fri sida (FS) och drivsida (DS). Det är dock inte tillräckligt tydligt för att avgöra vilken specifik fel eller underhållsåtgärd den verkliga anteckningen kan beskriva. Det är möjligt att det finns en underliggande lagerfel som har lett till dessa åtgärder, men det kan också vara så att sensorerna har orsakat felaktiga mätningar. Utan ytterligare data är det svårt att göra en exakt bedömning av den verkliga anteckningen.

Annotation: 
Valsen utbytt på grund av haveriet [Date] på drivsidan.

Evaluations: 
delvis korrekt

Reflections: 
Prediktionen nämner både sensorproblem och lagerbyten, vilket delvis stämmer överens med den verkliga anteckningen som handlar om ett haveri och byte av valsen på drivsidan. Prediktionen är dock ospecifik och kan inte exakt identifiera haveriet som orsakade bytet, vilket gör den delvis korrekt. Det är möjligt att prediktionen inte hade tillräckligt med data för att göra en mer exakt bedömning.

Time delta: 
17
\end{lstlisting}

\subsubsection{Prediction 8}
\begin{lstlisting}[language={[LaTeX]TeX}]
Prediction: 
Efter att ha analyserat de tillgängliga MindRAG-chunkarna och deras tillhörande anteckningar, verkar det som att de flesta chunkarna indikerar en sensor- eller kabelrelaterad felanmärkning. De flesta MindRAG-chunkarna är associerade med anteckningar som beskriver sensorbyte eller kabelproblem, och tidsskillnaderna är nära anteckningsdatumet, vilket tyder på att dessa chunkar är relaterade till en sensor- eller kabelfelanmärkning. Det finns inga tydliga indikationer på allvarligare fel som lagerfel eller haveri i de tillgängliga chunkarna. Därför är min bedömning att den verkliga anteckningen sannolikt beskriver ett sensor- eller kabelrelaterat problem, och inga allvarliga maskinfel är närvarande i de analyserade chunkarna.

Annotation: 
Kabeln troligen bytt.

Evaluations: 
korrekt

Reflections: 
Prediktionen identifierade korrekt att det fanns ett kabel- eller sensorrelaterat problem, vilket stämmer överens med den verkliga anteckningen som indikerar att kabeln troligen har bytts. Prediktionen och anteckningen är i linje med varandra, och det finns ingen indikation på allvarligare fel, vilket också stämmer med anteckningen.

Time delta: 
18
\end{lstlisting}

\subsubsection{Prediction 9}
\begin{lstlisting}[language={[LaTeX]TeX}]
Prediction: 
Efter att ha analyserat trettio skivor av data, visar den första punkten konsekvent MindRAG-chunks som är associerade med glapp ("glapp"), medan den andra och tredje punkten indikerar sensorproblem med noteringen "Givaren skall bytas nästa stopp". Mönstren har förblivit konsekventa över alla skivor. Baserat på den nuvarande analysen verkar det troligt att anteckningen kan vara relaterad till sensorproblem, men förekomsten av glapp i den första punkten bör inte ignoreras. Det är möjligt att den verkliga anteckningen handlar om sensorproblem, men det kan också finnas en underliggande glapp-problematik som behöver åtgärdas.

Annotation: 
Valsen bytt.

Evaluations: 
felaktig

Reflections: 
Prediktionen fokuserade på sensorproblem och glapp, medan den verkliga anteckningen indikerade att valsen hade bytts. Det finns ingen direkt koppling mellan de förutsagda problemen och den faktiska åtgärden som utfördes. Det är möjligt att analysen missade tecken på problem med valsen eller att dessa problem inte var uppenbara i de data som analyserades.

Time delta: 
8
\end{lstlisting}

\subsubsection{Prediction 10}
\begin{lstlisting}[language={[LaTeX]TeX}]
Time delta: 
8
Prediction: 
Efter att ha analyserat de tillgängliga MindRAG-chunkarna och deras tillhörande anteckningar, har jag identifierat flera potentiella problem:

1. **[Point]**: Det finns konsekventa indikationer på ett lagerproblem, med arbetsorder för lagerbyte både FS och DS nära anteckningsdatumet. Detta tyder på att ett lagerproblem identifierades och schemalades för byte.

2. **[Point]**: Det finns tydliga indikationer på BPFO, men på låga nivåer, vilket tyder på tidiga stadier av ett lagerfel.

3. **[Point]*: Det finns frekventa omnämnanden av smörjproblem, med anteckningar om stigande envelopevärden och behovet av att kontrollera smörjningen.

Baserat på dessa observationer, verkar det mest sannolikt att den verkliga anteckningen handlar om ett lagerproblem, särskilt med tanke på arbetsordern för lagerbyte. Det är också möjligt att smörjproblem kan ha påverkat sensoravläsningarna, men huvudfokus verkar vara på lagerproblemet. Det är viktigt att övervaka både smörjningen och de tidiga BPFO-indikationerna för att förhindra ytterligare problem.

Annotation: 
Vals utbytt på grund av lagerskada på drivsidan.

Evaluations: 
korrekt

Reflections: 
Prediktionen identifierade korrekt ett lagerproblem och noterade att det fanns arbetsorder för lagerbyte, vilket stämmer överens med den verkliga anteckningen om att valsen byttes på grund av lagerskada på drivsidan. Prediktionen nämnde också tidiga indikationer på BPFO, vilket kan relatera till den identifierade lagerskadan. Det fanns också omnämnanden av smörjproblem, men huvudfokus i prediktionen var på lagerproblemet, vilket var korrekt.

Time delta: 
18
\end{lstlisting}
\end{document}